\documentclass[11pt]{article}
\usepackage[T1]{fontenc}
\usepackage[utf8]{inputenc}
\usepackage{lmodern,amsmath,amssymb,amsthm,mathtools}
\usepackage[margin=1in]{geometry}
\usepackage{microtype,booktabs,longtable,array,enumitem}
\usepackage{tikz}
\usetikzlibrary{arrows.meta,positioning}
\usepackage{aliascnt}
\usepackage[hidelinks]{hyperref}
\usepackage[nameinlink,noabbrev]{cleveref}
\allowdisplaybreaks
\newtheorem{theorem}{Theorem}[section]
\newaliascnt{lemma}{theorem}
\newtheorem{lemma}[lemma]{Lemma}
\aliascntresetthe{lemma}
\newaliascnt{proposition}{theorem}
\newtheorem{proposition}[proposition]{Proposition}
\aliascntresetthe{proposition}
\newaliascnt{corollary}{theorem}

\aliascntresetthe{corollary}
\theoremstyle{definition}
\newaliascnt{definition}{theorem}
\newtheorem{definition}[definition]{Definition}
\aliascntresetthe{definition}
\newaliascnt{remark}{theorem}

\aliascntresetthe{remark}
\crefname{lemma}{lemma}{lemmas}
\Crefname{lemma}{Lemma}{Lemmas}
\crefname{proposition}{proposition}{propositions}
\Crefname{proposition}{Proposition}{Propositions}
\crefname{corollary}{corollary}{corollaries}
\Crefname{corollary}{Corollary}{Corollaries}
\crefname{definition}{definition}{definitions}
\Crefname{definition}{Definition}{Definitions}
\crefname{remark}{remark}{remarks}
\Crefname{remark}{Remark}{Remarks}
\numberwithin{equation}{section}
\DeclareMathOperator{\TV}{TV}
\DeclareMathOperator{\KL}{KL}
\DeclareMathOperator{\VC}{VC}
\DeclareMathOperator{\Var}{Var}
\DeclareMathOperator{\Unif}{Unif}
\DeclareMathOperator{\vol}{vol}
\newcommand{\E}{\mathbb E}
\newcommand{\R}{\mathbb R}
\newcommand{\ind}{\mathbf1}
\newcommand{\HH}{\mathsf h}
\newcommand{\eps}{\varepsilon}
\newcommand{\cstar}{c_\star}
\newcommand{\proofref}[1]{The complete proof is in \Cref{#1}.}
\title{Strictly Replicable Realizable Learning:\\Unknown Marginals, Finite-Class Interpolation,\\and the Cost of Shared Randomness}
\author{Guangjian Zhang\thanks{Email: \texttt{zgj1226029469@outlook.com}.}}
\date{October 23, 2026}
\hypersetup{pdftitle={Strictly Replicable Realizable Learning: Unknown Marginals, Finite-Class Interpolation, and the Cost of Shared Randomness},pdfauthor={Guangjian Zhang}}
\begin{document}
\maketitle
\begin{abstract}
Strict replicability requires two independent training samples, processed with the same random seed, to produce exactly the same hypothesis. We study realizable Boolean learning under unknown and known instance marginals. For unknown marginals, we obtain the worst-case sample complexity
$\Theta(((\log M)^2/\rho^2+\log(1/\delta))/\eps)$ for classes of cardinality at most $M$, throughout a fixed small-parameter range. This resolves the distribution-free finite-class sample-complexity question up to universal constants and rules out a general $\rho^{-1}$ dependence. The upper bound uses a truncated-power distribution with a hard empirical-error budget. For known marginals, a control-variate learner uses $O(((1+\log M)^{3/2}/\rho+\log(1/\delta))/\eps)$ samples, and explicit uniform-marginal instances establish the matching class-size exponent. On a product of $d$ finite threshold chains of length $k$, we determine the exact rate $\Theta((dk\log(1+\sqrt d/(k\rho))+\log(1/\delta))/\eps)$. We also characterize deterministic learning, the very-small-$\rho$ regime for arbitrary finite classes, and every cardinality--replicability regime for finite classes of VC dimension one. A VC--covering-number bound yields the fixed-accuracy exponent $3/2$ independently of cardinality. Finally, we separate statistical complexity from implementation costs: we prove a sharp fixed-block trade-off, short-seed existence bounds, and a shared rejection-filter reduction from optimal learning of the full Boolean class to efficient public sampling of ordinary Bernoulli product distributions. Detailed proofs include the geometric lower bounds, the discrete posterior-kernel comparison, and the finite-bit constructions.
\end{abstract}
\tableofcontents
\section{Introduction}
Two statistically accurate classifiers need not be the same classifier. Replicable learning asks for a stronger form of reproducibility: independent datasets must lead to the same output when the learner reuses its random seed. The definition introduced by Impagliazzo, Lei, Pitassi, and Sorrell \cite{ILPS} makes this requirement algorithmic. The randomness is shared; the observations are not. For a finite hypothesis class, this creates a basic question: how much more data is needed to make the entire prediction function repeatable?

The finite-class problem sits between ordinary PAC learning and high-dimensional stability. A consistent learner needs only logarithmically many samples in the class size at fixed accuracy and confidence. General replicable constructions have quadratic logarithmic dependence, up to additional parameter factors \cite{BGH,BBE}. Larsen, Mathiasen, Pabbaraju, and Svendsen \cite{LMPS} constructed a realizable class with a nearly $(\log |H|)^{3/2}$ lower bound and a nearly matching upper bound for that uniform-marginal instance. In their open-problems discussion \cite[Section 1.1]{LMPS}, they asked for the exact finite-class sample complexity, leaving the choice between class-size exponents $3/2$ and two unresolved. They also tentatively conjectured that inverse-linear dependence on the replicability parameter might suffice for arbitrary classes and distributions.

\paragraph{Resolution of the finite-class question.}
We resolve this distribution-free sample-complexity question up to universal constant factors. \Cref{thm:u} determines the joint worst-case dependence on cardinality, accuracy, confidence, and replicability throughout \eqref{eq:domain}. The class-size exponent is two, and the replication cost is inverse-quadratic. The latter gives a negative answer to the inverse-linear conjecture. Our earlier version, published on September 9, 2026 \cite{ZhangOriginal}, established the anchored realizable lower bound underlying both conclusions. The truncated-power learner developed here matches that bound jointly in all four parameters, with an additive confidence term.

The distinction between unknown and known marginals organizes the rest of the paper. A distribution-free learner must work across varying marginals; a learner that knows the marginal can use its geometry directly. We construct a general known-marginal learner, determine exact finite-grid rates, and quantify implementation costs.

\paragraph{Three obstacles and the mechanisms that address them.}
First, controlling every empirical risk separately wastes logarithmic factors. We instead control the change of an entire distribution over hypotheses. For unknown marginals, deleting one observation from a truncated-power distribution has small total squared Hellinger energy. A hard cutoff makes accuracy hold simultaneously for all seeds. For known marginals, an initial predictor supplies a control variate. Its mistakes, rather than the full loss, determine the variance of the correction. A reverse-KL estimate then controls the normalized distribution without a maximum over hypothesis-wise errors.

Second, a finite class behaves differently when $\rho$ becomes very small. A continuum approximation cannot see the final discrete scale. We retain the finite threshold grid in the lower bound. Its posterior is uniform on random intervals of unobserved cuts. Comparing this kernel to a compressed Neumann resolvent and then to Gaussian noise produces the exact logarithmic interpolation. On the upper-bound side, the mean risk of a consistent threshold anchor has the same discrete saturation.

Third, sampling one correct output distribution is weaker than coupling all runs with one shared random tape. We treat the coupling as an explicit algorithmic component. Fixed-block learners have a matching locality penalty. Finite lookup tables show that very short shared seeds suffice statistically. A rejection filter then reduces the optimal full-dimensional implementation problem to public sampling of untruncated Bernoulli products, while preserving additive confidence cost.

\paragraph{Contributions.}
Our results describe three resources: samples, shared random bits, and computation. Let $M$ bound the class cardinality, let $\eps$ be the accuracy tolerance, and let $\delta$ and $\rho$ bound accuracy failure and disagreement between two runs. The rates below use the parameter ranges specified in their linked theorems.
\begin{enumerate}[leftmargin=*,itemsep=.45em]
\item \textbf{The optimal unknown-marginal finite-class rate.}
We construct a proper learner with sample complexity
\[
\Theta\left(\frac{(\log M)^2}{\eps\rho^2}
+\frac{\log(1/\delta)}\eps\right)
\]
in the worst case. The new truncated-power upper bound matches our earlier anchored lower bound. A smooth weight profile controls replication, while a hard empirical cutoff gives the additive confidence term. \Cref{thm:u,thm:anchor} state the two directions; \Cref{sec:foundations,sec:proof-u} give their complete proofs.

\item \textbf{Known marginals reduce both principal exponents.}
A control-variate learner uses
$O(((1+\log M)^{3/2}/\rho+\log(1/\delta))/\eps)$ samples. Uniform-marginal constructions establish the class-size exponent $3/2$ and the optimal inverse-replicability exponent one when the remaining dependence is polynomial in $\log M$. These are general exponent conclusions, distinct from the exact fixed-class formula below. See \Cref{thm:k,sec:grid-consequences}.

\item \textbf{An exact finite-grid interpolation and two saturation results.}
On a product of $d$ threshold chains with $k+1$ choices per chain, we prove
\[
\Theta\left(\frac{dk}\eps\log\left(1+\frac{\sqrt d}{k\rho}\right)
+\frac{\log(1/\delta)}\eps\right).
\]
The same formula connects inverse-$\rho$ growth to logarithmic growth on a fixed finite grid. A separate rank-adaptive construction gives the deterministic finite-class minimax rate $\Theta((M\log(1/\rho)+\log(1/\delta))/\eps)$; this is also the unrestricted rate for $\rho\le1/(16M^2)$. \Cref{thm:grid,thm:det} state the results. The continuous posterior construction and its explicit cardinality translation are given in \Cref{sec:continuous-kernel}.

\item \textbf{Intrinsic complexity and a complete VC-one characterization.}
For VC dimension $v$, a validated one-inclusion anchor and a proper cover yield
$O((v^{3/2}\log(C/\eps)/\rho+\log(1/\delta))/\eps)$ samples, independently of class cardinality. The fixed-accuracy VC exponent is sharp. For VC dimension at most one, a shared-phase tree learner gives the exact cardinality-dependent rate
\[
\Theta\left(\frac M\eps\log\left(1+\frac1{M\rho}\right)
+\frac{\log(1/\delta)}\eps\right).
\]
It uses $O(\log(1/\rho))$ shared bits and can be made proper. On uniform threshold classes, an explicit implementation avoids enumerating either the class or the domain. See \Cref{thm:vc,thm:vc1,sec:compact-threshold}.

\item \textbf{A sharp locality cost.}
For the full Boolean class on $d$ points, fixed blocks of size at most $s$ yield the rate
$O(d^3/(s\eps\rho^2)+\log(1/\delta)/\eps)$ with bounded finite-bit running time $\operatorname{poly}(n,d,\log(1/\rho))2^{O(s)}$, including parameter-encoding costs. For equal-sized blocks, a matching lower bound holds in the fixed-block observation model. Its proof controls dependence between blocks at a fixed sample size. \Cref{thm:block} and \Cref{sec:locality-proof} give the algorithm and the model-specific optimality statement.

\item \textbf{Short seeds at the optimal statistical rate.}
Finite-grid compression achieves the optimal unknown-marginal sample order on the full Boolean class using only $O(\log d+\log\log(n+1)+\log(1/\rho))$ shared bits, with an explicit finite-table storage bound. We also prove an alternative distribution-averaged compression, an $\Omega(1/\rho)$ lower bound on the number of seed values already on a two-point domain, and an atomic-seed obstruction for a continuous zero-additive-error interface. \Cref{thm:seed,prop:atomic,sec:seed-average} separate these guarantees.

\item \textbf{A public filtering reduction with bounded stopping cost.}
We replace hard-truncated proposal distributions by untruncated Gibbs proposals and a shared empirical-risk filter. The filter makes at most $4/3$ proposals in expectation and $O(\log(1/\rho))$ after truncation, while retaining the optimal finite-class sample order and additive confidence. On the full Boolean class the proposal is an ordinary Bernoulli product. Thus an efficient constant-distortion public sampler for the relevant product family suffices for optimal-rate efficient learning. The reduction and an explicit finite-list implementation are proved in \Cref{thm:filter,sec:finite-proposal}.

\item \textbf{Positive compression tools and explicit mechanism obstructions.}
A fixed sufficient statistic preserves TV and Hellinger distances exactly, giving a public sampler for a fixed-baseline one-parameter product family through a dynamic program using $O(d^2)$ rational operations. Complementary results quantify the costs of black-box access, coordinatewise coupling, shared-capacity greedy rounding, a cardinality-first knapsack objective, and sample-dependent cost compression. The knapsack obstruction holds for two samples from one fixed distribution, with an explicit $d^4/(\eps\rho^2)$ small-$\rho$ asymptotic. Each mechanism and its guarantee has a separate entry in \Cref{sec:result-index} and a proof in \Cref{sec:obstructions}.
\end{enumerate}

\paragraph{Reading order.}
\Cref{sec:model,sec:statistical} introduce the interfaces and develop the statistical modules in the order above. \Cref{sec:computation} then treats implementation and shared randomness. Each module specifies its inputs, actions, output, and direct guarantee. The appendices supply complete proofs in dependency order. Two diagrams in \Cref{sec:proof-map} show the principal dependencies; \Cref{sec:result-index} also locates the auxiliary results and alternative constructions individually.

\paragraph{Related work.}
Our truncated-power construction attains the optimal unknown-marginal finite-class rate jointly in all four parameters, with additive $\log(1/\delta)/\eps$ and no extra logarithmic factors. This sharpens the general replicable constructions of Bun et al.\ \cite{BGH} and the additive-confidence guarantees of Banihashem et al.\ \cite{BBE}. Bun et al.\ also prove quadratic-dimensional lower bounds in the agnostic setting. Our anchored family establishes the joint quadratic obstruction under realizability. The shared-sampling step fits the correlated-sampling framework of Kalavasis et al.\ \cite{KKMV}; \Cref{lem:clock,sec:foundations-gibbs} give the concrete couplings used by our learners.

Our anchored lower bound combines a square-mass realizable experiment, an explicit accuracy-volume estimate, and a ball-expansion argument. These ingredients yield the joint $d^2/(\eps\rho^2)$ obstruction, including improper learners. Its geometric setting connects to the isoperimetric approach to high-dimensional replicability developed by Hopkins et al.\ \cite{HIKLY}. Dewasurendra \cite{Dewa} studies average-distortion partitions for distribution estimation, a related statistical problem. \Cref{sec:anchor-proof} proves our realizable PAC bound directly from the anchored experiment and its volume estimates.

For known marginals, we derive posterior-kernel comparisons and finite-grid estimates, using Borell's Gaussian noise-stability inequality in the form stated by Eldan \cite{Eldan}. The VC refinement combines the classical one-inclusion expected-risk guarantee \cite{HLW} with our validated anchor and control-variate learner. The finite-class Gibbs analysis in \Cref{sec:foundations-gibbs} uses an expected KL--generalization identity connected to the work of Wang, Lei, and Fienberg \cite{WLF}.

\section{Model and notation}\label{sec:model}
Let $H=\{h_1,\ldots,h_m\}$ be a finite class of measurable Boolean functions on a measurable space $X$. For a marginal $p$ and target $h^*\in H$, let $D_{p,h^*}$ denote the distribution of $(X,h^*(X))$ with $X\sim p$. Write
\[
L(f)=\Pr_{X\sim p}[f(X)\ne h^*(X)],\qquad
d_p(f,g)=\Pr_{X\sim p}[f(X)\ne g(X)].
\]
All equality events and parameterized maps used below are measurable. For finite $H$, its finitely many evaluation patterns provide a finite measurable reduction when needed. Lower-bound instance spaces are finite, so even improper Boolean outputs form a finite set.

\begin{definition}[Strict replicable PAC learning]\label{def:model}
A learner $A$ with sample budget $n$ is $(\eps,\delta,\rho)$-valid if, for every admitted realizable distribution,
\begin{equation}\label{eq:model}
\Pr_{S,r}[L(A(S;r))>\eps]\le\delta,
\qquad
\Pr_{S,S',r}[A(S;r)\ne A(S';r)]\le\rho.
\end{equation}
Here $S,S'$ are independent samples of size $n$ and $r$ is a shared seed independent of both. Equality means equality at every point of $X$, including points of zero $p$-mass. All output randomness belongs to $r$. Upper bounds below return a member of $H$; lower bounds permit arbitrary measurable Boolean outputs.
\end{definition}

In interface $U$, one algorithm must work for every unknown marginal and every target. In interface $K$, the quantifiers are $\forall p\,\exists A_p\,\forall h^*\in H$: the algorithm knows $p$, but not the target. Distances under $p$ can be used exactly in statistical constructions. Computational statements specify separately how such distances are represented. Interface $K$-unif restricts $p$ to uniform distributions on finite domains.

Let $N_H^U$ and $N_{H,p}^K$ denote the corresponding minimum sample budgets. Define
\begin{equation}\label{eq:minimax}
N^{U,*}(M)=\sup_{|H|\le M}N_H^U,
\qquad N^{K,*}(M)=\sup_{|H|\le M}\sup_p N_{H,p}^K,
\end{equation}
with $(\eps,\delta,\rho)$ suppressed only when unambiguous. The domain may vary in the supremum. Write $N^{K\text{-unif},*}$ for its uniform-marginal version. Singleton classes need no samples. A known-marginal worst-case lower bound is witnessed by a specified marginal; the algorithm is still allowed to know it.

We use $P_d=\{0,1\}^{[d]}$ for the full Boolean class. A fixed enumeration resolves every tie and selects every component representative. This also fixes the whole function when two hypotheses have $d_p$-distance zero. For distributions on a finite set,
\[
\TV(P,Q)=\tfrac12\sum_h|P(h)-Q(h)|,
\qquad \HH(P,Q)=\|\sqrt P-\sqrt Q\|_2.
\]
The same definitions use integrals for densities. Logarithms are natural unless a base is shown. Constants hidden by $O,\Omega,\Theta$ are absolute unless a subscript explicitly indicates otherwise. We use the common small-parameter domain
\begin{equation}\label{eq:domain}
0<\eps\le1/16,\qquad 0<\delta,\rho\le10^{-4}
\end{equation}
for joint lower-bound comparisons. Individual upper bounds have wider ranges, stated in their theorems.

\section{Statistical modules and their guarantees}\label{sec:statistical}
\subsection{Unknown marginals: sample a hypothesis within a hard error budget}\label{sec:module-u}
The input is a realizable sample $S$ and an enumeration of $H$. Count the errors $C_h$ of each hypothesis. Set
\[
k=\max\{4,\lceil2\log m\rceil\},\qquad B=\lfloor n\eps/2\rfloor,
\qquad w_S(h)=(1-C_h/B)_+^k.
\]
A consistent hypothesis has weight one. Use independent shared exponential clocks $E_h$ and return the index minimizing $E_h/w_S(h)$, treating a zero weight as an infinite score. This returns a proper hypothesis with fewer than $B$ empirical errors. The cutoff is responsible for accuracy; the smooth power near the cutoff is responsible for stability.

\begin{theorem}[Optimal unknown-marginal finite-class rate]\label{thm:u}
For $m\ge2$ and $0<\eps,\delta,\rho<1$, the preceding learner is valid with
\begin{equation}\label{eq:u-budget}
n=\left\lceil\max\left\{\frac{32k^2}{\eps\rho^2},
\frac{8\log(m/\delta)}\eps\right\}\right\rceil.
\end{equation}
For $M\ge2$ in \eqref{eq:domain},
\[
N^{U,*}(M)=\Theta\left(\frac{(\log M)^2}{\eps\rho^2}
+\frac{\log(1/\delta)}\eps\right).
\]
\end{theorem}
\proofref{sec:proof-u} The upper bound controls the expected change of the normalized square-root weight vector when observations are deleted. The matching obstruction is our anchored lower bound from the earlier version \cite{ZhangOriginal}. Its quantifier uses one learner across varying marginals.

\begin{theorem}[Anchored realizable obstruction]\label{thm:anchor}
On $X_d=\{0,\ldots,d\}$ let $H_d^{\rm anc}$ contain all $2^d$ Boolean functions with value zero at $0$. For $t\in B_2^d$ put
\begin{equation}\label{eq:anchor-family}
D_t(0,0)=1-8\eps\|t\|_2^2,\qquad
D_t(i,\ind\{t_i>0\})=8\eps t_i^2.
\end{equation}
If one randomized, possibly improper learner is valid for every $D_t$, where $d\ge1$, $\eps\le1/16$, and $\delta,\rho\le10^{-3}$, then, with $\eta=10\rho+10\delta/9$,
\begin{align}
n&\ge\left(\frac{1-25\eta}{240\eta}\right)^2\frac{d^2}\eps,
\label{eq:anchor-eta}\\
n&\ge\left(\frac{3-400\rho-388\delta}{6208\rho}\right)^2\frac{d^2}\eps
\ge\cstar\frac{d^2}{\eps\rho^2},\qquad
\cstar=\left(\frac{553}{1552000}\right)^2.
\label{eq:anchor-rho}
\end{align}
\end{theorem}
\proofref{sec:anchor-proof} The proof assigns high-probability output cells to each fixed seed. Accurate cells are small, distinct cells are separated, and expanding them forces an uncovered volume charged to replication failure. Keeping complete cells separates the accuracy budget $\delta$ from the stability denominator $\rho$.

The same obstruction gives $N_H^U\ge\cstar(v-1)^2/(\eps\rho^2)$ for every class with $\VC(H)=v\ge2$. A product-marginal version has lower bound $\cstar d^2/(2\eps\rho^2)$ for $\eps\le1/32$. Their precise reductions are proved in \Cref{sec:foundations-reductions}. At $\eps=\delta=\rho=10^{-4}$, the anchored lower bound is $10^{12}\cstar\lfloor\log_2M\rfloor^2$, in particular at least $814\lfloor\log_2M\rfloor^2$.

\subsection{Known marginals: correct an anchor rather than every empirical risk}\label{sec:module-k}
Knowing $p$ makes $d_p(h,g)$ available before observing a second sample. Split the data into two blocks of $N$ observations. Choose the first consistent $g$ from the first block. On the second block compute
\begin{equation}\label{eq:control-estimator}
\widetilde L(h)=d_p(h,g)+\frac1N\sum_{i=1}^N
\left[\ind\{h(X_i)\ne Y_i\}-\ind\{h(X_i)\ne g(X_i)\}\right].
\end{equation}
The bracket is zero whenever $g$ is correct. Give $h$ weight $\exp(-\lambda\widetilde L(h))$, where $\lambda=8\log(2m)/\eps$. Draw a shared radius $R\sim\Unif[\eps/4,\eps/2]$, retain only hypotheses with $d_p(h,g)\le R$, and use the same exponential-clock rule. The retained set contains $g$. Its diameter around the anchor protects accuracy, while the low-variance correction and the shared radius protect replication.

\begin{theorem}[Known-marginal control variates]\label{thm:k}
Let $a=1+\log m$. For $m\ge2$ and $0<\eps,\delta,\rho<1$, the preceding proper learner is valid using $2N$ samples, where
\begin{equation}\label{eq:k-budget}
N=\left\lceil\max\left\{\frac{128a^{3/2}}{\eps\rho},
\frac{2\log(m/\delta)}\eps\right\}\right\rceil.
\end{equation}
At every fixed triple in \eqref{eq:domain},
\[
\lim_{M\to\infty}\frac{\log N^{K,*}(M)}{\log\log M}
=\lim_{M\to\infty}\frac{\log N^{K\text{-unif},*}(M)}{\log\log M}=\frac32.
\]
\end{theorem}
\proofref{sec:proof-k} The exponent assertion combines the upper bound with the finite-grid lower bound below. It is an exponent statement: the displayed upper bound and the general cardinality lower bound need not agree up to a constant factor.

At fixed accuracy and confidence, inverse-replicability exponent one is also optimal among guarantees with polynomial dependence on $\log|H|$. More precisely, no bound $P(\log|H|)\rho^{-b}$ with fixed polynomial $P$ and $b<1$ can hold uniformly, even in interface $K$-unif. \Cref{sec:grid-consequences} proves this using finite threshold chains whose length grows as $1/\rho$.

\subsection{Finite threshold products: retain the final discrete scale}\label{sec:module-grid}
Fix $d,k\ge1$ and $\eps\le1/16$. Let $b=\lfloor(8\eps)^{-1}\rfloor$, $q=1/b$, and take the known uniform marginal on
\[
X=[b]\times[d]\times[k],\qquad
h_j(\ell,i,x)=\ind\{\ell=1\}\ind\{x\le j_i\},
\quad j\in\{0,\ldots,k\}^d.
\]
Thus $|H_{d,k}|=(k+1)^d$ and its active layer has mass $q\in[8\eps,16\eps]\cap(0,1/2]$. The instance depends on $\eps,d,k$, but neither on the sample size nor on the learner.

On the first sample block, set each anchor coordinate to its largest observed positive location, or zero if there is none. Apply the control-variate module on the second block, now with $\lambda=16d/\eps$. This exploits the product geometry rather than paying for $\log|H_{d,k}|$ in the temperature. For the lower bound, draw a uniform target vector, observe its labels, and resample a target from the posterior. This creates two target parameters sharing the same actual sample. Their posterior intervals determine the stability cost.

\begin{theorem}[Exact finite-grid interpolation]\label{thm:grid}
For $d,k\ge1$ and \eqref{eq:domain}, the above fixed class and fixed uniform marginal satisfy
\begin{equation}\label{eq:grid-rate}
N_{H_{d,k},p}^K
=\Theta\left[\frac{dk}{\eps}\log\left(1+\frac{\sqrt d}{k\rho}\right)
+\frac1\eps\log\frac1\delta\right].
\end{equation}
An explicit lower bound is the maximum of
$2^{-26}dk\log(1+\sqrt d/(k\rho))/\eps$ and
$\log(1/(2\delta))/(32\eps)$. A proper upper bound is $2N$, where
\[
N=\left\lceil\max\left\{\frac{4096dk}{q}\log\left(1+\frac{\sqrt d}{k\rho}\right),
\frac4\eps\left(d\log2+\log\frac1\delta\right)\right\}\right\rceil.
\]
\end{theorem}
\proofref{sec:grid-proof} If $k\rho\ge\sqrt d$, the main term is $d^{3/2}/(\eps\rho)$. With $d,k$ fixed and $\rho$ tending to zero, it is logarithmic in $1/\rho$. The discrete posterior comparison in \Cref{lem:grid-kernel} is the technical step that supports both regimes. A complementary continuous-posterior construction and its explicit finite-uniform reduction appear in \Cref{sec:continuous-kernel}.

\subsection{Deterministic learning: construct components with rank-dependent radii}\label{sec:module-det}
For a general known marginal, sort the $m$ distances from each $h$. If $r_h(j)$ is the $j$th distance, choose the largest $k_h$ with $r_h(k_h)\le a k_h$, and connect $h$ to its radius-$ak_h$ ball. Select one fixed representative per connected component. Given a sample, find the first consistent hypothesis and return its component representative. The rank condition makes far-away candidates progressively more expensive to leave consistent; this removes the extra $\log m$ from a uniform-distance union bound.

\begin{theorem}[Deterministic minimax and finite-class saturation]\label{thm:det}
For every known $p$ and $m\ge2$, a deterministic proper learner uses at most
\begin{equation}\label{eq:det-budget}
\left\lceil\frac{4m\log(1+2/\rho)+\log(m/\delta)}\eps\right\rceil
\end{equation}
samples. For $M\ge2$, $\eps\le1/16$, $\delta\le1/4$, and $\rho\le1/8$,
\[
N_{\rm det}^{K,*}(M)\asymp N_{\rm det}^{K\text{-unif},*}(M)
\asymp\frac{M\log(1/\rho)+\log(1/\delta)}\eps.
\]
For unrestricted randomized learners, the same rate holds when
$\delta\le1/16$ and $\rho\le1/(16M^2)$.
\end{theorem}
\proofref{sec:det-proof} The deterministic lower bound uses a threshold chain and an event that must change between its endpoints. The randomized saturation assertion uses the chain interpolation lower bound in \Cref{lem:chain}. Together with \Cref{thm:k,thm:grid}, this gives the general envelopes
\begin{align}
\frac c\eps\left[\mathcal F_M(\rho)+\log\frac1\delta\right]
&\le N^{K\text{-unif},*}(M)\le N^{K,*}(M),\label{eq:envelope-lower}\\
N^{K,*}(M)&\le\frac C\eps\left[
\min\left\{\frac{(1+\log M)^{3/2}}\rho,M\log\frac1\rho\right\}
+\log\frac1\delta\right],\label{eq:envelope-upper}\\
\mathcal F_M(\rho)&=\max_{(k+1)^d\le M}dk\log\left(1+\frac{\sqrt d}{k\rho}\right).
\nonumber
\end{align}
These envelopes are used together on \eqref{eq:domain}. They give a complete characterization in the saturation region of \Cref{thm:det}; matching them throughout the intermediate cardinality--replicability region remains an open problem.

\subsection{VC-sensitive learning: combine an accurate anchor with a small cover}\label{sec:module-vc}
Choose a proper $\eps/256$-cover $F\subseteq H$ in $d_p$, with $q=|F|$. If $q=1$, return its center. Otherwise train a one-inclusion predictor on the first sample block and an empirical-risk minimizer over $F$ on the second. A third, independent block selects the empirically better of these two candidates. This anchor combines small expected risk with a high-probability risk bound. Use the fourth block in \eqref{eq:control-estimator}, with candidate set $F$ and temperature $16\log(2q)/\eps$. If clipping produces an empty set, return a fixed member of $F$.

\begin{theorem}[VC--covering-number bound]\label{thm:vc}
For finite $H$ with $\VC(H)=v$ and a proper cover of size $q\ge2$, the preceding proper learner is valid with $4N$ samples, where
\[
N=\left\lceil\max\left\{\frac{1024(v+\sqrt v\log(2q))}{\eps\rho},
\frac{512\log(6(q+1)/\delta)}\eps\right\}\right\rceil.
\]
A cover can be chosen with $\log q\le2v\log(2048e/\eps)$. Consequently,
\[
N_{H,p}^K=O\left(\frac{v^{3/2}\log(C/\eps)}{\eps\rho}
+\frac{\log(1/\delta)}\eps\right).
\]
For the supremum over all finite classes of VC dimension at most $v\ge1$, the fixed-accuracy rate is
$\Theta_\eps(v^{3/2}/\rho+\log(1/\delta))$ on \eqref{eq:domain}.
\end{theorem}
\proofref{sec:vc-proof} The subscript $\eps$ in the last assertion records fixed accuracy. The preceding all-accuracy upper bound retains its displayed logarithm.

When $v\le1$, the disagreement supports relative to a fixed $h_0$ form a laminar tree. Contract edges shorter than $\eps/(8m)$ for distance calculations. Read the deepest observed disagreement on the target path, then cut the tree at periodically spaced depths using a shared phase. Return a fixed leaf associated with the resulting component. All representatives and leaf projections are fixed before seeing the data.

\begin{theorem}[All cardinality regimes for VC dimension one]\label{thm:vc1}
For $M\ge2$, $\eps\le1/16$, $\delta\le1/16$, and $\rho\le1/8$, the worst-case known-marginal complexity over finite $H$ with $|H|\le M$ and $\VC(H)\le1$ is
\[
\Theta\left[\frac M\eps\log\left(1+\frac1{M\rho}\right)
+\frac{\log(1/\delta)}\eps\right].
\]
The same lower bound holds with a uniform marginal. A proper upper bound for $m=|H|\ge2$ uses
\[
N=\left\lceil\max\left\{\frac{8m}\eps\log\left(1+\frac8{m\rho}\right),
\frac4\eps\log\frac1\delta\right\}\right\rceil
\]
samples and $\lceil\log_2(2/\rho)\rceil$ shared random bits. For explicit finite $H$, finite $X$, and rational $p$, the construction has polynomial total running time in the explicit input length and $N$.
\end{theorem}
\proofref{sec:vc1-proof} The algorithm is a shared-phase procedure on the entire tree, not a separate random choice on each branch. This is what preserves the rate when $H$ has many leaves.
For uniform thresholds on a compactly specified integer interval, \Cref{sec:compact-threshold} implements the same guarantee by scanning the sample and then performing a constant number of rational arithmetic operations. It does not build the explicit tree or enumerate the threshold class. That subsection also gives integer-computable sample budgets within a factor three of the displayed logarithmic budgets.

\subsection{The full Boolean class as a reference point}\label{sec:module-pd}
The constructions in our earlier version \cite{ZhangOriginal} distinguish knowing the masses from merely learning the labels particularly clearly. For $d\ge2$, $\eps\le1/16$, and $\delta,\rho\le10^{-3}$,
\begin{equation}\label{eq:pd-comparison}
N_{P_d}^U=\Theta\left(\frac{d^2}{\eps\rho^2}+\frac{\log(1/\delta)}\eps\right),
\quad
\sup_p N_{P_d,p}^K=\Theta\left(\frac d\eps\log\frac d\rho+\frac{\log(1/\delta)}\eps\right).
\end{equation}
For the known-marginal upper bound, retain coordinates with $p_i\ge\eps/(2d)$, use their observed labels, and set all other coordinates to zero. For the unknown-marginal upper bound, the Poisson--ball algorithm rounds a signed square-root empirical vector using a common ball sampler. Both complete proofs and the matching lower bounds are given in \Cref{sec:foundations-ball,sec:foundations-known-pd}. \Cref{thm:u} supplies a second unknown-marginal construction, and the next section separates its statistical rate from its implementation costs.

\section{Implementation modules and their precise guarantees}\label{sec:computation}
An implementation has three distinct costs: observations, shared random bits, and computation. We count preprocessing and storage when claiming a uniform polynomial-time algorithm. An algorithm polynomial in an explicit list of $m$ hypotheses can still be exponential in the $d$-bit representation of $P_d$.

\subsection{Local blocks: a matching information-dependence trade-off}\label{sec:module-block}
Partition the $d$ coordinates of $P_d$ into $g=\lceil d/s\rceil$ fixed blocks of at most $s$ coordinates. Within each block enumerate its $2^{s_j}$ label vectors and use truncated-power weights with exponent $k_0=4s$ and cutoff $B=\lfloor n\eps/(2g)\rfloor$. Sample each block using shared rejection proposals. Reserve each block's random-tape addresses in advance, quantize the acceptance threshold downwards, and cap its number of proposals. If no proposal is accepted, return a fixed locally consistent vector. Concatenate the block outputs.

\begin{theorem}[Fixed-block trade-off]\label{thm:block}
For $1\le s\le d$ and $0<\eps,\delta,\rho<1$, the block learner has sample complexity
\[
O\left(\frac{d^3}{s\eps\rho^2}+\frac{\log(1/\delta)}\eps\right)
\]
and deterministic running-time and random-bit bounds
$\operatorname{poly}(n,d,\log(1/\rho),\text{parameter bit length})2^{O(s)}$ for rational parameters. For $d\ge2$ and $s\mid d$, this sample rate is optimal on \eqref{eq:domain} among learners whose output on a block can inspect only that block's labeled observations and their positions, together with the public seed and public parameters.
\end{theorem}
\proofref{sec:locality-proof} The lower bound permits arbitrary computation and arbitrary dependence among blocks' shared seeds. Its restriction is access to other blocks' observations. Thus $s=\Theta(\log d)$ yields a polynomial-time algorithm with rate $d^3/(\eps\rho^2\log d)$; the theorem identifies the locality cost of that implementation.

\subsection{Short shared seeds at the optimal statistical rate}\label{sec:module-seed}
Represent the sample by its signed square-root count vector
$x_i=(2b_i-1)\sqrt{c_i/n}$, with zero in unseen coordinates. First consider a common rejection sampler from a radius-$\sqrt{\eps/2}$ ball centered at $x$, and output the signs of the accepted point. Every accepted point respects the empirical error budget. Next restrict the input to the finite set of possible count vectors and select a finite collection of common seeds that approximates all pairwise disagreement probabilities. Store their output functions as a table and choose one row uniformly.

\begin{theorem}[Finite-grid seed compression]\label{thm:seed}
On $P_d$, a proper learner uses
\[
n=\left\lceil\max\left\{\frac{128d^2}{\eps\rho^2},
\frac{8(d\log2+\log(1/\delta))}\eps\right\}\right\rceil
\]
samples and $O(\log d+\log\log(n+1)+\log(1/\rho))$ public bits. The construction is a finite table, of size at most $Kd(2n+1)^d$ bits, with
$K$ the least power of two at least
$16[\log2+2d\log(2n+1)]/\rho$.
For $\eps\le1/16$ and $\delta<1/4$, every unknown-marginal learner on the anchored two-point class with seed support at most $K$ satisfies
$K\ge(1-4\delta)/(2\rho)$, for every finite sample budget.
\end{theorem}
\proofref{sec:randomness-proof} The table size is part of the implementation cost. A second construction in \Cref{sec:seed-average} compresses seeds by controlling the average over two samples from each realizable distribution; it supplies an alternative route to logarithmic shared-bit complexity. \Cref{prop:atomic} gives both the continuous-sphere atomic obstruction and its finite-grid support-size counterpart. For a known-marginal threshold chain, \eqref{eq:chain-seed} separately quantifies how the sample lower bound depends on the number of seed atoms.

\subsection{Shared rejection filtering: reduce learning to ordinary product proposals}\label{sec:module-filter}
Let $L_0=\lceil\log_2m\rceil$, $\lambda_0=32L_0/\eps$, and $a=1-\lambda_0/n$. Use the untruncated Gibbs proposal
\[
P_S(h)\propto a^{C_h(S)}.
\]
At each of $T=\lceil\log_4(32/\rho)\rceil$ independent shared-tape slots, propose a hypothesis with a public sampler. Draw an independent shared midpoint from a $K$-point grid in $(\eps/4,\eps/2)$, where $K$ is the least power of two at least $64/\rho$. Accept if the proposal's empirical loss is below that midpoint. After $T$ unsuccessful trials, output the first consistent hypothesis. Each possible output satisfies the empirical budget, including the fallback.

\begin{theorem}[Public filtering reduction]\label{thm:filter}
Suppose a public proposal sampler, reading only its own input distribution, has marginal TV error at most $\zeta$ and coupled disagreement at most
\[
c\HH(P,Q)+\zeta,\qquad c\ge1,\quad \zeta\le\rho/64.
\]
The filter is valid with
\[
n=\left\lceil\max\left\{\frac{2^{14}(c+1)^2L_0^2}{\eps\rho^2},
\frac{8\log(m/\delta)}\eps\right\}\right\rceil.
\]
Its infinite version makes at most $4/3$ proposals in expectation; its finite version makes at most $T$ proposals. An explicit finite-list implementation has bounded finite-bit running time polynomial in $m,n$, and the parameter bit lengths.
\end{theorem}
\proofref{sec:filter-proof} The analysis charges a possible difference in stopping times once, through a geometric renewal bound. It does not multiply the sample budget by $T$.

For $P_d$, the proposal is a product of Bernoulli coordinates. If coordinate $i$ appears $c_i>0$ times with label $b_i$, its probability of output one is
\[
p_i=\begin{cases}(1+a^{c_i})^{-1},&b_i=1,\\
a^{c_i}/(1+a^{c_i}),&b_i=0,\end{cases}
\]
and $p_i=1/2$ if unseen. A polynomial-time public sampler with the theorem's constant Hellinger distortion on this family would therefore give optimal-statistical-rate polynomial-time learning of $P_d$. This is the remaining computational interface. The reduction removes the global truncation and the knapsack constraint from the proposal distribution.

\subsection{What can be compressed, and what must stay coupled}\label{sec:module-obstructions}
A fixed sufficient statistic can provide an efficient public sampler. If
$P_\theta(v)=\pi_\theta(T(v))Q_{T(v)}(v)$ with a parameter-independent conditional kernel, first couple the statistic and then use a shared conditional seed. For
$P_t(v)\propto t^{|v|}\prod_i a_i^{v_i}$ with fixed $a_i>0$, the statistic $|v|$ has $d+1$ states and its conditional sampler uses an $O(d^2)$ dynamic program. \Cref{prop:lift} proves the exact preservation of TV and Hellinger distances in this construction.

Six further results identify which implementation shortcuts preserve this common coupling and which incur additional costs.
\begin{enumerate}[leftmargin=*,itemsep=.3em]
\item \textbf{Black-box access.} A hypothesis-evaluation oracle requires $\Omega(m)$ queries even without replication, counting preprocessing and prediction queries (\Cref{prop:oracle}).
\item \textbf{Coordinatewise coupling.} Shared coordinatewise uniforms can have mismatch $\Theta(d|\eta|)$ between nearby Bernoulli products whose TV distance is $O(\sqrt d|\eta|)$ (\Cref{sec:normalization}).
\item \textbf{Greedy capacity coupling.} A shared-capacity greedy rule on binary costs has exponentially growing sensitivity to a common rescaling of those costs (\Cref{sec:greedy}).
\item \textbf{A global knapsack objective.} A cardinality-first rule has a $d^4/(\eps\rho^2)$ same-distribution asymptotic obstruction at fixed $d,\eps$. The proof also gives an exact population-boundary perturbation argument and a finite-bit realization (\Cref{prop:knapsack}).
\item \textbf{Sample-dependent compression.} A two-coordinate example has identical cost distributions but coupled output mismatch tending to $1/2$ while TV tends to zero. The example embeds into the actual empirical Gibbs family used by our filter (\Cref{sec:fixed-statistic}).
\item \textbf{Exact fixed statistics for all products.} A parameter-independent conditional kernel for the entire positive Bernoulli product family forces its statistic to be injective, with at least $2^d$ states (\Cref{sec:fixed-statistic}).
\end{enumerate}
These are separate, proved statements about the specified information or coupling models. Alongside the filter, \Cref{sec:rounding-interface} gives an alternative sufficient interface based directly on square-root empirical vectors. The result index in \Cref{sec:result-index} links each construction and obstruction to its detailed argument.

\section{Conclusion}
We resolve the distribution-free finite-class sample-complexity question posed by Larsen et al.\ \cite[Section 1.1]{LMPS}. Our matching bounds determine the joint dependence on class cardinality, accuracy, confidence, and replicability up to universal constants on \eqref{eq:domain}. The quadratic class-size obstruction and inverse-quadratic replication cost established in our earlier version \cite{ZhangOriginal} are attained by the truncated-power learner in \Cref{thm:u}.

Knowing the instance marginal changes the attainable rates. Control variates reduce the principal exponents to $3/2$ in $\log M$ and one in $1/\rho$. Finite threshold products reveal how the latter rate crosses over to logarithmic growth in $1/\rho$ on a fixed finite grid. Rank-adaptive components and a shared-phase tree algorithm give exact deterministic and VC-one characterizations. The implementation results then quantify the costs of locality and shared randomness, and reduce optimal-rate efficient learning of the full Boolean class to a concrete public product-sampling interface.

\clearpage
\appendix
\section{Proof map and result index}\label{sec:proof-map}
The arrows in the next two diagrams record the main proof dependencies, not every supporting inequality. All complete proofs follow. The geometric lemmas and foundational lower bounds come first. Subsequent appendices follow the statistical and implementation modules in the main text.

\begin{figure}[htbp]
\centering
\begin{tikzpicture}[box/.style={draw=black,fill=white,text=black,align=center,font=\small,text width=4.1cm,minimum height=1cm},arr/.style={-{Stealth},draw=black},node distance=8mm and 9mm]
\node[box] (geo) {Compact geometry\\\Cref{sec:geometry}};
\node[box,right=of geo] (del) {Deletion energy and clocks\\\Cref{lem:deletion,lem:clock}};
\node[box,below=of geo] (anc) {Anchored obstruction\\\Cref{thm:anchor}};
\node[box,below=of del] (pow) {Truncated-power sampler\\\Cref{lem:power}};
\node[box,below=of anc] (unknown) {Unknown-marginal optimum\\\Cref{thm:u}};
\node[box,below=of pow] (control) {Control variates and clipping\\\Cref{lem:control,lem:clip}};
\node[box,below=of unknown] (kernel) {Posterior kernel and Gaussian comparison\\\Cref{lem:grid-kernel,lem:gaussian}};
\node[box,below=of control] (known) {Known-marginal upper bound\\\Cref{thm:k}};
\node[box,below=of kernel] (grid) {Exact grid rate; class-size and VC exponents\\\Cref{thm:grid,thm:k,thm:vc}};
\node[box,below=of known] (vc) {Validated one-inclusion anchor\\\Cref{lem:one-inclusion,lem:validation}};
\draw[arr] (geo)--(anc); \draw[arr] (anc)--(unknown);
\draw[arr] (del)--(pow); \draw[arr] (pow)--(unknown);
\draw[arr] (control)--(known); \draw[arr] (kernel)--(grid);
\draw[arr] (control.west) -- ++(-.45,0) |- (grid.north east); \draw[arr] (vc)--(grid);
\end{tikzpicture}
\caption{Main statistical dependencies. The chain lemma separately supplies the small-grid, deterministic, and VC-one lower bounds. Rank balls and laminar shared-phase trees supply their corresponding upper bounds.}
\end{figure}
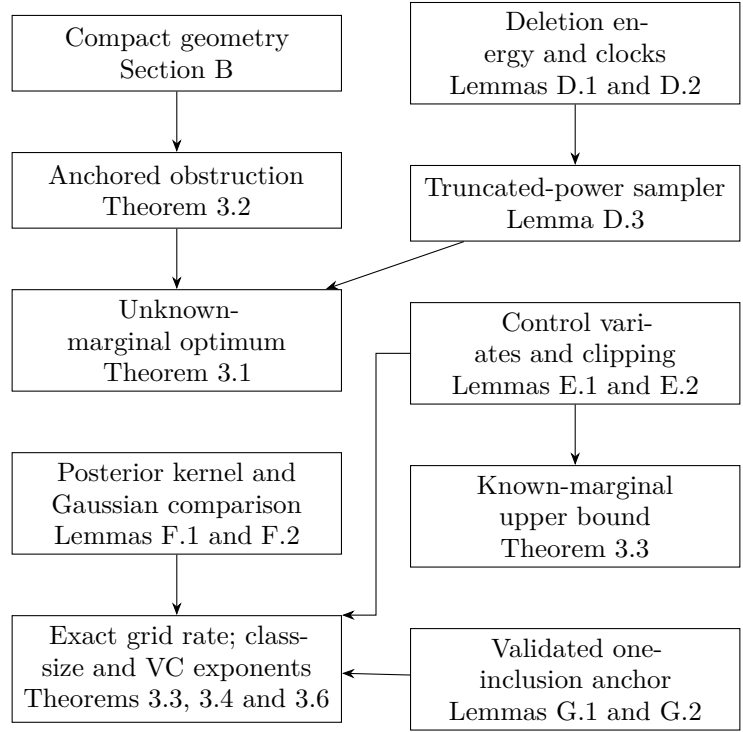

\begin{figure}[htbp]
\centering
\begin{tikzpicture}[box/.style={draw=black,fill=white,text=black,align=center,font=\small,text width=4.1cm,minimum height=1.15cm},arr/.style={-{Stealth},draw=black},node distance=10mm and 9mm]
\node[box] (power) {Truncated-power stability\\\Cref{lem:power}};
\node[box,right=of power] (local) {Local collision + fixed-size covariance\\\Cref{lem:local-collision,lem:block-covariance}};
\node[box,below=of power] (blocks) {Finite-bit block upper bound\\\Cref{thm:block}};
\node[box,below=of local] (lower) {Matching block lower bound\\\Cref{thm:block}};
\node[box,below=of blocks] (ball) {Ball coupling + finite-grid compression\\\Cref{lem:finite-seeds}};
\node[box,below=of lower] (seed) {Short seeds; atomic obstructions\\\Cref{thm:seed,prop:atomic}};
\node[box,below=of ball] (gibbs) {Gibbs deletion + score envelopes\\\Cref{lem:gibbs-displacement,lem:score-envelope}};
\node[box,below=of seed] (filter) {Synchronized rejection filter\\\Cref{thm:filter}};
\draw[arr] (power)--(blocks); \draw[arr] (local)--(lower);
\draw[arr] (ball)--(seed); \draw[arr] (gibbs)--(filter);
\end{tikzpicture}
\caption{Implementation dependencies. The filter reduces optimal full-dimensional computation to a public product sampler. The sufficient-statistic lift and the explicit mechanism counterexamples delimit that interface.}
\end{figure}
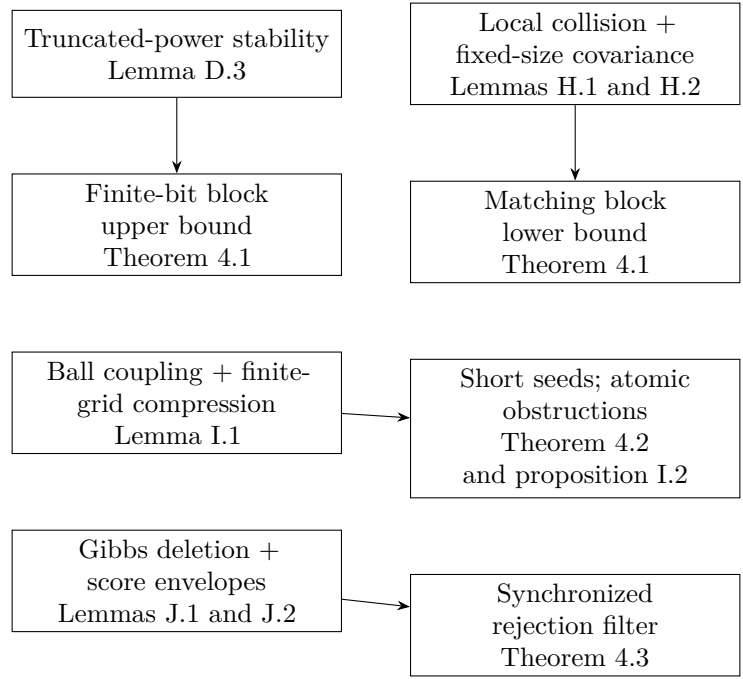

\clearpage
\subsection{Index of results}\label{sec:result-index}
The index includes auxiliary guarantees, alternative constructions, and mechanism-specific lower bounds as well as the main theorems. Each row identifies a mathematical claim and the place where its full argument appears. The first group contains the foundational results from our earlier version \cite{ZhangOriginal}. The remaining groups cover the statistical and implementation results, together with the supporting tools that connect the modules.

\begin{longtable}{@{}>{\raggedright\arraybackslash}p{.20\textwidth}>{\raggedright\arraybackslash}p{.50\textwidth}>{\raggedright\arraybackslash}p{.23\textwidth}@{}}
\toprule
Result & Guarantee and role & Statement or complete argument \\
\midrule\endhead
\multicolumn{3}{@{}l}{\textbf{Foundational results}}\\
Anchored obstruction & Unknown-marginal $d^2/(\eps\rho^2)$ lower bound, with accuracy failure separated from the replication denominator. & \Cref{thm:anchor,sec:anchor-proof} \\
Compact geometry & Self-contained compact Brunn--Minkowski input for output-cell expansion. & \Cref{sec:geometry} \\
Gibbs learner & Finite-class Gibbs upper bound and its KL--generalization analysis. & \Cref{prop:foundations-gibbs,sec:foundations-gibbs} \\
Poisson--ball learner & Full Boolean class upper bound, ball coupling, truncation, and fallback. & \Cref{prop:foundations-ball,sec:foundations-ball} \\
Known full Boolean class & Matching known-marginal rate and its contrast with the unknown-marginal rate. & \Cref{sec:module-pd,sec:foundations-known-pd} \\
Confidence and reductions & Additive confidence lower bound; VC-shattering and product-marginal consequences. & \Cref{sec:foundations-confidence,sec:foundations-reductions} \\
\midrule
\multicolumn{3}{@{}l}{\textbf{Statistical constructions and exact regimes}}\\
Truncated-power learner & Optimal unknown-marginal finite-class rate with a hard empirical budget. & \Cref{thm:u,sec:proof-u} \\
Control variates & Proper known-marginal learner with class-size exponent $3/2$ and inverse-replicability exponent one. & \Cref{thm:k,sec:proof-k} \\
Exponent lower bounds & Uniform-marginal constructions certify both exponents, with their distinct quantifiers. & \Cref{sec:grid-consequences} \\
Finite-grid interpolation & A single exact formula for inverse-$\rho$ growth and fixed-grid logarithmic saturation. & \Cref{thm:grid,sec:grid-proof} \\
Continuous posterior kernel & A second posterior-kernel route, explicit finite-uniform transfer, constants, and cardinality translation. & \Cref{prop:continuous,sec:continuous-kernel} \\
Deterministic components & Rank-adaptive balls remove the extra logarithmic cardinality factor from the deterministic replication term. & \Cref{thm:det,lem:rank,sec:det-proof} \\
Randomized saturation & The same deterministic rate is optimal among randomized learners when $\rho\le1/(16M^2)$. & \Cref{thm:det,sec:det-proof} \\
Threshold-chain bounds & Full chain interpolation, deterministic strengthening, confidence term, and a finite-atom seed refinement. & \Cref{lem:chain}; \eqref{eq:chain-seed} \\
General cardinality envelopes & Explicit lower and upper envelopes throughout the common parameter range. & \eqref{eq:envelope-lower}--\eqref{eq:envelope-upper}; \Cref{sec:grid-consequences} \\
VC--covering bound & Four independent data blocks combine small mean anchor risk, validation, and a proper cover. & \Cref{thm:vc,sec:vc-proof} \\
Fixed-accuracy VC rate & A cardinality-free matching exponent $3/2$ at fixed accuracy. & \Cref{thm:vc}; \eqref{eq:vc-lower} \\
VC-one characterization & Proper shared-phase tree learning with matching bounds in every cardinality--replicability regime. & \Cref{thm:vc1,sec:vc1-proof} \\
Compact thresholds & Implementation without enumerating the class or domain; exact rational choices of sample budgets. & \Cref{sec:compact-threshold} \\
\midrule
\multicolumn{3}{@{}l}{\textbf{Implementation and shared randomness}}\\
Finite-bit block learner & Explicit samples, runtime, random-bit budgets, fixed tape addresses, and bounded rejection. & \Cref{thm:block,sec:locality-proof} \\
Matching locality bound & Local collision and fixed-size covariance yield the matching fixed-block sample rate. & \Cref{lem:local-collision,lem:block-covariance,sec:locality-proof} \\
Pairwise seed compression & Finite empirical grid, logarithmic shared bits, hard error budget, and explicit table storage. & \Cref{thm:seed,lem:finite-seeds} \\
Average seed compression & An independent construction uniform over realizable distributions, using a simplex net rather than all input pairs. & \Cref{sec:seed-average} \\
Two-point seed lower bound & At least $(1-4\delta)/(2\rho)$ seed values in the unknown-marginal two-point experiment. & \Cref{thm:seed,sec:randomness-proof} \\
Continuous and grid atoms & No atomic seed for the continuous zero-additive-error interface; an explicit atom-mass bound on the empirical grid. & \Cref{prop:atomic} \\
Public filtering & Optimal statistical order from Gibbs proposals, hard-budget filtering, and synchronized stopping. & \Cref{thm:filter,sec:filter-proof} \\
Proposal and score bounds & Gibbs deletion energy and data-dependent score control justify the filter's constant expected cost. & \Cref{lem:gibbs-displacement,lem:score-envelope} \\
Bounded-bit proposals & Exact joint-law analysis followed by marginal and coupling error bounds for finite-bit truncation. & \Cref{sec:finite-proposal} \\
Product-sampler reduction & An efficient public sampler for the specified Bernoulli products suffices for optimal-rate efficient $P_d$ learning. & \Cref{sec:module-filter} \\
Square-root interface & An alternative sufficient public rounding map acts directly on signed empirical count vectors. & \Cref{sec:rounding-interface} \\
\midrule
\multicolumn{3}{@{}l}{\textbf{Compression tools and explicit obstructions}}\\
Black-box query bound & Linear dependence on the indexed class size when preprocessing and prediction queries are charged. & \Cref{prop:oracle} \\
Normalizer computation & Exact truncated-power normalizer and marginal sampling by an $O(dn)$ arithmetic dynamic program. & \Cref{sec:normalization} \\
Coordinatewise mismatch & A dimension-dependent loss for the coordinatewise shared-uniform coupling. & \Cref{sec:normalization} \\
Fixed-statistic lift & Exact preservation of TV, Hellinger distance, and the statistic sampler's disagreement probability. & \Cref{prop:lift} \\
One-parameter product family & A fixed-baseline product family has a $d+1$-state statistic and a conditional dynamic program using $O(d^2)$ rational operations. & \Cref{sec:fixed-statistic} \\
Dynamic cost counterexample & Equal cost laws can produce different coupled vectors; the example embeds in the empirical Gibbs family. & \Cref{sec:fixed-statistic} \\
Injectivity obstruction & An exact fixed statistic for every positive Bernoulli product must have at least $2^d$ states. & \Cref{sec:fixed-statistic} \\
Greedy-capacity obstruction & Exact moving-boundary calculation exhibits exponential sensitivity for binary-cost greedy packing. & \Cref{sec:greedy} \\
Knapsack obstruction & Population perturbation, same-distribution $d^4$ sample asymptotic, and a bounded-bit realization for one specified objective. & \Cref{prop:knapsack} \\
\bottomrule
\end{longtable}

\section{The geometric input}\label{sec:geometry}
The packing proofs in \Cref{sec:anchor-proof,sec:locality-proof} use the compact-set Brunn--Minkowski inequality. We give a self-contained derivation. Write $B=B_2^d$ and $\mu(E)=\vol(E)/\vol(B)$. For nonempty compact sets $K,L\subset\mathbb R^d$, the conclusion is $\vol(K+L)^{1/d}\ge\vol(K)^{1/d}+\vol(L)^{1/d}$. If one volume vanishes, a translate of the other set lies in the sum, so only positive volumes need proof.
\subsection{Prékopa--Leindler for bounded compactly supported functions}

Let $0<\theta<1$. Let $f,g:\mathbb R^d\to[0,\infty)$ be bounded Borel functions with compact support, and let $h:\mathbb R^d\to[0,\infty]$ be Borel. We prove
\begin{equation}
\begin{gathered}
h((1-\theta)x+\theta y)
\ge
f(x)^{1-\theta}g(y)^\theta
\quad\text{for all }x,y\\
\Longrightarrow\qquad
\int h
\ge
\left(\int f\right)^{1-\theta}
\left(\int g\right)^\theta.
\end{gathered}
\label{eq:geometry-01}
\end{equation}
If either integral on the right is zero, the conclusion is immediate.

We begin in dimension one.

\subsection{The transport inequality: interval indicators first}

Let $Z:(0,1)\to\mathbb R$ be bounded and strictly increasing. Extend its almost-everywhere derivative $Z'$ by zero on the exceptional set.

First take an interval $I=(a,b)$. The preimage
\[
J=\{u:Z(u)\in(a,b)\}
\]
is an interval (possibly empty or degenerate). For a monotone function, the integral of its derivative over an interval is at most its total increase there. Hence
\[
\int_0^1\mathbf1_{(a,b)}(Z(u))Z'(u)\,du
=
\int_J Z'(u)\,du
\le b-a.
\]
The derivative-versus-increase inequality can also be obtained by applying Fatou's lemma to nonnegative forward difference quotients on compact subintervals of $J$, then approaching its endpoints.

Define the measure
\[
\nu(E)=\int_0^1\mathbf1_E(Z(u))Z'(u)\,du.
\]
The interval calculation shows $\nu(I)\le|I|$. Since an open subset of $\mathbb R$ is a countable disjoint union of intervals, $\nu(O)\le|O|$ for every open $O$. Outer regularity of Lebesgue measure then gives $\nu(E)\le|E|$ for every Borel $E$. Approximation by nonnegative simple functions proves
\begin{equation}
\boxed{
\int_{\mathbb R} h(x)\,dx
\ge
\int_0^1 h(Z(u))Z'(u)\,du
}
\qquad(h\ge0\text{ Borel}).
\label{eq:geometry-02}
\end{equation}
Only an inequality is asserted. Jumps or singular components of $Z$ do not invalidate it; they are precisely why an unjustified equality would be unsafe.

\subsection{Generalized quantiles and the one-dimensional proof}

Write
\[
A=\int_{\mathbb R}f>0,
\qquad
B_0=\int_{\mathbb R}g>0,
\]
and let
\[
F(x)=A^{-1}\int_{-\infty}^x f(s)\,ds,
\qquad
G(x)=B_0^{-1}\int_{-\infty}^x g(s)\,ds.
\]
These distribution functions are continuous. Define their generalized quantiles by
\[
Q_f(u)=\inf\{x:F(x)\ge u\},
\qquad
Q_g(u)=\inf\{x:G(x)\ge u\},
\quad 0<u<1.
\]

Continuity gives
\[
F(Q_f(u))=u,
\qquad
G(Q_g(u))=u.
\]
In particular, the quantiles are strictly increasing, though they may have jumps across gaps in the supports. Moreover,
\[
|\{u:Q_f(u)\le x\}|=F(x),
\]
so $Q_f$ sends uniform measure on $(0,1)$ to density $f/A$. The corresponding statement holds for $Q_g$.

By the Lebesgue differentiation theorem, $F'(x)=f(x)/A$ almost everywhere. Under the pushforward law $f(x)\,dx/A$, almost every $x$ is such a differentiation point and has $0<f(x)<\infty$. At those points the usual inverse-difference-quotient argument is valid: positive derivative makes the inverse continuous locally, and
\[
F(Q_f(u+s))-F(Q_f(u))=s
\]
gives
\begin{equation}
Q_f'(u)=\frac{A}{f(Q_f(u))},
\qquad
Q_g'(u)=\frac{B_0}{g(Q_g(u))}
\quad\text{for almost every }u.
\label{eq:geometry-03}
\end{equation}

Set
\[
Z(u)=(1-\theta)Q_f(u)+\theta Q_g(u).
\]
This is bounded and strictly increasing. Apply \eqref{eq:geometry-02}, the hypothesis in \eqref{eq:geometry-01}, and \eqref{eq:geometry-03}:
\begin{equation}
\begin{aligned}
\int h
&\ge
\int_0^1
h(Z(u))
\left((1-\theta)Q_f'(u)+\theta Q_g'(u)\right)\,du\\
&\ge
\int_0^1
f(Q_f(u))^{1-\theta}g(Q_g(u))^\theta\\
&\hspace{2.6cm}\cdot
\left(
(1-\theta)\frac{A}{f(Q_f(u))}
+
\theta\frac{B_0}{g(Q_g(u))}
\right)\,du\\
&\ge
A^{1-\theta}B_0^\theta.
\end{aligned}
\label{eq:geometry-04}
\end{equation}
The last step is weighted arithmetic--geometric mean, applied where the two density values are finite and positive; the exceptional set has measure zero. This elementary inequality follows, for example, from $\log x\le x-1$.

This proves \eqref{eq:geometry-01} in dimension one without assuming that the densities are everywhere positive or that the quantiles are absolutely continuous.

\subsection{Induction on dimension}

Assume \eqref{eq:geometry-01} holds in dimension $d-1$. Write points as $(x',s)$, and define
\[
F_1(x')=\int_{\mathbb R}f(x',s)\,ds,
\quad
G_1(y')=\int_{\mathbb R}g(y',s)\,ds,
\quad
H_1(z')=\int_{\mathbb R}h(z',s)\,ds.
\]
The first two functions are bounded Borel functions with compact support.

For fixed $x',y'$, apply the one-dimensional result to the corresponding slices and to
$z'=(1-\theta)x'+\theta y'$. The pointwise hypothesis in \eqref{eq:geometry-01} gives
\[
H_1((1-\theta)x'+\theta y')
\ge
F_1(x')^{1-\theta}G_1(y')^\theta.
\]
Slices of zero integral cause no difficulty, since the right side is then zero.

Apply the $(d-1)$-dimensional result to $F_1,G_1,H_1$, and use Tonelli. This proves \eqref{eq:geometry-01} in dimension $d$.

\subsection{From Prékopa--Leindler to Brunn--Minkowski}

For compact $K,L$, apply \eqref{eq:geometry-01} to
\[
f=\mathbf1_K,\qquad
g=\mathbf1_L,\qquad
h=\mathbf1_{(1-\theta)K+\theta L}.
\]
The Minkowski combination is compact, so its indicator is Borel. We obtain
\[
\operatorname{vol}((1-\theta)K+\theta L)
\ge
\operatorname{vol}(K)^{1-\theta}
\operatorname{vol}(L)^\theta.
\]

For positive volumes, let
\[
\alpha=\operatorname{vol}(K)^{1/d},
\qquad
\beta=\operatorname{vol}(L)^{1/d},
\qquad
\theta=\frac{\beta}{\alpha+\beta}.
\]
The scaled sets $K/\alpha$ and $L/\beta$ have volume one, and
\[
(1-\theta)\frac K\alpha+\theta\frac L\beta
=
\frac{K+L}{\alpha+\beta}.
\]
The preceding inequality gives volume at least one for this set. Rescaling proves
\begin{equation}
\operatorname{vol}(K+L)
\ge(\alpha+\beta)^d.
\label{eq:geometry-05}
\end{equation}

In particular, for a nonempty compact set $K\subseteq\mathbb R^d$ and $b>0$,
\[
\mu(K+bB)
\ge
\left(\mu(K)^{1/d}+b\right)^d,
\]
which is exactly the compact-set inequality used in \Cref{lem:cell-packing}.~$\square$

\section{Unknown-marginal foundations}\label{sec:foundations}
This appendix gives complete proofs of the foundational results from our earlier version \cite{ZhangOriginal}.
\subsection{Geometric notation and the realizable family}

Let
\[
B=B_2^d,\qquad V_d=\operatorname{vol}(B),\qquad V_0=1.
\]
For measurable $E\subseteq\mathbb R^d$, write
\[
\mu(E)=\frac{\operatorname{vol}(E)}{V_d}.
\]
The restriction of $\mu$ to $B$ is a probability measure.

On
\[
X_d=\{0,1,\ldots,d\},
\]
define the anchored class
\[
H_d^{\mathrm{anc}}=\{h_v:v\in\{0,1\}^d\},
\qquad
h_v(0)=0,\quad h_v(i)=v_i.
\]
Its cardinality is $2^d$. After relabeling the domain,
$H_{d-1}^{\mathrm{anc}}$ is a subclass of $P_d$.

For $0<q\le1/2$ and $t\in B$, define
\begin{equation}
\begin{aligned}
D_t^{(q)}(0,0)&=1-q\|t\|^2,\\
D_t^{(q)}\!\left(i,\mathbf 1\{t_i>0\}\right)&=qt_i^2,
\qquad 1\le i\le d.
\end{aligned}
\label{eq:foundations-01}
\end{equation}
All unlisted atoms have probability zero. This distribution is realizable by $H_d^{\mathrm{anc}}$, including when some coordinates vanish.

For $0<\varepsilon\le1/16$, set
\[
D_t^\varepsilon=D_t^{(8\varepsilon)}.
\]

For distributions $P,Q$ on a finite space, use
\[
\mathsf H(P,Q)^2
=\sum_z(\sqrt{P(z)}-\sqrt{Q(z)})^2
=2(1-b),
\qquad
b=\sum_z\sqrt{P(z)Q(z)},
\]
and
\[
\operatorname{TV}(P,Q)=\frac12\sum_z|P(z)-Q(z)|.
\]
For distributions on $\mathbb R^d$ with Lebesgue densities $f,g$ we use the same definition with $\frac12\int|f-g|$; for uniform laws on two sets $E,E'$ of equal positive volume this equals $\operatorname{vol}(E\setminus E')/\operatorname{vol}(E)$. We use the convention $0\log0=0$.

\subsection{The parameterized lower bound}\label{sec:anchor-proof}
This subsection proves Theorem~\ref{thm:anchor}, the lower-bound module in Section~\ref{sec:module-u}. The four lemmas supply, in order, input continuity, small accurate regions, separated output cells, and a packing inequality. The geometric input was proved in Appendix~\ref{sec:geometry}.

We first prove four lemmas.

\begin{lemma}[Lipschitz control of the input experiment]\label{lem:input-continuity}
For every integer $d\ge1$, every $0<q\le1/2$, every $s,t\in B$, and every integer $n\ge1$,
\[
\mathsf H(D_t^{(q)},D_s^{(q)})^2
\le2q\|t-s\|^2,
\]
and
\begin{equation}
\operatorname{TV}\!\left((D_t^{(q)})^n,(D_s^{(q)})^n\right)
\le\sqrt{2qn}\,\|t-s\|.
\label{eq:foundations-02}
\end{equation}
In particular,
\[
\operatorname{TV}\!\left((D_t^\varepsilon)^n,(D_s^\varepsilon)^n\right)
\le4\sqrt{\varepsilon n}\,\|t-s\|.
\]
\end{lemma}

\textbf{Proof.}
For coordinate $i$, the square-root probabilities of its two possible labeled atoms are
\[
\sqrt q\,((-t_i)_+,(t_i)_+).
\]
The map
\[
u\longmapsto((-u)_+,u_+)
\]
is $1$-Lipschitz. For inputs of the same sign this is immediate. For inputs of opposite signs, the squared image distance is $u^2+v^2$, which is at most $(u-v)^2$. At zero the label changes, but the corresponding probability is zero.

The anchor square-root probability is
\[
F(t)=\sqrt{1-q\|t\|^2}.
\]
On $B$,
\[
\|\nabla F(t)\|
=
\frac{q\|t\|}{\sqrt{1-q\|t\|^2}}
\le
\frac q{\sqrt{1-q}}
\le\sqrt q.
\]
Integrating along the segment from $s$ to $t$, which stays in the convex ball $B$, bounds the squared anchor contribution by $q\|t-s\|^2$. Summing all contributions proves the Hellinger bound.

For completeness, if $b=\sum_z\sqrt{P(z)Q(z)}$, then
\begin{equation}
\begin{aligned}
\mathsf H(P^n,Q^n)^2
&=2(1-b^n)
\le n\mathsf H(P,Q)^2,\\
\operatorname{TV}(P,Q)
&\le\frac12\mathsf H(P,Q)\sqrt{2+2b}
\le\mathsf H(P,Q).
\end{aligned}
\label{eq:foundations-03}
\end{equation}
The first line follows from multiplicativity of affinity and
$1-b^n\le n(1-b)$. The second is Cauchy--Schwarz applied to
$|P-Q|=|\sqrt P-\sqrt Q|(\sqrt P+\sqrt Q)$.
Combining these inequalities proves \eqref{eq:foundations-02}.~$\square$

For any fixed seed, an output event is a deterministic event of the input sample. Total variation cannot increase under this map: its probability difference is bounded by the supremum of input-event probability differences. Thus the same Lipschitz constant applies to every fixed-seed output probability.

\begin{lemma}[Volume of an accurate-output region]\label{lem:accurate-region-volume}
For every integer $d\ge1$, every $0<\varepsilon\le1/16$, and every Boolean function $f$ on $X_d$,
\begin{equation}
\mu\{t\in B:L_{D_t^\varepsilon}(f)\le\varepsilon\}
\le
\left(\frac{24}{25}\right)^d.
\label{eq:foundations-04}
\end{equation}
The accurate-output region is compact.
\end{lemma}

\textbf{Proof.}
If $f(0)=1$, its loss is at least the anchor mass, hence at least $1/2>\varepsilon$. The region is then empty.

Suppose $f(0)=0$. Reflecting coordinates according to $f(i)$, and using invariance of the ball under these reflections, reduces accuracy to
\[
\sum_{i:\,t_i<0}t_i^2\le\frac18.
\]

Let $G=(G_1,\ldots,G_d)$ have independent standard Gaussian coordinates, and let $R$ be independent with
$\Pr(R\le r)=r^d$, $0\le r\le1$. Then
\[
T=R\frac G{\|G\|}
\]
is uniform on $B$. Indeed, Gaussian rotational invariance gives a uniform direction, and the radial distribution of uniform volume on $B$ is $r^d$. The event $G=0$ has probability zero.

Put
\[
Z=\sum_{G_i<0}G_i^2,
\qquad
W=\|G\|^2.
\]
If $R>3/4$ and $T$ is accurate, then $Z\le2W/9$. Exponential Markov with parameter $1/2$, followed by independence of the coordinates, gives
\begin{equation}
\begin{aligned}
\Pr[T\text{ is accurate}]
&\le
\left(\frac34\right)^d
+
\mathbb E\exp\!\left(\frac12\left(\frac{2W}{9}-Z\right)\right)\\
&=
\left(\frac34\right)^d
+
\left[
\frac12(7/9)^{-1/2}
+
\frac12(16/9)^{-1/2}
\right]^d\\
&=
\left(\frac34\right)^d
+
\left(\frac3{2\sqrt7}+\frac38\right)^d\\
&\le
\left(\frac34\right)^d
+
\left(\frac{17}{18}\right)^d.
\end{aligned}
\label{eq:foundations-05}
\end{equation}
Here the Gaussian integral is obtained by rescaling the standard Gaussian density. The last inequality follows from
\[
\frac{17}{18}-\frac38=\frac{41}{72},
\qquad
\left(\frac{41}{72}\right)^2-\frac9{28}
=\frac{103}{36288}>0.
\]

For $d\ge9$, divide the final expression in \eqref{eq:foundations-05} by $(24/25)^d$. The resulting ratio is at most
\[
\left(\frac{25}{32}\right)^9
+
\left(\frac{425}{432}\right)^9.
\]
Now
\[
\left(\frac{25}{32}\right)^3
=\frac{15625}{32768}<\frac12,
\]
and
\[
\left(\frac{432}{425}\right)^9
=
\left(1+\frac7{425}\right)^9
\ge
1+\frac{63}{425}
=
\frac{488}{425}.
\]
Therefore
\[
\left(\frac{25}{32}\right)^9
+
\left(\frac{425}{432}\right)^9
<
\frac18+\frac{425}{488}
=
\frac{243}{244}<1.
\]

It remains to handle $d\le8$. For $f(0)=0$,
\[
L_{D_t^\varepsilon}(f)+L_{D_{-t}^\varepsilon}(f)
=
8\varepsilon\|t\|^2.
\]
If both $t$ and $-t$ are accurate, then $\|t\|\le1/2$. If $E$ denotes the accurate region, symmetry gives
\begin{equation}
2\mu(E)
=
\mu(E\cup(-E))+\mu(E\cap(-E))
\le1+2^{-d}.
\label{eq:foundations-06}
\end{equation}
For $d=1$, this yields $3/4<24/25$. For $2\le d\le8$,
\[
\mu(E)
\le\frac58
<
\frac{17}{25}
\le
\left(\frac{24}{25}\right)^8
\le
\left(\frac{24}{25}\right)^d.
\]
The nontrivial rational comparison is
\[
\left(\frac{24}{25}\right)^8-\frac{17}{25}
=
\frac{6315548551}{152587890625}>0.
\]

Finally, the loss is continuous in $t$. Each coordinate contribution is a squared positive or negative part, and the anchor contribution is continuous. Thus $E$ is closed in compact $B$, and hence compact.~$\square$

\begin{lemma}[High-probability output cells]\label{lem:output-cells}
Fix a finite-output experiment indexed by $t\in B$, a learner satisfying \eqref{eq:model} for every parameter, and a threshold $u\in(1/2,1)$. For a seed $r$, define
\[
p_f(t,r)=\Pr_S[A(S;r)=f],
\qquad
p_{\max}(t,r)=\max_f p_f(t,r),
\]
and
\[
C_f(r)=\{t\in B:p_f(t,r)\ge u\}.
\]

Suppose every fixed-seed output-event probability is $L$-Lipschitz in $t$, where $L>0$. Then the cells are compact and pairwise disjoint, and distinct cells have distance at least
\[
\frac{2u-1}{L}.
\]
For the family $D_t^\varepsilon$, define
\[
H_r=\mu\{t:p_{\max}(t,r)<u\},
\]
and
\[
I_r
=
\sum_f
\mu\!\left(
C_f(r)\cap\{t:L_{D_t^\varepsilon}(f)>\varepsilon\}
\right).
\]
Then
\begin{equation}
\mathbb E_r H_r\le\frac{\rho}{1-u},
\qquad
\mathbb E_r I_r\le\frac{\delta}{u}.
\label{eq:foundations-07}
\end{equation}
\end{lemma}

\textbf{Proof.}
In the anchored experiment there are at most $2^{d+1}$ Boolean output functions on $X_d$, including improper outputs. The same argument applies to any finite output space. Lipschitz continuity makes each cell closed in $B$, hence compact. Two output probabilities cannot both exceed $1/2$, so the cells are disjoint.

If $t\in C_f(r)$, $s\in C_g(r)$, and $f\ne g$, then
\[
p_f(t,r)\ge u,
\qquad
p_f(s,r)\le1-u.
\]
Therefore
\[
L\|t-s\|\ge2u-1.
\]

For a fixed $t,r$, the collision probability of two independent output draws is $\sum_f p_f(t,r)^2$. Since
\[
\sum_f p_f(t,r)^2\le p_{\max}(t,r),
\]
replicability implies
\[
\mathbb E_r[1-p_{\max}(t,r)]\le\rho.
\]
On $\{p_{\max}<u\}$, one has $1-p_{\max}>1-u$. Since $p_f(t,r)$ is continuous in $t$ and measurable in $r$ (Section~\ref{sec:model}), it is jointly measurable, so expectations over $r$ and integrals over $t$ may be interchanged by Tonelli's theorem. Integrating over $t$ proves the first inequality in \eqref{eq:foundations-07}.

If $t\in C_f(r)$ and $f$ is inaccurate at $t$, then the fixed-seed probability $\Pr_S[L_{D_t^\varepsilon}(A(S;r))>\varepsilon]$ is at least $p_f(t,r)\ge u$. More precisely, disjointness of the cells gives
\[
u\sum_f
\mathbf1\{t\in C_f(r),\ L_{D_t^\varepsilon}(f)>\varepsilon\}
\le
\Pr_S[L_{D_t^\varepsilon}(A(S;r))>\varepsilon].
\]
Integrating over $t,r$ and using PAC accuracy proves the second inequality. Both conclusions are averages over seeds, not pointwise assertions about each seed.~$\square$

\begin{lemma}[Packing with exceptional large cells]\label{lem:cell-packing}
Let $d\ge1$, $a>0$, and $0<w\le1$. Let a finite collection of compact subsets $C_f\subseteq B$ be pairwise separated by distance at least $2a$. Let $h=1-\sum_f\mu(C_f)$ be the uncovered normalized volume (the cells are disjoint because $a>0$). Let $s$ be the total normalized volume of cells satisfying
\[
\mu(C_f)\le w^d.
\]
Then
\begin{equation}
h\ge
s\left[
\left(\frac{1+a/w}{1+a}\right)^d-1
\right]
\ge
s\,\frac{da(w^{-1}-1)}{1+a}.
\label{eq:foundations-08}
\end{equation}
\end{lemma}

\textbf{Proof.}
Zero-volume cells do not affect the volume sums and may be omitted. For every remaining cell, set
\[
U_f=C_f+aB^\circ.
\]
These open neighborhoods are disjoint: membership in two of them would give two cell points at distance strictly less than $2a$. They are contained in $(1+a)B$.

For $0<b<a$, the compact-set Brunn--Minkowski inequality \cite{Gardner} (proved in Appendix~\ref{sec:geometry}) yields
\[
\mu(C_f+bB)
\ge
\left(\mu(C_f)^{1/d}+b\right)^d.
\]
As $b\uparrow a$, the sets $C_f+bB$ increase to $U_f$. Continuity from below therefore gives
\[
\mu(U_f)
\ge
\left(\mu(C_f)^{1/d}+a\right)^d.
\]

Every cell has normalized volume at most one, so its expansion factor is at least $(1+a)^d$. A cell with volume at most $w^d$ has expansion factor at least $(1+a/w)^d$. Consequently,
\[
(1+a)^d
\ge
(1-h-s)(1+a)^d+s(1+a/w)^d.
\]
Rearranging proves the first inequality in \eqref{eq:foundations-08}. The second follows from
\[
\frac{1+a/w}{1+a}
=
1+\frac{a(w^{-1}-1)}{1+a}
\]
and Bernoulli's inequality.~$\square$

\paragraph{Proof of Theorem~\ref{thm:anchor}: exclusion of $n=0$.}

A zero-sample output depends only on the seed. \Cref{lem:accurate-region-volume} implies that its average accuracy probability, averaged over uniform $t\in B$ and the seed, is at most
\[
(24/25)^d\le24/25<1-\delta.
\]
This contradicts accuracy for every $t$. Hence $n\ge1$.

\paragraph{Proof of \eqref{eq:anchor-eta}.}

Choose $u=9/10$. Intersect each cell with its accurate region:
\[
K_f(r)=C_f(r)\cap\{t:L_{D_t^\varepsilon}(f)\le\varepsilon\}.
\]
These are compact, and their total normalized volume is $1-H_r-I_r$. By \Cref{lem:output-cells} and Fubini's theorem (joint measurability of $p_f$ was noted in the proof of \Cref{lem:output-cells}), some seed $r_*$ satisfies
\[
\sum_f\mu(K_f(r_*))
\ge
1-10\rho-\frac{10}{9}\delta
=
1-\eta.
\]

Each $K_f(r_*)$ has volume at most $(24/25)^d$. By \Cref{lem:input-continuity,lem:output-cells}, distinct cells are separated by at least
\[
\frac{4/5}{4\sqrt{\varepsilon n}}
=
\frac1{5\sqrt{\varepsilon n}}
=
2a,
\qquad
a=\frac1{10\sqrt{\varepsilon n}}.
\]
Applying the neighborhood argument from \Cref{lem:cell-packing} to the accurate cells gives
\begin{equation}
(1-\eta)\left(1+\frac{25a}{24}\right)^d
\le
(1+a)^d.
\label{eq:foundations-09}
\end{equation}
Dividing by $(1+a)^d$ and using Bernoulli,
\[
\frac{d(1-\eta)}
{24(10\sqrt{\varepsilon n}+1)}
\le\eta.
\]
Thus
\begin{equation}
\begin{aligned}
240\eta\sqrt{\varepsilon n}
&\ge d(1-\eta)-24\eta\\
&\ge d(1-25\eta).
\end{aligned}
\label{eq:foundations-10}
\end{equation}
The last step uses $d\ge1$. No upper bound on $a$ has been assumed.

The parameter range gives
\[
\eta\le\frac1{90},
\qquad
1-25\eta\ge\frac{13}{18}.
\]
Squaring \eqref{eq:foundations-10} proves \eqref{eq:anchor-eta}, including
\[
\left(\frac{13/18}{240}\right)^2
=
\frac{169}{18662400}.
\]

\paragraph{Proof of \eqref{eq:anchor-rho}.}

Now choose
\[
u=\frac34,
\qquad
w=\frac{97}{100},
\qquad
a=\frac1{16\sqrt{\varepsilon n}}.
\]
The cell separation is at least
\[
\frac{1/2}{4\sqrt{\varepsilon n}}=2a.
\]
\Cref{lem:output-cells} gives
\[
\mathbb E H_r\le4\rho,
\qquad
\mathbb E I_r\le\frac43\delta.
\]

Call a cell large if its normalized volume exceeds $w^d$. By \Cref{lem:accurate-region-volume}, the accurate fraction of a large cell is at most
\[
\left(\frac{96}{97}\right)^d
\le\frac{96}{97}.
\]
Thus at least $1/97$ of its volume is inaccurate. The total volume of large cells is therefore at most $97I_r$.

Let $s_r$ be the total volume of the remaining small cells. Then
\begin{equation}
\mathbb E s_r
\ge
1-4\rho-\frac{388}{3}\delta
=:s_0
\ge\frac{13}{15}.
\label{eq:foundations-11}
\end{equation}
Apply \Cref{lem:cell-packing} separately for each seed and then take expectations. Since $w^{-1}-1=3/97$,
\[
4\rho
\ge
\frac{3s_0d}{97(16\sqrt{\varepsilon n}+1)}.
\]
It follows that
\begin{equation}
\begin{aligned}
6208\rho\sqrt{\varepsilon n}
&\ge3s_0d-388\rho\\
&\ge d(3-400\rho-388\delta)\\
&\ge\frac{553}{250}d.
\end{aligned}
\label{eq:foundations-12}
\end{equation}
The middle line uses $d\ge1$, and the last uses
$\delta,\rho\le10^{-3}$.

Squaring proves \eqref{eq:anchor-rho}, because
\[
\left(\frac{553/250}{6208}\right)^2
=
\left(\frac{553}{1552000}\right)^2
=
c_\star.
\]
This completes the proof of Theorem~\ref{thm:anchor}.~$\square$

No relation between $n$ and $d$ was assumed. In fact the bound itself yields $n\ge c_\star d^2/(\varepsilon\rho^2)>2d^2\ge d$, because
\[
\frac{c_\star}{\varepsilon\rho^2}
\ge
\frac{305809}{150544}>2.
\]

\subsection{A finite-class Gibbs upper bound}\label{sec:foundations-gibbs}
This Gibbs upper bound provides a baseline for the filtering module in \Cref{sec:module-filter}. Its confidence dependence is improved by \Cref{thm:u}.
\begin{proposition}[Gibbs baseline]\label{prop:foundations-gibbs}
For finite $H$ of size $m\ge2$ and $0<\eps,\delta,\rho<1$, a proper unknown-marginal learner uses
\begin{equation}\label{eq:foundations-gibbs}
n=\left\lceil\frac{32\log m\log(m/\delta)}{\eps\rho^2}\right\rceil.
\end{equation}
\end{proposition}

\begin{lemma}[Public rejection-sampling coupling]\label{lem:rejection-coupling}
Let $P,Q$ be probability distributions on a finite set $H$ of size $m$. Use a common sequence of independent proposals
\[
(h_j,u_j),\qquad
h_j\sim\operatorname{Unif}(H),\quad
u_j\sim\operatorname{Unif}[0,1].
\]
The $P$-sampler accepts the first proposal with $u_j\le P(h_j)$, and the $Q$-sampler uses $u_j\le Q(h_j)$. If no proposal is ever accepted (a null event), the sampler outputs a fixed default element of $H$; for the ball sampler of Proposition~\ref{prop:foundations-ball} the default is the point $Y=0$, so that the output is the all-zero function.

Each output has the required marginal distribution, and
\begin{equation}
\Pr[\text{the outputs differ}]
\le
\frac{2\operatorname{TV}(P,Q)}
{1+\operatorname{TV}(P,Q)}
\le
2\operatorname{TV}(P,Q).
\label{eq:foundations-13}
\end{equation}

The same guarantee applies to uniform distributions on equal-positive-measure sets contained in a common finite-measure proposal space.
\end{lemma}

\textbf{Proof.}
The probability that one proposal is accepted by the $P$-sampler and has value $h$ is $P(h)/m$. The total acceptance probability is $1/m$, so the first accepted value has law $P$. The sampler terminates almost surely. The same holds for $Q$.

Write $v=\operatorname{TV}(P,Q)$. For one proposal, the acceptance intersection and union have probabilities
\[
\frac{1-v}{m}
\quad\text{and}\quad
\frac{1+v}{m},
\]
respectively. The first proposal in the union lies in the intersection with probability
\[
\frac{1-v}{1+v}.
\]
On this event, both samplers accept the same proposal. Therefore disagreement is at most $2v/(1+v)$.

This is an upper bound, not an equality: even after a proposal accepted by only one sampler, the eventual output values can still coincide.

For equal-measure sets, replace the two acceptance regions in $H\times[0,1]$ by the two sets themselves. The same intersection-over-union calculation applies.~$\square$

\paragraph{Proof of Proposition~\ref{prop:foundations-gibbs}.}

Take $n$ as in \eqref{eq:foundations-gibbs}, and define
\begin{equation}
\lambda=\frac{2\log(m/\delta)}{\varepsilon},
\qquad
\gamma=\frac{\lambda}{n},
\qquad
P_S(h)=
\frac{\exp(-\gamma\sum_{z\in S}\ell_h(z))}{Z_S},
\label{eq:foundations-14}
\end{equation}
where $\ell_h(x,y)=\mathbf1\{h(x)\ne y\}$ and $Z_S$ is the sum of the unnormalized weights. Output a sample from $P_S$ using \Cref{lem:rejection-coupling} with public randomness.

The sample-size choice ensures
\[
\gamma\le\frac{\rho^2}{16\log m}\le\frac12.
\]
The learner is proper and does not use the unknown distribution.

\textbf{Accuracy.}
Realizability supplies a hypothesis with zero sample loss almost surely, so $Z_S\ge1$. For any $h$ with $L_D(h)>\varepsilon$,
\[
\mathbb E_S\exp\!\left(-\gamma\sum_{z\in S}\ell_h(z)\right)
=
\left(1-L_D(h)(1-e^{-\gamma})\right)^n.
\]
Consequently,
\begin{equation}
\begin{aligned}
\mathbb E_S P_S\{h:L_D(h)>\varepsilon\}
&\le
m\exp[-n\varepsilon(1-e^{-\gamma})]\\
&\le
me^{-\lambda\varepsilon/2}
=\delta,
\end{aligned}
\label{eq:foundations-15}
\end{equation}
using $1-e^{-\gamma}\ge\gamma/2$ for $0\le\gamma\le1/2$.

\textbf{An empirical-risk bound.}
Let $\widehat L_S(h)=\frac1n\sum_{z\in S}\ell_h(z)$, so that $P_S(h)=e^{-\lambda\widehat L_S(h)}/Z_S$ by \eqref{eq:foundations-14} and $\lambda=n\gamma$. Let $\widehat L_S(P)=\sum_hP(h)\widehat L_S(h)$, and similarly define $L_D(P)$. The Gibbs entropy identity is
\[
\lambda\widehat L_S(P_S)+\log Z_S
=
\mathsf{Ent}(P_S)
\le\log m.
\]
Since $Z_S\ge1$,
\[
\widehat L_S(P_S)\le\frac{\log m}{\lambda}.
\]

\textbf{Leave-one-out control with the same $\gamma$.}
Let $T$ contain $n-1$ examples, and define $Q_T$ using the same coefficient $\gamma$ as in \eqref{eq:foundations-14}, not $\lambda/(n-1)$. For an additional example $z$, let $P^z=P_{T\cup\{z\}}$. Then
\[
\frac{P^z(h)}{Q_T(h)}
=
\frac{e^{-\gamma\ell_h(z)}}
{Q_T(e^{-\gamma\ell(z)})}
\in[e^{-\gamma},e^\gamma].
\]
Thus, for every $z,z'$,
\[
P^{z'}\ell(z)\le e^{2\gamma}P^z\ell(z).
\]
Average over independent $T,z,z'$. On the left, $z$ is a fresh example for the sample $T\cup\{z'\}$; on the right, exchangeability identifies the expected loss on the distinguished training example with expected empirical risk. Therefore
\begin{equation}
\mathbb E_S[L_D(P_S)-\widehat L_S(P_S)]
\le
(e^{2\gamma}-1)\,
\mathbb E_S\widehat L_S(P_S).
\label{eq:foundations-16}
\end{equation}

\textbf{The KL identity.}
For independent samples $S,S'$,
\[
\operatorname{KL}(P_S\|P_{S'})
=
\lambda P_S(\widehat L_{S'}-\widehat L_S)
+\log Z_{S'}-\log Z_S.
\]
Taking expectations, the two log-partition terms cancel. Independence turns the cross-sample risk into population risk. Hence
\begin{equation}
\begin{aligned}
\mathbb E_{S,S'}\operatorname{KL}(P_S\|P_{S'})
&=
\lambda\,
\mathbb E_S[L_D(P_S)-\widehat L_S(P_S)]\\
&\le
(e^{2\gamma}-1)\log m\\
&\le4\gamma\log m.
\end{aligned}
\label{eq:foundations-17}
\end{equation}
The last inequality holds for $\gamma\le1/2$. For example, convexity on $[0,1]$ gives $e^x-1\le(e-1)x\le2x$, and one may take $x=2\gamma$.

This is an expected KL--generalization identity of the type studied in \cite{WLF}.

\textbf{Replicability.}
For affinity $b$, Jensen's inequality gives
\[
\mathsf H(P,Q)^2
=
2(1-b)
\le-2\log b
\le\operatorname{KL}(P\|Q).
\]
Indeed,
\[
\log b
=
\log\mathbb E_P\sqrt{Q/P}
\ge-\tfrac12\operatorname{KL}(P\|Q).
\]
Together with $\operatorname{TV}\le\mathsf H$, \Cref{lem:rejection-coupling}, and Jensen,
\begin{equation}
\begin{aligned}
\Pr[A(S;r)\ne A(S';r)]
&\le2\mathbb E_{S,S'}\operatorname{TV}(P_S,P_{S'})\\
&\le2\sqrt{\mathbb E_{S,S'}\operatorname{KL}(P_S\|P_{S'})}\\
&\le4\sqrt{\frac{\lambda\log m}{n}}
\le\rho.
\end{aligned}
\label{eq:foundations-18}
\end{equation}
This proves Proposition~\ref{prop:foundations-gibbs}.~$\square$

\subsection{The Poisson--ball upper bound for \texorpdfstring{$P_d$}{the full Boolean class}}\label{sec:foundations-ball}
This subsection proves the guarantee for the geometric algorithm in \Cref{sec:module-pd}. The short-seed construction in \Cref{sec:module-seed} will use a fixed-sample variant of the same ball-coupling estimate.
\begin{proposition}[Poisson--ball learner]\label{prop:foundations-ball}
For $d\ge1$, $0<\eps\le1$, and $0<\delta,\rho<1$, a proper learner for $P_d$ uses
\begin{equation}\label{eq:foundations-ball}
n=\left\lceil\max\left\{\frac{128d^2}{\eps\rho^2},
\frac{16(d\log3+\log(2/\delta))}{\eps}\right\}\right\rceil.
\end{equation}
\end{proposition}

We provide the Poisson estimates and the ball-translation estimate before analyzing the algorithm.

\begin{lemma}[Poisson splitting]\label{lem:poisson-splitting}
For every integer $d\ge1$, every $n>0$, and every probability vector $p$ on $[d]$, let $M_0\sim\operatorname{Pois}(n/2)$. Draw, independently of $M_0$, $M_0$ i.i.d.\ points from $p$, and let $N_i$ be their counts. Then the $N_i$ are independent and
\[
N_i\sim\operatorname{Pois}(\mu_i),
\qquad
\mu_i=np_i/2.
\]
In particular, $\mu_i=0$ implies $N_i=0$ almost surely.
\end{lemma}

\textbf{Proof.}
For $s_i\in[0,1]$,
\begin{equation}
\begin{aligned}
\mathbb E\prod_i s_i^{N_i}
&=
\mathbb E\left(\sum_i p_i s_i\right)^{M_0}\\
&=
\exp\!\left(\frac n2\left(\sum_i p_i s_i-1\right)\right)\\
&=
\prod_i\exp(\mu_i(s_i-1)).
\end{aligned}
\label{eq:foundations-19}
\end{equation}
This is the product of the Poisson probability generating functions. The zero-mean assertion follows directly.~$\square$

\begin{lemma}[Scalar square-root Poisson estimates]\label{lem:poisson-square-root}
Let $N\sim\operatorname{Pois}(\mu)$, where $\mu\ge0$, and put
\[
W_\mu=(\sqrt N-\sqrt\mu)^2.
\]
If $\mu=0$, then $W_\mu=0$ almost surely. If $\mu>0$, then
\[
\mathbb E W_\mu\le1.
\]
For every $\mu\ge0$ and $u\ge0$,
\[
\Pr[W_\mu\ge u]\le2e^{-u},
\qquad
\mathbb E e^{W_\mu/2}\le3.
\]
\end{lemma}

\textbf{Proof.}
The case $\mu=0$ is immediate; at $u=0$, the tail bound is also valid.

For $\mu>0$,
\[
W_\mu
=
\frac{(N-\mu)^2}{(\sqrt N+\sqrt\mu)^2}
\le
\frac{(N-\mu)^2}{\mu}.
\]
Since $\operatorname{Var}(N)=\mu$,
\begin{equation}
\mathbb E W_\mu\le1.
\label{eq:foundations-20}
\end{equation}

Define the Poisson rate function
\[
I_\mu(x)=x\log(x/\mu)-x+\mu,
\qquad x\ge0.
\]
With $v=\sqrt{x/\mu}$,
\[
I_\mu(x)-(\sqrt x-\sqrt\mu)^2
=
2\mu v(v\log v-v+1)\ge0.
\]
The expression remains valid at $v=0$ under $0\log0=0$.

The moment generating function
\[
\mathbb E e^{sN}=\exp(\mu(e^s-1))
\]
and exponential Markov imply
\[
\Pr[N\ge x]\le e^{-I_\mu(x)}\quad(x\ge\mu),
\]
and
\[
\Pr[N\le x]\le e^{-I_\mu(x)}\quad(0\le x\le\mu).
\]
For $x>0$, choose $s=\log(x/\mu)$ in both cases ($s\ge0$ for the upper tail; $s\le0$ for the lower tail, where $\{N\le x\}=\{e^{sN}\ge e^{sx}\}$); $x=0$ follows by a limit, or directly from $\Pr[N=0]=e^{-\mu}$.

Apply these bounds at
\[
x_+=(\sqrt\mu+\sqrt u)^2,
\]
and, when $u\le\mu$, at
\[
x_-=(\sqrt\mu-\sqrt u)^2.
\]
When $u>\mu$, the lower branch is empty. Thus
\[
\Pr[W_\mu\ge u]\le2e^{-u}.
\]
Finally, tail integration gives
\begin{equation}
\begin{aligned}
\mathbb E e^{W_\mu/2}
&=
1+\frac12\int_0^\infty e^{u/2}\Pr[W_\mu>u]\,du\\
&\le
1+\int_0^\infty e^{-u/2}\,du
=3.
\end{aligned}
\label{eq:foundations-21}
\end{equation}
$\square$

\begin{lemma}[Vector error and truncation]\label{lem:poisson-vector}
Let $d\ge1$, let $n\ge1$ be an integer, let $p$ be a probability vector, and let $h\in P_d$. In the untruncated Poisson experiment, define
\[
t_i=(2h(i)-1)\sqrt{p_i},
\qquad
\widetilde t_i=(2h(i)-1)\sqrt{2N_i/n}.
\]
Then, for every $\varepsilon>0$,
\[
\mathbb E\|\widetilde t-t\|^2\le\frac{2d}{n},
\qquad
\Pr[\|\widetilde t-t\|^2>\varepsilon/4]
\le3^d e^{-n\varepsilon/16},
\]
and
\begin{equation}
\Pr[M_0>n]
\le e^{-n(\log2-1/2)}
\le e^{-n/8}.
\label{eq:foundations-22}
\end{equation}
\end{lemma}

\textbf{Proof.}
For $\mu_i=np_i/2$,
\[
\|\widetilde t-t\|^2
=
\frac2n\sum_i(\sqrt{N_i}-\sqrt{\mu_i})^2.
\]
A zero $\mu_i$ contributes zero almost surely. \Cref{lem:poisson-square-root} gives the expectation bound. Independence from \Cref{lem:poisson-splitting} gives
\[
\mathbb E\exp\!\left(\frac n4\|\widetilde t-t\|^2\right)
\le3^d.
\]
Exponential Markov proves the stated error tail.

For truncation, exponential Markov with parameter $\log2$ yields
\[
\Pr[M_0>n]
\le
\exp\!\left(\frac n2-n\log2\right).
\]
The elementary inequality
\[
\log x\ge\frac{2(x-1)}{x+1}\qquad(x\ge1)
\]
follows by differentiation. It gives $\log2\ge2/3>5/8$, proving the final bound.~$\square$

\begin{lemma}[Total variation under translation of a ball]\label{lem:ball-translation}
For every integer $d\ge1$, every $R>0$, and every $x,x'\in\mathbb R^d$, let $b=\|x-x'\|$. Then
\begin{equation}
\operatorname{TV}\!\left(
\operatorname{Unif}(x+RB),
\operatorname{Unif}(x'+RB)
\right)
\le
\frac{V_{d-1}}{V_d}\frac bR
\le
\sqrt d\,\frac bR.
\label{eq:foundations-23}
\end{equation}
\end{lemma}

\textbf{Proof.}
Rotate so that the translation is parallel to the first coordinate. On each perpendicular slice, the two intervals have the same length and are translated by $b$. The volume lost from one interval is at most $b$. Integrating over the perpendicular $(d-1)$-ball gives lost volume at most
\[
bV_{d-1}R^{d-1}.
\]
For equal-volume uniform laws, total variation is the lost volume divided by $V_dR^d$, proving the first inequality.

Slicing the unit ball gives
\[
\frac{V_d}{V_{d-1}}
=
\int_{-1}^{1}(1-u^2)^{(d-1)/2}\,du.
\]
For $d\ge3$, on $|u|\le1/\sqrt d$, Bernoulli's inequality gives
\[
(1-u^2)^{(d-1)/2}
\ge
1-\frac{d-1}{2d}
\ge\frac12.
\]
Therefore the integral is at least $1/\sqrt d$. For $d=1$, the ratio is $V_1/V_0=2$. For $d=2$, it is $\pi/2$, and the required inequality is immediate.~$\square$

\paragraph{The algorithm.}

Take $n$ as in \eqref{eq:foundations-ball}. The public seed first generates
\[
M_0\sim\operatorname{Pois}(n/2),
\]
independently of the public candidate sequence described below.

If $M_0>n$, output the all-zero function. Otherwise read the first $M_0$ examples. Let $N_i$ be the count of point $i$. If $N_i>0$, let $y_i$ be the label of the first occurrence of $i$ among the examples read (under realizability all occurrences carry the same label almost surely) and set
\[
\widehat t_i=(2y_i-1)\sqrt{2N_i/n};
\]
if $N_i=0$, set $\widehat t_i=0$.

Put
\[
R=\frac{\sqrt\varepsilon}{2}.
\]
Generate a public sequence of uniform candidates from
\[
(\sqrt2+R)B.
\]
Accept the first candidate $Y$ in $\widehat t+RB$, and output
\begin{equation}
g(i)=\mathbf1\{Y_i>0\},
\label{eq:foundations-24}
\end{equation}
with the displayed fixed convention at zero.

When $M_0\le n$,
\[
\|\widehat t\|^2=\frac{2M_0}{n}\le2,
\]
so the acceptance ball is contained in the proposal ball. The accepted point is uniform on $\widehat t+RB$, and acceptance occurs almost surely. Since every Boolean function on $[d]$ belongs to $P_d$, the learner is proper.

\paragraph{Accuracy.}

Fix any unknown marginal $p$ and any target $h\in P_d$. Let
\[
t_i=(2h(i)-1)\sqrt{p_i}.
\]
If $g(i)\ne h(i)$, the signs of $Y_i$ and $t_i$ disagree, allowing a zero endpoint, and
\[
p_i\le(Y_i-t_i)^2.
\]
The case $p_i=0$ is trivial. Summing gives
\begin{equation}
L_D(g)\le\|Y-t\|^2.
\label{eq:foundations-25}
\end{equation}

Realizability also gives the identity
\[
\widehat t_i=(2h(i)-1)\sqrt{2N_i/n},
\]
including unseen coordinates, where both sides are zero.

Analyze an infinite i.i.d. sample stream so that the untruncated estimator $\widetilde t$ is defined even when $M_0>n$. The actual algorithm never reads more than $n$ examples. On $M_0\le n$, its estimator equals $\widetilde t$.

The sample-size choice and \Cref{lem:poisson-vector} ensure
\[
\Pr[\|\widetilde t-t\|^2>\varepsilon/4]
\le\delta/2.
\]
They also ensure
\[
\Pr[M_0>n]\le\delta/2,
\]
because $0<\varepsilon\le1$ and
\[
\frac n8
\ge
\frac2\varepsilon
\left(d\log3+\log\frac2\delta\right)
\ge\log\frac2\delta.
\]
Outside these two bad events,
\[
\|Y-t\|
\le
\|Y-\widehat t\|+\|\widehat t-t\|
\le
R+\frac{\sqrt\varepsilon}{2}
=
\sqrt\varepsilon.
\]
Equation \eqref{eq:foundations-25} proves accuracy.

\paragraph{Strict replicability.}

Run the algorithm on two independent sample streams using the same public seed. In particular, the two runs share $M_0$. On $M_0>n$, both output the all-zero function.

Let $\widetilde t,\widetilde t'$ denote the two untruncated estimators. Conditional on $M_0\le n$ and the two samples, \Cref{lem:rejection-coupling,lem:ball-translation} apply to the two equal-volume acceptance balls, because the candidate sequence is independent of $M_0$ and the samples. Equality of accepted points implies equality of the whole output functions. Therefore
\[
\begin{aligned}
\Pr[g\ne g']
&\le
\frac{2\sqrt d}{R}\,
\mathbb E\!\left[
\mathbf1\{M_0\le n\}
\|\widetilde t-\widetilde t'\|
\right]\\
&\le
\frac{2\sqrt d}{R}\,
\mathbb E\|\widetilde t-\widetilde t'\|.
\end{aligned}
\]
The count vectors are not asserted to be independent across the two runs: they share $M_0$. Independence is unnecessary. The triangle inequality and the marginal second-moment bounds give
\[
\mathbb E\|\widetilde t-\widetilde t'\|
\le
\mathbb E\|\widetilde t-t\|
+
\mathbb E\|\widetilde t'-t\|
\le2\sqrt{2d/n}.
\]
Consequently,
\begin{equation}
\Pr[g\ne g']
\le
\frac{8\sqrt2\,d}{\sqrt{\varepsilon n}}
\le\rho.
\label{eq:foundations-26}
\end{equation}
This proves \eqref{eq:foundations-ball}.~$\square$

\subsection{The confidence lower bound}\label{sec:foundations-confidence}
The following two-target argument supplies the confidence terms in Theorem~\ref{thm:u} and equation~\eqref{eq:pd-comparison}.

\begin{lemma}[Confidence cost for arbitrary learners]\label{lem:confidence-cost}
For every integer $d\ge2$, every $0<\varepsilon\le1/4$, and every $0<\delta<1/2$, every randomized, possibly improper learner that satisfies the accuracy condition (the first line of \eqref{eq:model}) for all distributions realizable by $P_d$ must satisfy
\[
n\ge\frac1{4\varepsilon}\log\frac1{2\delta}.
\]

The same worst-case lower bound holds in the known-marginal interface. No replicability assumption is needed.
\end{lemma}

\textbf{Proof.}
Use two points of masses $1-2\varepsilon$ and $2\varepsilon$, with all other points having mass zero. Consider two targets that agree on the common point and have opposite labels on the rare point.

The event that the rare point is unseen has probability $(1-2\varepsilon)^n$. Conditional on this event, the labeled sample distribution is identical for the two targets. Every Boolean output has the wrong rare-point label for at least one target, and that error has mass $2\varepsilon>\varepsilon$. Summing the two failure probabilities gives
\[
2\delta\ge(1-2\varepsilon)^n.
\]
Thus
\begin{equation}
n\ge
\frac{\log(1/(2\delta))}
{-\log(1-2\varepsilon)}
\ge
\frac1{4\varepsilon}\log\frac1{2\delta},
\label{eq:foundations-27}
\end{equation}
since
\[
-\log(1-2\varepsilon)
=
\int_0^{2\varepsilon}\frac{du}{1-u}
\le4\varepsilon
\]
for $\varepsilon\le1/4$. The argument uses a single fixed marginal, so it also applies when that marginal is known.~$\square$

\subsection{The known-marginal characterization}\label{sec:foundations-known-pd}
This subsection proves the known-marginal half of \eqref{eq:pd-comparison}.
\begin{proposition}\label{prop:foundations-known-pd}
Let $d\ge2$, $\eps\le1/16$, and $\delta,\rho\le10^{-3}$. A deterministic proper learner, given $p$, is valid whenever
\[
n\ge\max\left\{\frac{2d}{\eps}\log\frac{2d}{\rho},
\frac2\eps\bigl(d\log2+\log(1/\delta)\bigr)\right\}.
\]
Conversely, put $k=d-1$ and assign mass $8\eps/k$ to each of $k$ rare points and the remaining mass to an anchor. Every valid learner given this marginal requires
\[
n\ge \frac{k}{16\eps}\log\frac{k}{6\rho}\ge\frac d{64\eps}\log\frac d\rho.
\]
\end{proposition}

\paragraph{Upper bound in Proposition~\ref{prop:foundations-known-pd}.}

Fix any known marginal $p$, and define
\[
T=\left\{i:p_i\ge\frac{\varepsilon}{2d}\right\}.
\]
For $i\in T$, output its true label if it has been observed, and output zero otherwise. Outside $T$, output zero.

If both runs cover $T$, the resulting whole functions coincide. Hence
\begin{equation}
\Pr[\text{the outputs differ}]
\le
2d\,e^{-n\varepsilon/(2d)}.
\label{eq:foundations-28}
\end{equation}

The total mass outside $T$ is at most $\varepsilon/2$. Inaccuracy therefore implies that the unseen points of $T$ have total mass greater than $\varepsilon/2$. For every subset $U\subseteq T$ of such mass, the probability that all its points are unseen is at most $e^{-n\varepsilon/2}$. A union bound over at most $2^d$ subsets gives
\begin{equation}
\Pr[L_D(g)>\varepsilon]
\le
2^d e^{-n\varepsilon/2}.
\label{eq:foundations-29}
\end{equation}
The stated sample-size condition makes \eqref{eq:foundations-28} at most $\rho$ and \eqref{eq:foundations-29} at most $\delta$.

\paragraph{Lower bound in Proposition~\ref{prop:foundations-known-pd}.}

Fix the explicit marginal in the proposition. Set $k=d-1$ and
$a=8\varepsilon/k$. The anchor has label zero, and the $k$ rare labels form a vector
\[
H\sim\operatorname{Unif}(\{0,1\}^k).
\]
Let $A_p$ be any randomized, possibly improper learner, given $p$, that is valid for every target.

For each seed $r$ and target $H$, choose a modal output function $g_r(H)$, breaking ties in a fixed order. Replicability implies
\[
\mathbb E_{r,H}\Pr_S[A_p(S;r)\ne g_r(H)]\le\rho.
\]
An accurate output has at most $k/8$ erroneous rare labels, since each rare point has mass $a=8\varepsilon/k$. Hence, for every sample $S$, $\operatorname{Ham}(g_r(H),H)\le k/8+k\,\mathbf 1\{\operatorname{Ham}(g_r(H),H)>k/8\}$, and the indicator is at most $\mathbf 1\{A_p(S;r)\ne g_r(H)\}+\mathbf 1\{L_{D_{p,H}}(A_p(S;r))>\varepsilon\}$. Taking expectations over $S$ and then over $(r,H)$, and using accuracy and the replicability bound above, gives
\begin{equation}
\mathbb E_{r,H}\operatorname{Ham}(g_r(H),H)
\le
k\left(\frac18+\delta+\rho\right),
\label{eq:foundations-30}
\end{equation}
where Hamming distance here counts only rare coordinates.

For a function-valued map $g$ on the cube, define its total edge influence by
\[
I(g)=
\sum_{i=1}^k
\Pr_H[g(H)\ne g(H^{\oplus i})].
\]
If both endpoints of an $i$-edge predict their respective $i$-th target bit correctly, their output functions differ. Consequently,
\begin{equation}
\begin{aligned}
\mathbb E_r I(g_r)
&\ge
k-2\mathbb E_{r,H}\operatorname{Ham}(g_r(H),H)\\
&\ge
\frac{373}{500}k.
\end{aligned}
\label{eq:foundations-31}
\end{equation}

Let
\[
q=(1-a)^n,
\qquad
\theta=\min\{1,(2kq)^{-1}\}.
\]
Given the sample positions, independently select each unseen rare coordinate with probability $\theta$, obtaining a set $J$. The additional selection coins are independent of $H,r$, and the sample positions. Define
\[
H'=H\oplus\mathbf1_J.
\]

No sampled label changes, because $J$ contains only unseen coordinates. Moreover, conditional on the positions and $J$, the map $H\mapsto H'$ is a bijection of the cube. Hence
\[
(r,H',X_{1:n})\overset d=(r,H,X_{1:n}).
\]
The learner receives exactly the same labeled sample under $H$ and $H'$. Comparing its common output to the two modes gives
\begin{equation}
\Pr[g_r(H)\ne g_r(H')]\le2\rho.
\label{eq:foundations-32}
\end{equation}

For each $i$,
\begin{equation}
\begin{aligned}
\Pr[J=\{i\}]
&\ge
\Pr[i\in J]
-
\sum_{j\ne i}\Pr[i,j\in J]\\
&\ge
\theta q-(k-1)\theta^2q^2\\
&\ge\frac{\theta q}{2}.
\end{aligned}
\label{eq:foundations-33}
\end{equation}
Indeed, a rare coordinate is unseen with probability $q$. For distinct rare coordinates, the probability both are unseen is
\[
(1-2a)^n\le(1-a)^{2n}=q^2,
\]
where $a=8\varepsilon/k\le1/2$ (as $\varepsilon\le1/16$ and $k\ge1$) gives $0\le1-2a\le(1-a)^2$.
When $k=1$, there are no pair terms. Finally,
$\theta kq\le1/2$, which proves the last line of \eqref{eq:foundations-33}.

The set $J$ is independent of $(r,H)$. Keeping only singleton $J$ events in \eqref{eq:foundations-32}, and using \eqref{eq:foundations-31}, yields
\begin{equation}
2\rho
\ge
\frac{\theta q}{2}\,\mathbb E_r I(g_r)
\ge
\frac{373}{1000}
\min\{kq,1/2\}.
\label{eq:foundations-34}
\end{equation}
Since $\rho\le10^{-3}$, the minimum cannot equal $1/2$. Therefore
\[
kq\le\frac{2000}{373}\rho<6\rho.
\]

Because $a\le1/2$ (see above),
\[
-\log(1-a)\le2a,
\qquad
q\ge e^{-2an}.
\]
Thus
\[
n\ge
\frac{k}{16\varepsilon}\log\frac{k}{6\rho}.
\]
Also $k\ge d/2$, and $d/\rho\ge2000>144$, so
\[
\log\frac{k}{6\rho}
\ge
\log\frac{d}{12\rho}
\ge
\frac12\log\frac d\rho.
\]
Hence
\begin{equation}
n\ge
\frac{k}{16\varepsilon}\log\frac{k}{6\rho}
\ge
\frac d{64\varepsilon}\log\frac d\rho.
\label{eq:foundations-35}
\end{equation}

This argument also excludes $n=0$: then $q=1$, making \eqref{eq:foundations-34} impossible.

Combining \eqref{eq:foundations-35} with \Cref{lem:confidence-cost} proves the lower half of \eqref{eq:pd-comparison}. The two lower bounds may use different explicit marginals; this is legitimate because $n^{\mathrm{known}}_{P_d}$ takes a supremum over marginals. The upper bound proves the other half, with absolute constants.

For a point-mass marginal, one sample reveals the only relevant target label; all other output labels can be fixed to zero. Thus the lower bound cannot be interpreted as holding for every marginal.~$\square$

\subsection{VC and product-marginal reductions}\label{sec:foundations-reductions}
These reductions prove the generalizations of the anchored obstruction stated after \Cref{thm:anchor}.
\begin{proposition}\label{prop:foundations-vc}
This lower-bound reduction also applies to measurable classes $H$ that are not finite, with the same unknown-marginal validity requirement. If $\VC(H)=v\ge2$, $\eps\le1/16$, and $\delta,\rho\le10^{-3}$, then
\begin{equation}\label{eq:foundations-vc}
N_H^U\ge\cstar(v-1)^2/(\eps\rho^2).
\end{equation}
At $\eps=\delta=\rho=10^{-4}$ this is at least $814(v-1)^2$. Infinite VC dimension precludes a finite uniform sample bound.
\end{proposition}

\paragraph{Proof of Proposition~\ref{prop:foundations-vc}.}

Choose a shattered set of $v$ points. Designate one point as an anchor, and identify the other $v-1$ points with the non-anchor coordinates of $H_{v-1}^{\mathrm{anc}}$.

For every parameter $t$, place the corresponding anchored distribution entirely on this shattered set. Shattering ensures realizability by $H$. Restrict the learner's output function to the shattered set. Its loss is unchanged because the distribution is supported there. Moreover, equality of the entire original output functions implies equality of their restrictions, so restriction can only increase collision probability.

The restricted learner is therefore a randomized, possibly improper learner to which Theorem~\ref{thm:anchor} applies with dimension $v-1$. This gives \eqref{eq:foundations-vc}.

At the fixed parameters,
\[
10^{12}c_\star
=
\frac{4778265625}{37636}
>814,
\]
which yields the stated fixed-parameter bound.~$\square$

\subsection{The product-marginal construction}
\begin{proposition}\label{prop:foundations-product}
On $\{0,\ldots,d\}\times\mathbb Z_2$, let $h_v(a,b)=v_a\oplus b$, where $v_0=0$. For $t\in B_2^d$, take the product marginal $w(t)\times\Unif(\mathbb Z_2)$, where $w_i(t)=16\eps t_i^2$ and $w_0(t)=1-16\eps\|t\|^2$, and choose $v_i=\ind\{t_i>0\}$. If $\eps\le1/32$ and $\delta,\rho\le10^{-3}$, one learner valid throughout this family requires
\begin{equation}\label{eq:foundations-product}
n\ge\cstar d^2/(2\eps\rho^2).
\end{equation}
In particular, $n\ge700d^2$ at the fixed parameters $10^{-4}$.
\end{proposition}

\paragraph{Proof of Proposition~\ref{prop:foundations-product}.}

Let $F$ be any Boolean output function on
$\{0,\ldots,d\}\times\mathbb Z_2$. For each block $a$, round the pair
\[
(F(a,0),F(a,1))
\]
to whichever of the two complementary parity patterns
\[
(0,1),\qquad(1,0)
\]
has fewer disagreements with it. Break ties by a fixed rule, and let $f(a)$ be the resulting block bit.

If the true block bit is $v_a$, a wrong rounded bit requires block error at least $1/2$. Hence
\[
\mathbf1\{f(a)\ne v_a\}
\le
2\Pr_{b\sim\operatorname{Unif}(\mathbb Z_2)}
[F(a,b)\ne v_a\oplus b].
\]
Writing $Q_t$ for the labeled product distribution and
$D_t^{(16\varepsilon)}$ for its collapsed anchored distribution, we obtain
\begin{equation}
L_{D_t^{(16\varepsilon)}}(f)
\le2L_{Q_t}(F).
\label{eq:foundations-36}
\end{equation}
Thus product accuracy at threshold $\varepsilon$ implies rounded accuracy at threshold $2\varepsilon$. Whole-function equality is preserved by this deterministic rounding.

We apply the proof of Theorem~\ref{thm:anchor} directly to the product input experiment. No simulation using shared random block bits is required.

Indeed, $Q_t$ is obtained by splitting each labeled atom of
$D_t^{(16\varepsilon)}$ into two distinguishable atoms of half its probability. Consequently,
\begin{equation}
\sum_z\sqrt{Q_t(z)Q_s(z)}
=
\sum_z
\sqrt{D_t^{(16\varepsilon)}(z)
      D_s^{(16\varepsilon)}(z)}.
\label{eq:foundations-37}
\end{equation}
The Hellinger calculation and sample-experiment Lipschitz bound are therefore unchanged by duplication: by \eqref{eq:foundations-03}, $\operatorname{TV}(Q_t^n,Q_s^n)\le\sqrt{2\cdot16\varepsilon n}\,\|t-s\|=4\sqrt{\varepsilon'n}\,\|t-s\|$ with $\varepsilon'=2\varepsilon$, which is the constant of \Cref{lem:input-continuity} at accuracy $\varepsilon'$.

With $\varepsilon'=2\varepsilon$, one has
$16\varepsilon=8\varepsilon'$ and
$\varepsilon'\le1/16$. The rounded outputs have the accurate-region volume bound from \Cref{lem:accurate-region-volume} at threshold $\varepsilon'$. Their high-probability cells satisfy \Cref{lem:output-cells} with input experiment $Q_t^n$: its proof uses only replicability of the rounded output under the input law and the accuracy bound $\Pr_{S,r}[L_{D_t^{(16\varepsilon)}}(f)>\varepsilon']\le\Pr_{S,r}[L_{Q_t}(F)>\varepsilon]\le\delta$ from \eqref{eq:foundations-36}. \Cref{lem:cell-packing} applies unchanged. The \eqref{eq:anchor-rho} proof therefore gives
\[
n\ge
c_\star\frac{d^2}{\varepsilon'\rho^2}
=
c_\star\frac{d^2}{2\varepsilon\rho^2}.
\]

At the fixed parameters,
\[
\frac{10^{12}c_\star}{2}
=
\frac{4778265625}{75272}
>700.
\]
This proves Proposition~\ref{prop:foundations-product}.~$\square$

\section{Shared sampling and the unknown-marginal upper bound}\label{sec:proof-u}
This appendix certifies the truncated-power module in \Cref{sec:module-u}. We first prove the two reusable stability tools. The same deletion argument will support the block algorithm, and the same public sampler will support the known-marginal modules.

\begin{lemma}[Exponential-clock public sampling]\label{lem:clock}
For probability vectors $P,Q$ on one fixed finite index set, let $E_h$ be independent rate-one exponential variables shared between the two runs. The rule $\arg\min_h E_h/P(h)$ has law $P$, with zero denominators interpreted as infinity. The two outputs disagree with probability at most $2\TV(P,Q)/(1+\TV(P,Q))$.
\end{lemma}
\begin{proof}
For $P(h)>0$, $E_h/P(h)$ is exponential with rate $P(h)$, independently across $h$. Integrating the density of its being the first clock gives probability $P(h)/\sum_gP(g)=P(h)$. Ties have probability zero and are resolved by the fixed index order.

If $P(h),Q(h)>0$, condition on $E_h=t$. It wins both competitions exactly when
$E_g\ge t\max\{P(g)/P(h),Q(g)/Q(h)\}$ for every $g$. Integrating $t$ gives simultaneous winning probability
\[
\left[\sum_g\max\{P(g)/P(h),Q(g)/Q(h)\}\right]^{-1}
\ge\frac{\min\{P(h),Q(h)\}}{\sum_g\max\{P(g),Q(g)\}}.
\]
Sum over $h$. The numerator sum is $1-\TV(P,Q)$ and the denominator is $1+\TV(P,Q)$. Coordinates with a zero mass contribute zero to this lower bound. Subtract the resulting collision probability from one.
\end{proof}

\begin{lemma}[Deletion energy]\label{lem:deletion}
Let $F(S)$ take values in a finite-dimensional Hilbert space, with $S$ an independent sample of size $n$. For each $i$, let $F_{-i}$ be any function of the sample with observation $i$ removed, keeping all public parameters fixed. If $S'$ is an independent copy, then
\[
\E\|F(S)-F(S')\|^2\le2\sum_{i=1}^n\E\|F(S)-F_{-i}\|^2.
\]
If $0\le\psi\le\psi'$ coordinatewise and $\psi\ne0$, then
\begin{equation}\label{eq:normalization}
\left\|\frac{\psi'}{\|\psi'\|}-\frac\psi{\|\psi\|}\right\|^2
\le\frac{\|\psi'-\psi\|^2}{\|\psi\|^2}.
\end{equation}
\end{lemma}
\begin{proof}
The variance tensorization inequality for independent coordinates gives
$\Var(F)\le\sum_i\E\|F-\E[F\mid S_{-i}]\|^2$; it follows by the martingale decomposition, or by applying scalar Efron--Stein in an orthonormal basis and summing. Conditional expectation is the least-squares predictor from $S_{-i}$, so each summand is at most $\E\|F-F_{-i}\|^2$. Finally $\E\|F-F'\|^2=2\Var(F)$.
For the second statement put $t=\|\psi'\|/\|\psi\|\ge1$ and let $a$ be the cosine of the angle between the vectors. The right side minus the left side is $(t-1)(t+1-2a)\ge0$.
\end{proof}

\begin{lemma}[Truncated-power stability]\label{lem:power}
Let $C_h$ be sample error counts for a finite realizable Boolean class of size $m$. Fix an integer $k\ge\max\{4,2\log m\}$ and an integer $B\ge2k$. Put
$P_S(h)=(1-C_h/B)_+^k/Z_S$. Then
\begin{equation}\label{eq:power-energy}
\sum_i\|\sqrt{P_S}-\sqrt{P_{S_{-i}}}\|_2^2\le\frac{e^2k^2}{4B},
\qquad
\E\TV(P_S,P_{S'})\le\frac{ek}{\sqrt{2B}}.
\end{equation}
The deletion distributions use the original $B,k$.
\end{lemma}
\begin{proof}
Write $u_h=(1-C_h/B)_+$, $\psi_h=u_h^{k/2}$, and $Z=\sum_hu_h^k\ge1$. The last inequality follows from realizability. Deleting an erroneous observation changes coordinate $h$ from $u_h^{k/2}$ to $v_h^{k/2}$, where $v_h=(1-(C_h-1)/B)_+$. For $C_h\ge1$, the mean-value theorem gives
\[
0\le v_h^{k/2}-u_h^{k/2}\le\frac{k}{2B}v_h^{k/2-1}.
\]
There are exactly $C_h$ deletions of this type. Apply \eqref{eq:normalization} before summing:
\[
\sum_i\|\sqrt{P_S}-\sqrt{P_{S_{-i}}}\|^2
\le\frac{k^2}{4B^2Z}\sum_h C_hv_h^{k-2}
\le\frac{k^2}{4BZ}\sum_hv_h^{k-2}.
\]
For the final inequality, a positive summand has $1\le C_h\le B$, since counts and $B$ are integers. Coordinates with $C_h=0$ contribute zero before this bound.

For all $h$, $v_h\le u_h+1/B$. H\"older's inequality and Minkowski's inequality yield
\[
\sum_hv_h^{k-2}\le m^{2/k}\left(\sum_hv_h^k\right)^{(k-2)/k},
\qquad
\left(\sum_hv_h^k\right)^{1/k}\le Z^{1/k}+m^{1/k}/B.
\]
Divide by $Z$, use $Z\ge1$, $m^{2/k}\le e$, and $m^{1/k}\le\sqrt e<2$. Thus
\[
\frac{\sum_hv_h^{k-2}}Z
\le e\left(1+\frac{m^{1/k}}B\right)^{k-2}
\le e\exp(k\sqrt e/B)\le e^2.
\]
This proves the deterministic energy bound. Apply \Cref{lem:deletion}, followed by $\TV(P,Q)\le\HH(P,Q)$ and Jensen, to obtain the expected-TV estimate.
\end{proof}

\begin{proof}[Proof of \Cref{thm:u}]
The chosen sample size gives $B\ge16k^2/\rho^2-1\ge15k^2/\rho^2\ge2k$. Every returned hypothesis has $C_h<B\le n\eps/2$. For fixed $h$ with $L(h)>\eps$, the multiplicative binomial lower-tail bound gives
\[
\Pr[C_h\le n\eps/2]\le e^{-n\eps/8}.
\]
For completeness, apply exponential Markov to $e^{-tC_h}$ and optimize over $t>0$, or use $\Pr[C_h\le nL(h)/2]\le e^{-nL(h)/8}$. A union bound over $m$ hypotheses is at most $\delta$ under \eqref{eq:u-budget}. This event controls all possible clock outcomes simultaneously.

By \Cref{lem:clock,lem:power}, the replication failure is at most
\[
2\E\TV(P_S,P_{S'})\le\frac{\sqrt2ek}{\sqrt B}
\le\sqrt{2e^2/15}\,\rho<\rho.
\]
For the minimax lower bound take $d=\lfloor\log_2M\rfloor\ge1$ in \Cref{thm:anchor}. Its class has size $2^d\le M$, and $d$ is within absolute factors of $\log M$. The two-hypothesis rare-point construction in \Cref{sec:foundations-confidence} gives the separate confidence term even when $M=2$. The maximum of the two lower bounds is at least half their sum. The upper bound follows uniformly for all $m\le M$, since $k=O(1+\log m)$ and the term $\log m/\eps$ is absorbed by $(\log M)^2/(\eps\rho^2)$ for $M\ge2$. This proves the claimed rate.
\end{proof}

\section{Known-marginal estimation, chains, and deterministic components}\label{sec:proof-k}
The control-variate estimates below certify \Cref{thm:k} and are reused by the finite-grid and VC modules. We then prove the threshold-chain obstruction and the rank-ball construction needed for \Cref{thm:det}.

\begin{lemma}[Control-variate distribution estimate]\label{lem:control}
Let $g$ be a Boolean predictor independent of a correction block of size $N$, and put $r=L(g)$. For a finite candidate set $F$, define $\widetilde L$ by \eqref{eq:control-estimator}. Let
\[
Q(f)=\frac{e^{-\lambda L(f)}}{\sum_{u\in F}e^{-\lambda L(u)}},\qquad
P(f)=\frac{e^{-\lambda\widetilde L(f)}}{\sum_{u\in F}e^{-\lambda\widetilde L(u)}}.
\]
If $\lambda/N\le1/4$, then
\[
\E[\KL(Q\Vert P)\mid g]\le\lambda^2r/N,
\qquad \E\TV(P,Q)\le\lambda\sqrt{\E r/(2N)}.
\]
\end{lemma}
\begin{proof}
For a single correction define $Z_f=\ind\{f(X)\ne Y\}-\ind\{f(X)\ne g(X)\}$. It vanishes whenever $g(X)=Y$, so $|Z_f|\le1$ and $\E Z_f^2\le r$. Moreover, $d_p(f,g)+\E Z_f=L(f)$. For $0\le t\le1/4$ and $|z|\le1$, Taylor's formula gives
$e^{-tz}\le1-tz+(t^2e^t/2)z^2$. Using $\log u\le u-1$ and $e^{1/4}/2<1$, we obtain
\[
\log\E e^{-t(Z_f-\E Z_f)}\le t^2\E Z_f^2\le t^2r.
\]
Independence within the correction block therefore gives, for $\xi_f=\widetilde L(f)-L(f)$,
$\E[e^{-\lambda\xi_f}\mid g]\le e^{\lambda^2r/N}$ and $\E[\xi_f\mid g]=0$.

Retaining the normalizing constants gives the identity
\[
\KL(Q\Vert P)=\lambda\sum_fQ(f)\xi_f+
\log\sum_fQ(f)e^{-\lambda\xi_f}.
\]
The first term has zero conditional mean. Jensen applied to the logarithm bounds the second conditional mean by $\lambda^2r/N$. Pinsker's inequality $\TV^2\le\KL/2$, followed by Jensen over all data, proves the second assertion. Dependence among corrections for different hypotheses is allowed throughout.
\end{proof}

\begin{lemma}[Shared-radius clipping]\label{lem:clip}
Suppose $Q\{f:L(f)\le\eps/8\}\ge1/2$. For a shared $R\sim\Unif[\eps/4,\eps/2]$, put
$A_g(R)=\{f:d_p(f,g)\le R\}$ and $B_*(R)=\{f:L(f)\le R\}$.
Let $P^R=P(\cdot\mid A_g(R))$ when the set is nonempty and otherwise take any fixed point mass; let $Q^R=Q(\cdot\mid B_*(R))$. Then
\begin{equation}\label{eq:clip}
\E_R\TV(P^R,Q^R)\le2\TV(P,Q)+8L(g)/\eps.
\end{equation}
For two independent data runs using the same radius and clocks, their final disagreement is at most
\begin{equation}\label{eq:cv-collision}
8\E\TV(P,Q)+32\E L(g)/\eps.
\end{equation}
\end{lemma}
\begin{proof}
Write $r=L(g)$. If $r>\eps/8$, the right side of \eqref{eq:clip} is at least one. Otherwise $|d_p(f,g)-L(f)|\le r$ implies that both clipping sets contain $\{L\le\eps/8\}$, hence have $Q$-mass at least $1/2$. In particular the actual set is nonempty. For any common conditioning set $A$,
\[
\TV(P|A,Q|A)\le\frac{\TV(P,Q)}{\max\{P(A),Q(A)\}}
\le\frac{\TV(P,Q)}{Q(A)}.
\]
To verify the first inequality, suppose $P(A)\ge Q(A)$. The common mass after normalization is at least
$\sum_{f\in A}\min\{P(f),Q(f)\}/P(A)\ge1-\TV(P,Q)/P(A)$; interchange the laws in the other case.
For the two $Q$-conditionals, direct normalization gives
$\TV(Q|A,Q|B)\le Q(A\triangle B)/\min\{Q(A),Q(B)\}\le2Q(A\triangle B)$.
For each $f$, the radii for which its two memberships differ form an interval of length at most $r$. Averaging over an interval of length $\eps/4$ gives $\E_RQ(A_g\triangle B_*)\le4r/\eps$. The triangle inequality proves \eqref{eq:clip}.

The reference $Q^R$ depends on the true target and shared radius but not on either sample. Insert its clock output between the two actual outputs and apply \Cref{lem:clock} to each pair. The result is $4\E\TV(P^R,Q^R)$, which yields \eqref{eq:cv-collision}.
\end{proof}

\begin{proof}[Proof of the upper bound in \Cref{thm:k}]
A first consistent anchor from $N$ examples has
\[
\Pr[L(g)>u]\le\min\{1,me^{-Nu}\},\qquad
\E L(g)\le(1+\log m)/N=a/N.
\]
The expectation follows by integrating the tail on $[0,\infty)$; integrating only to one can only decrease it. With $\lambda=8\log(2m)/\eps$, a zero-risk hypothesis supplies a partition-function lower bound one, so
$Q\{L>\eps/8\}\le me^{-\lambda\eps/8}=1/2$.
Also $\log(2m)\le a$ and the sample budget ensures $\lambda/N\le1/4$.

Combine \Cref{lem:control,lem:clip} to bound disagreement by
\[
4\sqrt2\lambda\sqrt{a}/N+32a/(N\eps)
\le\frac{32(\sqrt2+1)a^{3/2}}{N\eps}<\rho.
\]
Every output lies within $\eps/2$ of $g$, so its risk is at most $L(g)+\eps/2$. Failure therefore has probability at most $me^{-N\eps/2}\le\delta$. The class-size exponent part is completed after the grid lower bound in \Cref{sec:grid-consequences}.
\end{proof}

\subsection{A fixed finite chain detects the discrete replication scale}\label{sec:chain-proof}
This lemma supplies the lower bounds in the deterministic, saturation, and VC-one modules. It also handles small dimensions in the product-grid theorem.

\begin{lemma}[Threshold-chain interpolation]\label{lem:chain}
Consider $J+1$ nested thresholds whose endpoint disagreement mass is $s$, with each adjacent disagreement interval of mass $s/J$. Suppose $2\eps<s\le1/2$ and $\delta\le1/16$. Any valid learner, including randomized improper learners, satisfies
\begin{equation}\label{eq:chain}
n\ge\frac{J}{128s}\log\left(1+\frac1{J\rho}\right),
\qquad
n\ge\frac1{2s}\log\frac1{2\delta}.
\end{equation}
If the learner is deterministic and $\delta\le1/4$, its first bound strengthens to
$n\ge J\log(1/(2\rho))/(2s)$.
If the shared seed has at most $K$ atoms and $\delta<1/8$, it also satisfies
\begin{equation}\label{eq:chain-seed}
n\ge\frac{J}{2s}\left[\log\frac{1-8\delta}{2\rho\min\{K,J+1\}}\right]_+.
\end{equation}
\end{lemma}
\begin{proof}
Let $\mathcal E=\{f:d_p(f,h_0)\le\eps\}$ and, at a fixed seed $r$, set $u_j(r)=\Pr_S[A(S;r)\in\mathcal E]$ under target $h_j$. Accuracy and disjoint endpoint accuracy balls give $\E_ru_0\ge1-\delta$ and $\E_ru_J\le\delta$. Markov's inequality shows that seeds with $u_0\ge3/4$, $u_J\le1/4$ have total probability at least $1-8\delta$.

Adjacent sample experiments share exactly the event of avoiding their single disagreement interval, of probability $\theta=(1-s/J)^n$. Thus
$|u_{j+1}-u_j|\le1-\theta\le L:=ns/J$. For each target and seed, two outputs disagree with probability at least $2u_j(1-u_j)$.
The endpoint accuracy conditions exclude $n=0$, since then the output law is independent of the target. Hence $L>0$.

If $L\le1/4$, a sequence crossing from $3/4$ to $1/4$ with steps at most $L$ has at least $1/(4L)$ entries strictly between these values. Indeed, use the last entry at least $3/4$ before the first subsequent entry at most $1/4$: traversing a gap $1/2$ takes at least $1/(2L)$ steps, leaving at least $1/(2L)-1\ge1/(4L)$ intermediate entries. Each contributes at least $3/8$. Average over at least half the seed mass and sum over the $J+1\le2J$ targets to obtain
\[
L\ge\min\{1/4,3/(128J\rho)\}.
\]
Independently of this small-step case, some adjacent pair straddles $1/2$. Since their difference is at most $1-\theta$, at least one member lies in $[\theta/2,1-\theta/2]$ and contributes at least $\theta/2$. Averaging gives $\theta\le8J\rho$, hence
$L\ge\tfrac12[\log(1/(8J\rho))]_+$, using $-\log(1-s/J)\le2s/J$.

For $z>0$,
\[
\max\{\min(1/4,3z/128),\tfrac12[\log(z/8)]_+\}
\ge\tfrac1{128}\log(1+z).
\]
For $z\le16$, use $\log(1+z)\le z$ before the minimum saturates, and $\log17<4$ after it does. For $z\ge16$, $\log(z/8)\ge\frac18\log(1+z)$ follows by checking $z=16$ and differentiating. Substitute $z=1/(J\rho)$ to obtain the first bound.

For deterministic learners, the endpoint conditions hold without averaging seeds. The straddling target alone forces $\rho\ge\theta/2$, proving the stronger bound. With at most $K$ seed atoms, some good atom has mass at least $(1-8\delta)/K$. Alternatively distribute good seed mass among its selected straddling targets: some target receives at least $(1-8\delta)/(J+1)$. Either argument gives
$\rho\ge(1-8\delta)\theta/(2\min\{K,J+1\})$, proving \eqref{eq:chain-seed}.

Finally, the endpoint samples coincide whenever they avoid total disagreement mass $s$. Their common output cannot be accurate for both endpoints. Therefore $2\delta\ge(1-s)^n$; use $-\log(1-s)\le2s$ for the confidence bound.
\end{proof}

\subsection{Rank-adaptive components}\label{sec:det-proof}
We now verify the construction in \Cref{sec:module-det}; its metric statement applies even when distinct hypotheses have distance zero.

\begin{lemma}[Rank balls]\label{lem:rank}
On a finite pseudometric space of $m$ indexed points, let
$k_h=\max\{j:r_h(j)\le aj\}$ and $B_h=\{g:d(h,g)\le ak_h\}$. Connect every ball to its center. Each component has diameter at most $4am$. For every $h$, if $g$ is outside its component and
$R_h(g)=|\{u:d(h,u)\le d(h,g)\}|$, then $d(h,g)>aR_h(g)$. Consequently
\[
\sum_{g\notin\operatorname{comp}(h)}e^{-nd(h,g)}\le(e^{na}-1)^{-1}.
\]
\end{lemma}
\begin{proof}
The index $1$ is admissible, since $r_h(1)=0$. The ball contains exactly $k_h$ points: if it contained $j>k_h$, then $r_h(j)\le ak_h\le aj$, contradicting maximality.
To connect two points in a component choose a shortest chain of intersecting balls between them. Nonadjacent balls in the chain are disjoint, since an intersection would shorten the chain. The odd balls are disjoint, and so are the even balls. The sum of their cardinalities is therefore at most $2m$. Traversing their diameters costs at most $2a\sum k_h\le4am$.
If an external point $g$ had $d(h,g)\le aR_h(g)$, then $R_h(g)$ would be an admissible rank and at most $k_h$. Thus $g\in B_h$, a contradiction. The $j$th closest external point has distance greater than $aj$. Summing the resulting geometric series proves the final assertion. Tied and zero distances are already included in $R_h$.
\end{proof}

\begin{proof}[Proof of \Cref{thm:det}]
Choose $0<\eta<\eps$ and $a=\eta/(4m)$. By \Cref{lem:rank}, returning the representative of the first consistent hypothesis $g$ increases risk by at most $\eta$. The probability of $L(g)>\eps-\eta$ is at most $me^{-n(\eps-\eta)}$.
For a fixed target $h^*$, a hypothesis $h$ outside its component stays consistent with probability at most $e^{-nd_p(h,h^*)}$. The lemma bounds their total probability by $(e^{n\eta/(4m)}-1)^{-1}$. Outside this event the output is the fixed target-component representative. Two runs disagree with probability at most twice this bound. It suffices that
\[
n\ge\max\{4m\log(1+2/\rho)/\eta,\ \log(m/\delta)/(\eps-\eta)\}.
\]
Take $A=4m\log(1+2/\rho)$, $B=\log(m/\delta)$ and $\eta=\eps A/(A+B)$ to obtain \eqref{eq:det-budget}.

For the lower bound let $J=M-1$, $b=\lceil(4\eps)^{-1}\rceil$, and take thresholds $h_j(x)=\ind\{x\le j\}$, $0\le j\le J$, on $[bJ]$ with uniform marginal. The active mass $s=1/b$ satisfies $2\eps<s\le4\eps\le1/4$. Apply the deterministic part of \Cref{lem:chain} and its confidence bound. Since $J\ge M/2$ and $\rho\le1/8$, these bounds match the upper bound within absolute constants.

For randomized learners use the interpolation bound of the same lemma. If $\rho\le1/(16M^2)$, then
\[
\log\left(1+\frac1{J\rho}\right)\ge\log(1/\rho)-\log M\ge\tfrac12\log(1/\rho).
\]
This gives $n\ge M\log(1/\rho)/(2048\eps)$, together with the same confidence term. The deterministic upper bound applies also to randomized learning. This proves saturation.
\end{proof}

For comparison, a uniform-radius component rule joins every pair at distance at most $\eps/(2(m-1))$, returns the first consistent hypothesis's fixed component representative, and uses
$n\ge\max\{2(m-1)\log(2m/\rho)/\eps,2\log(m/\delta)/\eps\}$.
Each component has diameter at most $\eps/2$, and a consistent external hypothesis has probability at most $e^{-n\eps/(2(m-1))}$. The same two-event proof gives this bound. Rank balls improve precisely its replication term.

\section{Finite threshold products and their posterior kernels}\label{sec:grid-proof}
This appendix proves \Cref{thm:grid}, the interpolation module in \Cref{sec:module-grid}. The proof has three parts: couple targets through their posterior, bound the resulting crossing of output cells, and construct a matching control-variate learner. A complementary continuous-kernel argument and its finite-uniform consequence are given in \Cref{sec:continuous-kernel}.

\subsection{Posterior resampling and modal cells}
Throughout the finite-grid proof, $b,q,H_{d,k}$ are as in \Cref{sec:module-grid}. Give the target coordinates $J_i$ independent uniform priors on $\{0,\ldots,k\}$. A fixed-seed output distribution $Q_{j,r}$ has a fixed modal output $M_r(j)$; put $a_r(j)=1-Q_{j,r}(M_r(j))$. Since $\sum_fQ(f)^2\le\max_fQ(f)$, replicability implies $\E_ra_r(j)\le\rho$. A wrong mode contributes at most the sum of the algorithm's accuracy failure and its probability of differing from that mode, so
$\Pr_r[L_j(M_r(j))>\eps]\le\delta+\rho$.

We first justify Poissonization. For the all-zero and all-$k$ targets, both modes are accurate and hence distinct on at least $1-2\delta-2\rho$ seed mass. Their sample experiments coincide on the event of avoiding the active layer. Comparing the same output with these two different modes gives
\[
2\rho\ge(1-2\delta-2\rho)(1-q)^n.
\]
On \eqref{eq:domain}, $(1-q)^n\le3\rho$ and
$n\ge\log(1/(3\rho))/(2q)\ge\log(1/(3\rho))$. Generate an ordered Poisson sample of mean $8n$. Use its first $n$ observations if available and otherwise return a fixed function. Its shortage probability satisfies
\[
t_n\le e^{-(7-\log8)n}\le e^{-4n}\le(3\rho)^4\le\rho/100.
\]
The resulting accuracy and replication parameters are
$\bar\delta=\delta+t_n$ and $\bar\rho=\rho+2t_n\le1.02\rho$.
The Poisson count belongs to the data experiment and is independent in the two runs; it is not added to the shared seed.

Each active atom now has independent Poisson count of mean
\begin{equation}\label{eq:grid-beta}
\beta=8nq/(dk),\qquad t=e^{-\beta},\qquad z=e^\beta-1.
\end{equation}
Observing location $x$ cuts the target-state chain between $x-1$ and $x$. Every cut is present independently with probability $1-t$. Conditional on the complete labeled sample, each target coordinate is uniform on its remaining interval; the coordinates are conditionally independent. Sample $J'$ independently from this posterior given the same sample $S$. Both $(J,S)$ and $(J',S)$ have the original joint law.

Let $K=K_{k,\beta}$ be the one-dimensional transition: generate the independent cuts, then select a uniform state in the interval containing the input state. For a fixed cut pattern, interval averaging is an orthogonal projection. Thus $K$ is a self-adjoint Markov operator on the uniform state space and $0\preceq K\preceq I$. The posterior transition on $d$ coordinates is $K^{\otimes d}$.

From this point, $Q_{j,r}$ and $M_r(j)$ denote the output law and mode of the Poissonized learner. For each seed let $C_f(r)=\{j:M_r(j)=f\}$ be its complete modal cells. Comparing the common output $A(S;r)$ to the two modes yields
\begin{align}
\E_r\sum_f\langle\ind_{C_f(r)},(I-K^{\otimes d})\ind_{C_f(r)}\rangle
&\le2\bar\rho,\label{eq:modal-crossing}\\
\Pr_{J,r}[L_J(M_r(J))>\eps]&\le\bar\delta+\bar\rho.
\label{eq:modal-accuracy}
\end{align}
The inner product uses the uniform target prior. The first bound concerns complete cells, not their accurate subsets. The second bound will control the mass assigned to large cells.

\subsection{Comparison with a compressed Neumann resolvent}
\begin{lemma}[Finite posterior-kernel comparison]\label{lem:grid-kernel}
Identify functions on $\{0,\ldots,k\}^d$ with functions constant on the corresponding equal-volume cells of $[0,1]^d$, then apply the coordinatewise Gaussian quantile map. Set
\[
A=\max\{1,2(k+1)(e^\beta-1)\},\qquad \tau=A^{-2}.
\]
For every such function $f$,
\begin{equation}\label{eq:grid-operator}
\langle f,(I-K^{\otimes d})f\rangle
\ge\frac1{26}\langle f,(I-\mathsf O_\tau)f\rangle,
\end{equation}
where $\mathsf O_\tau$ is the Gaussian Ornstein--Uhlenbeck operator with correlation $e^{-\tau}$.
\end{lemma}
\begin{proof}
\textit{Entry bounds.} For states $u\le v$ at distance $r=v-u$, their internal cuts are absent with probability $t^r$. Conditional on this event, the left and right extensions $L,R$ of their common interval are independent truncated geometric variables with means at most $1/z$. Jensen gives
\begin{equation}\label{eq:grid-entry}
K(u,v)\ge\frac{e^{-\beta r}}{r+1+2/z}.
\end{equation}
The event that all cuts are absent also gives $K(u,v)\ge e^{-\beta k}/(k+1)$.

Put $K_0=k+1$ and $a=\max\{K_0^{-1},2z\}$. On $[0,K_0]$ the Neumann resolvent $(I-\partial_x^2/a^2)^{-1}$ has kernel
\[
r_a(x,y)=\frac{a\cosh(a\min(x,y))\cosh(a(K_0-\max(x,y)))}{\sinh(aK_0)}.
\]
It satisfies the zero-derivative endpoint conditions and the derivative jump of the Green equation, which verifies the formula. Its integral in $y$ is one. Compress it to unit cells $I_i=[i,i+1]$ by
$\bar R_{ij}=\int_{I_i}\int_{I_j}r_a(x,y)\,dx\,dy$.
This matrix is symmetric and stochastic.

Since $aK_0\ge1$, $r_a(x,y)\le3ae^{-a|x-y|}$. For $i<j$ with $r=j-i$,
\[
\bar R_{ij}\le\frac{3(1-e^{-a})^2}{a}e^{-a(r-1)}.
\]
If $a=2z$, the inequality $1-e^{-2z}\le2z/(1+z)$ yields
$\bar R_{ij}\le6z(1+z)^{-2}e^{-2z(r-1)}$.
For $0<z<1$ that elementary inequality follows from
$\log((1+z)/(1-z))\ge2z$; for $z\ge1$ its right side is at least one.
The lower bound in \eqref{eq:grid-entry} at distance $r$ equals
\[
\frac{z}{2(1+z)^2}\,
\frac{e^{-\beta(r-1)}}{1+z(r-1)/(2(1+z))}.
\]
Using $\beta=\log(1+z)\le z$ and $1+u\le e^u$ gives $\bar R_{ij}\le12K(i,j)$.
If $a=K_0^{-1}$, then $2zK_0\le1$, so
$K(i,j)\ge e^{-zk}/K_0\ge1/(2K_0)$, whereas
$r_a(x,y)\le\coth(1)/K_0<2/K_0$. The same factor $12$ therefore applies in both cases.

\textit{Quadratic forms and quantiles.} For symmetric Markov kernels,
\[
\langle f,(I-K)f\rangle=\frac12\sum_{i,j}\pi_iK(i,j)(f(i)-f(j))^2,
\]
where $\pi_i=1/K_0$. The off-diagonal entry comparison gives
$I-K\succeq(I-\bar R)/12$.
Scale to $[0,1]$, and let $P$ be its equal-cell averaging projection with range $V$. Then
$\bar R=P(I+\mathcal L_N/A^2)^{-1}P|_V$, where $\mathcal L_N=-\partial_x^2$ has Neumann boundary conditions.

Let $\Phi,\phi$ be the Gaussian distribution function and density. The unitary map $Uf=f\circ\Phi$ satisfies, first on smooth functions and then by closure,
\[
\mathcal E_G(Uf)=\int_0^1\phi(\Phi^{-1}(u))^2|f'(u)|^2\,du
\le(2\pi)^{-1}\mathcal E_N(f).
\]
The Neumann form domain is contained in the transformed Gaussian form domain. The variational identity
\[
\langle g,(I+\mathcal L)^{-1}g\rangle
=\sup_f\{2\langle g,f\rangle-\|f\|^2-\mathcal E(f)\}
\]
therefore proves
$(I+\mathcal L_N/A^2)^{-1}\preceq(I+A^{-2}\mathcal L_G)^{-1}$.
Define $B=\frac{11}{12}I+\frac1{12}(I+\tau\mathcal L_G)^{-1}$. On $V$, $0\preceq K\preceq PBP$. Tensor products preserve this order: for two factors the difference is $(PBP-K)\otimes PBP+K\otimes(PBP-K)$, a sum of positive operators, and induction handles all factors. For $f\in V^{\otimes d}$, the compression disappears inside the quadratic form, giving
$\langle f,K^{\otimes d}f\rangle\le\langle f,B^{\otimes d}f\rangle$.

\textit{The full-dimensional spectral bound.} In the Hermite basis, write $x_i=\tau j_i\ge0$ and $X=\sum_i x_i$. The eigenvalue of $B^{\otimes d}$ is $\prod_i(1-x_i/(12(1+x_i)))$. Since $\sum_i x_i/(1+x_i)\ge X/(1+X)$,
\[
1-\prod_i\left(1-\frac{x_i}{12(1+x_i)}\right)
\ge1-e^{-X/(12(1+X))}
\ge\frac1{13}\frac X{1+X}
\ge\frac1{26}(1-e^{-X}).
\]
The middle inequality follows from $1-e^{-u/12}\ge u/13$ for $u\in[0,1]$; for the last, split at $X=1$. The OU eigenvalue is $e^{-X}$. This proves \eqref{eq:grid-operator} for every cell-constant $f$.
\end{proof}

\subsection{Small cells cross with sufficient probability}
\begin{lemma}[Gaussian crossing estimate]\label{lem:gaussian}
For a Gaussian set of mass $p=\Phi(-u)$ with $u\ge1$ and $0<\tau\le1$,
\[
\langle\ind_C,(I-\mathsf O_\tau)\ind_C\rangle
\ge\frac p{15}\min\{1,u\sqrt\tau\}.
\]
\end{lemma}
\begin{proof}
Borell's noise-stability inequality \cite[Theorem 1, $q=2$]{Eldan} says that a halfspace maximizes the probability that both correlated Gaussian points stay in a set of fixed measure. It suffices to bound the halfspace's probability of leaving.
Write $H=rG+\sigma Z$, where $r=e^{-\tau}$, $\sigma=\sqrt{1-r^2}\ge0.9\sqrt\tau$, and $G,Z$ are independent standard Gaussians. For $w=\min\{1/(2u),\sigma/2\}$ and $G\in[-u-w,-u]$, the conditional probability of $H>-u$ is at least $\Phi(-1/2)>3/10$. The band has probability at least $w\phi(u)/2$, since $uw+w^2/2\le5/8<\log2$. Mills' inequality $p\le\phi(u)/u$ then gives at least
$(3/20)wup\ge(p/15)\min\{1,u\sqrt\tau\}$; indeed $(3/40)0.9>1/15$.
The bound on $\sigma$ follows from $1-e^{-2\tau}\ge(1-e^{-2})\tau>0.81\tau$ on $[0,1]$.
\end{proof}

\begin{proof}[Lower bound in \Cref{thm:grid}]
First suppose $d\ge1024$. For any fixed improper output $f$, its active-coordinate error fractions
\[
Z_i(J_i)=\frac1k\sum_{x=1}^k\ind\{f(1,i,x)\ne\ind\{x\le J_i\}\}
\]
are independent, lie in $[0,1]$, and have means at least
$k^{-1}\sum_{x=1}^k\min\{x,k+1-x\}/(k+1)\ge1/4$.
The final inequality follows by summing the two arithmetic progressions on either side of the midpoint. Accuracy implies $d^{-1}\sum_iZ_i\le\eps/q\le1/8$. Hoeffding therefore gives prior accuracy mass at most $e^{-d/32}$.

Put $p_0=e^{-d/64}$. A modal cell of mass $p_f>p_0$ contains accurate mass at most $p_0^2$, and hence inaccurate mass at least $(1-p_0)p_f$. By \eqref{eq:modal-accuracy}, the average total mass of large cells is at most $(\bar\delta+\bar\rho)/(1-p_0)$. Thus small cells have average mass at least $0.99$.
For any small cell with positive mass $p$, write $p=\Phi(-u)$. Since
$p\ge\int_u^{u+1}\phi(v)\,dv\ge\phi(u+1)$ and $p\le e^{-d/64}$,
\[
u\ge\sqrt{d/32-\log(2\pi)}-1\ge\sqrt d/8\ge1.
\]
Combine \Cref{lem:grid-kernel,lem:gaussian}. The cell's crossing energy is at least
\[
\frac p{3120}\min\{1,\sqrt d/A\},\qquad 3120=26\cdot15\cdot8.
\]
Sum over small cells and compare with \eqref{eq:modal-crossing}:
$2\bar\rho\ge\frac1{3200}\min\{1,\sqrt d/A\}$.
As $6400\bar\rho\le6528\rho<1$, we must have $A>\sqrt d>1$, and
\[
e^\beta-1\ge\frac{\sqrt d}{13056(k+1)\rho}
\ge\frac{\sqrt d}{2^{14}(k+1)\rho}.
\]
Use \eqref{eq:grid-beta} and $q\le16\eps$ to conclude
\[
n\ge\frac{dk}{128\eps}\log\left(1+\frac{\sqrt d}{2^{14}(k+1)\rho}\right)
\ge\frac{2^{-22}dk}{\eps}\log\left(1+\frac{\sqrt d}{k\rho}\right).
\]
The second step uses $k+1\le2k$ and $\log(1+x/c)\ge c^{-1}\log(1+x)$ for $c\ge1$.

For $d<1024$, restrict to the diagonal chain $(j,\ldots,j)$, $j=0,\ldots,k$. Apply \Cref{lem:chain} with $J=k$ and $s=q$ to get
$n\ge k\log(1+1/(k\rho))/(2048\eps)$. Since
$\log(1+\sqrt d x)\le\sqrt d\log(1+x)$ and $d^{3/2}<2^{15}$, this is at least the claimed $2^{-26}$ multiple of the main term.
Finally the two diagonal endpoints have distance $q$ and identical samples whenever the active layer is missed. Thus $2\delta\ge(1-q)^n$, which gives $n\ge\log(1/(2\delta))/(32\eps)$. Both bounds hold on the same instance.
\end{proof}

\subsection{The matching finite-grid learner}
\begin{proof}[Upper bound in \Cref{thm:grid}]
Let $a_0=q/(dk)$ be the mass of an active atom and $\mu=Na_0$. For the first block choose the maximum-positive anchor $g_i$. It never exceeds the true threshold $j_i^*$. Put $r_i=a_0(j_i^*-g_i)$ and $r=\sum_i r_i=L(g)$. A gap of at least $\ell$ requires the last $\ell$ positive atoms to be unobserved, so
\begin{equation}\label{eq:grid-anchor-mean}
\E r\le da_0\sum_{\ell\ge1}e^{-Na_0\ell}
=\frac q{k(e^\mu-1)}.
\end{equation}
For any subset $I$ of coordinates and $u_i>0$, the event $r_i>u_i$ for all $i\in I$ requires avoidance of fixed disjoint terminal intervals of combined mass at least $\sum_{i\in I}u_i$. Hence its probability is at most $e^{-N\sum_{i\in I}u_i}$. Expand the product of
$e^{\theta r_i}=1+\int_0^{r_i}\theta e^{\theta u}\,du$ and integrate each subset term. For $0<\theta<N$ this gives
\[
\E e^{\theta r}\le\sum_{I\subseteq[d]}\left(\frac\theta{N-\theta}\right)^{|I|}
=\left(\frac N{N-\theta}\right)^d.
\]
At $\theta=N/2$, $\Pr[r>\eps/2]\le2^de^{-N\eps/4}\le\delta$. This proof works for fixed-size multinomial samples; it does not require coordinate independence.

Take $\lambda=16d/\eps$. The reference $Q(j)\propto e^{-\lambda L(h_j)}$ factors into truncated two-sided geometric distributions with ratios $e^{-\lambda a_0}$. The expected distance in an untruncated two-sided geometric law of parameter $u=\lambda a_0$ is
$2\sum_{\ell\ge1}\ell e^{-u\ell}/(1+2\sum_{\ell\ge1}e^{-u\ell})=1/\sinh u\le1/u$.
Truncation at either endpoint decreases this mean: relative to the infinite distance law, its multiplicity ratio is nonincreasing in distance. Therefore $\E_Q L\le d/\lambda=\eps/16$, and $Q\{L>\eps/8\}\le1/2$.

Apply the correction, clipping, and clocks of \Cref{lem:control,lem:clip}. Their disagreement bound is
\begin{equation}\label{eq:grid-collision}
4\sqrt2\lambda\sqrt{\E r/N}+32\E r/\eps.
\end{equation}
Write $x=\sqrt d/(k\rho)$ and $C_0=4096$. The sample budget ensures $\mu\ge C_0\log(1+x)$ and
\[
e^\mu-1\ge C_0x,\qquad \mu(e^\mu-1)\ge C_0^2x^2.
\]
The first is Bernoulli's inequality. For the second, $\log(1+x)\ge2x/(x+2)$ gives
$2\log(1+x)((1+x)^2-1)\ge4x^2$; scale by $C_0/2\ge1$ using $e^{tu}-1\ge t(e^u-1)$.
Substitute \eqref{eq:grid-anchor-mean} into \eqref{eq:grid-collision}, use $N=\mu dk/q$ and $q/\eps\le16$. The two terms are at most $\sqrt2\rho/4$ and $\rho/(8\sqrt d)$, respectively. Their sum is less than $\rho$.
The required condition $\lambda/N\le1/4$ follows from
\[
\lambda/N\le[16k\log(1+\sqrt d/(k\rho))]^{-1}
\le(16\log2)^{-1}<1/4.
\]
Here $k\log(1+y/k)\ge\log(1+y)$ for integer $k\ge1$, and $y=\sqrt d/\rho\ge1$.
Every output lies within $\eps/2$ of $g$, so the earlier tail bound for $r$ supplies accuracy. Since $dk\log(1+\sqrt d/(k\rho))\ge d\log2$, the upper and lower formulas give \eqref{eq:grid-rate}.
\end{proof}

\subsection{Consequences for cardinality and VC dimension}\label{sec:grid-consequences}
The envelopes \eqref{eq:envelope-lower}--\eqref{eq:envelope-upper} follow by maximizing the grid lower bound over $(k+1)^d\le M$ and taking the better of \Cref{thm:k,thm:det}. Maxima and sums differ by at most a factor two for the nonnegative lower-bound terms.
For fixed parameters, choose $k=\lceil\sqrt d/\rho\rceil$. Then
$dk\log(1+\sqrt d/(k\rho))\ge d^{3/2}/(2\rho)$ and
$\log|H_{d,k}|=O(d\log d)$. For every sufficiently large $M$, choose
$d=\lfloor c\log M/\log\log M\rfloor$ with a fixed sufficiently small $c>0$. The resulting class fits within $M$ and gives
$N^{K\text{-unif},*}(M)=\Omega((\log M/\log\log M)^{3/2})$.
Taking logarithms, dividing by $\log\log M$, and combining with \Cref{thm:k}'s upper bound proves both exponent limits there.

The grid class has VC dimension exactly $d$: one active point in each coordinate can be shattered, whereas any larger set includes either an inactive point or two points in the same threshold coordinate. The latter cannot receive the label pattern that decreases across a threshold. Taking $d=v$ and the same $k$ gives
\begin{equation}\label{eq:vc-lower}
\sup_{\substack{H\text{ finite}\,;\,\VC(H)\le v}}\sup_{p\text{ uniform}}N_{H,p}^K
\ge\max\left\{\frac{2^{-27}v^{3/2}}{\eps\rho},
\frac{\log(1/(2\delta))}{32\eps}\right\}.
\end{equation}
This supplies the lower bound invoked by \Cref{thm:vc}.

It also establishes the inverse-replicability exponent stated after \Cref{thm:k}. Fix $\eps,\delta$ in \eqref{eq:domain}, take $d=1$ and $k=\lceil1/\rho\rceil$, and let $\rho\downarrow0$. The grid lower bound is $\Omega(1/(\eps\rho))$, whereas $\log|H_{1,k}|=O(\log(1/\rho))$. For any fixed $b<1$ and any fixed polynomial $P$, $P(\log(1/\rho))\rho^{-b}=o(\rho^{-1})$. Thus a uniform known-marginal guarantee of the form $P(\log|H|)\rho^{-b}$ is impossible. The upper bound in \Cref{thm:k} attains exponent one. The family varies with $\rho$ in this argument; for a fixed finite class the saturation theorems describe a different limit.

\subsection{The continuous posterior route and its finite-uniform consequence}\label{sec:continuous-kernel}
A continuous posterior kernel gives a complementary lower-bound construction. We give the kernel comparison and its explicit constants below. The exact discrete theorem above removes the need for a very fine grid.

For a uniform threshold $u\in[0,1]$, a Poisson process of rate $\lambda$ cuts the interval. Let $K_\lambda$ average over the random interval containing $u$. It is an average of orthogonal projections, hence $0\preceq K_\lambda\preceq I$. If $u\le v$ and $z=v-u$, the two points have no intervening cut with probability $e^{-\lambda z}$, and the two outside extensions have means at most $1/\lambda$. Therefore
\[
K_\lambda(u,v)\ge\frac{\lambda e^{-\lambda z}}{2+\lambda z}
\ge\frac\lambda2 e^{-3\lambda z/2}.
\]
For $\lambda\ge1$, the Neumann resolvent with $a=3\lambda/2$ has kernel
$a\cosh(a\min(u,v))\cosh(a(1-\max(u,v)))/\sinh a
<4\lambda e^{-3\lambda|u-v|/2}$.
The preceding entry bound gives $I-K_\lambda\succeq(I-R_a)/8$.
The quantile form comparison in \Cref{lem:grid-kernel} gives
$R_a\preceq(I+\lambda^{-2}\mathcal L_G)^{-1}$, since $2\pi>9/4$.
Tensorize the positive operator
$B=7I/8+(I+\lambda^{-2}\mathcal L_G)^{-1}/8$.
Its Hermite eigenvalues satisfy
\[
1-\prod_i\left(1-\frac{x_i}{8(1+x_i)}\right)
\ge\frac19\frac X{1+X}\ge\frac1{18}(1-e^{-X}),\qquad X=\sum_i x_i.
\]
If $\lambda<1$, add independent cuts until rate one; the finer averaging projection is larger in positive-operator order. Thus in all cases
\begin{equation}\label{eq:continuous-operator}
I-K_\lambda^{\otimes d}\succeq\tfrac1{18}(I-\mathsf O_\tau),
\qquad \tau=\max\{1,\lambda\}^{-2}.
\end{equation}

\begin{proposition}[Explicit fine-grid joint lower bound]\label{prop:continuous}
For $d\ge64$ and \eqref{eq:domain}, take $k=\lceil2^{14}d^{3/2}/\rho^2\rceil$ in $H_{d,k}$. Every valid learner satisfies
\[
n\ge\max\left\{\frac{2^{-20}d^{3/2}}{\eps\rho},
\frac{\log(1/(2\delta))}{32\eps}\right\}.
\]
\end{proposition}
\begin{proof}
The small-step part of \Cref{lem:chain}, before its final coarsening, on the diagonal chain gives
$n\ge\min\{k/(4q),3(1-8\delta)/(64q\rho)\}\ge3/(128q\rho)\ge3/(64\rho)$.
In particular $e^{-n/4}\le\rho/100$ for $\rho\le10^{-4}$; taking logarithms verifies this at the endpoint, and the ratio decreases as $\rho$ decreases.

Replace the third coordinate by $[0,1]$ and map it to $\max\{1,\lceil kx\rceil\}$. Extend each finite output constantly on these intervals. For continuous target $t$, compare to $\lfloor kt\rfloor/k$. One observation changes label with probability at most $q/k$, so $n$ observations cost $\eta=nq/k$. If $n\ge d^{3/2}/(\eps\rho)$ the conclusion already holds; otherwise $\eta\le\rho/1024$. The continuous experiment has accuracy $\eps'=\eps+q/k$, failure $\delta+\eta$, and replication error $\rho+2\eta$.
Poissonize with mean $2n$, taking the first $n$ samples when possible. Shortage is at most $e^{-n/4}$. The resulting parameters obey $\bar\rho<1.025\rho$ and $\bar\delta\le\delta+(1/1024+1/100)\rho$. Each coordinate has cut rate $\lambda=2nq/d$.

The posterior-resampling and modal-cell argument leading to \eqref{eq:modal-crossing} applies verbatim, with uniform prior on $[0,1]^d$. For any output $f$, the accurate parameter set $A_f$ is compact: its risk changes by at most $(q/d)\|t-s\|_1$. If $t,s\in A_f$, then $\|t-s\|_1\le2\alpha d$, where $\alpha=\eps'/q\le1/7$. Brunn--Minkowski from \Cref{sec:geometry} and the $\ell_1$-ball volume give
\[
2^d\vol(A_f)\le\vol(A_f-A_f)\le\frac{2^d(2\alpha d)^d}{d!},
\qquad \vol(A_f)\le(2e/7)^d<e^{-d/4}.
\]
Thus complete cells of mass at most $e^{-d/8}$ carry at least $0.99$ average mass. For such a mass $p=\Phi(-u)$, the same Gaussian-tail argument as above gives $u\ge\sqrt d/4$. A weaker form of \Cref{lem:gaussian}, with constant $1/50$, and \eqref{eq:continuous-operator} give crossing at least
$p\min\{1,\sqrt d/\max(1,\lambda)\}/3600$.
Consequently $2\bar\rho\ge\min\{1,\sqrt d/\max(1,\lambda)\}/4000$. Since $8200\rho<1$, we obtain $\lambda\ge\sqrt d/(8200\rho)$, hence
$n\ge d^{3/2}/(262400\eps\rho)\ge2^{-20}d^{3/2}/(\eps\rho)$.
The finite-grid endpoint confidence bound is unchanged.
\end{proof}

For an explicit cardinality translation, set $L=\log M$,
$B_M=1+\log(L+e)+\log(1/\rho)$ and $D=\lfloor L/(8B_M)\rfloor$.
If $D\ge64$, choose $d=D$ in \Cref{prop:continuous}. The bound
$\log(k+1)\le15\log2+\frac32\log D+2\log(1/\rho)\le8B_M$
ensures $(k+1)^D\le M$, while $D\ge L/(16B_M)$. Thus
\[
N^{K\text{-unif},*}(M)\ge\max\left\{
\frac{2^{-26}}{\eps\rho}\left(\frac{\log M}{1+\log(\log M+e)+\log(1/\rho)}\right)^{3/2},
\frac{\log(1/(2\delta))}{32\eps}\right\}.
\]
The dimension condition is part of this explicit consequence; the grid envelope in \eqref{eq:envelope-lower} is available for all $M\ge2$.

\section{VC-sensitive anchors and shared-phase trees}\label{sec:vc-proof}
This appendix proves the two guarantees attached to the covering and tree modules in \Cref{sec:module-vc}. We first construct the anchor required by \Cref{thm:vc}. We then give a separate tree construction for \Cref{thm:vc1}. Neither construction uses the other one's upper bound.

\subsection{An expected-risk anchor}
\begin{lemma}[One-inclusion expectation bound]\label{lem:one-inclusion}
For a finite Boolean class of VC dimension $v$, there is an order-invariant, possibly improper predictor trained on $N$ examples with expected realizable risk at most $v/(N+1)$.
\end{lemma}
\begin{proof}
For a finite restriction $C\subseteq\{0,1\}^{t}$, join strings that differ in one coordinate. Its number of edges is at most $v|C|$. Here is the usual induction, included to specify the combinatorial input. Split at the last coordinate and write $U$ and $I$ for the union and intersection of the two projections. The edges internal to the projections are bounded by the edges of $U$ plus those of $I$. There are $|I|$ edges in the last coordinate. Moreover, $\VC(U)\le v$ and $\VC(I)\le v-1$: a set shattered by $I$, together with the last coordinate, is shattered by $C$. Induction therefore gives
\[
|E(C)|\le v|U|+(v-1)|I|+|I|=v|C|.
\]
The argument applies to every vertex subset. Assign each edge to one of its endpoints with capacity $v$ per vertex. Hall's condition holds, since any collection of edges has at least $|E|/v$ incident vertices by the preceding density bound. Direct every edge away from its assigned endpoint. The maximum outdegree is at most $v$.

Fix an ordering of the evaluation patterns of $H$ and a deterministic choice of such an orientation for each restriction. These choices depend on the set of instance patterns, not on sample order. For a realizable sample and a query already represented in it, return the observed label. Otherwise consider the extensions of the observed labeling to the query. If there is only one, return it. If there are two, return the label at the head of their one-inclusion edge. On a nonrealizable sample return the constant-zero predictor; this makes the definition total and order-invariant. Points with the same evaluation pattern are treated as one point. Since $H$ is finite, only finitely many patterns occur, so all choices can be made measurably in the finite evaluation partition.

On any realizable multiset of $N+1$ labeled examples, a leave-one-out error is impossible at a repeated pattern. At a uniquely occurring pattern it corresponds to an outgoing edge of the true restriction in the graph on the full set of patterns. There are at most $v$ such edges. Averaging the $N+1$ leave-one-out errors and then using exchangeability proves the claimed risk bound. This is the expectation guarantee of the one-inclusion construction \cite{HLW}.
\end{proof}

\begin{lemma}[Independent validation]\label{lem:validation}
Let two predictors, independent of a validation sample of size $N$, have risks $r_0,r_1$. The empirically better predictor $g$, with a fixed tie rule, satisfies, conditionally on the candidates,
\[
\Pr\{L(g)>4\min(r_0,r_1)+4t/N\}\le3e^{-t},\qquad
\E L(g)\le4\min(r_0,r_1)+9/N.
\]
\end{lemma}
\begin{proof}
For a binomial mean $\widehat r$, the exponential moment identity
$\E e^{uN\widehat r}=(1-r+re^u)^N$ and $1+x\le e^x$ give
\[
\Pr\{\widehat r>2r+t/N\}\le e^{-t},\quad
\Pr\{r>2\widehat r+2t/N\}\le e^{-t}.
\]
For the second inequality use $u=-1$ and $1-e^{-1}>1/2$; for the first use $u=1$ and $e-1<2$. Apply the first inequality to the better true-risk candidate and the second to both candidates. The selection rule then gives the stated tail bound. Integrating $\min\{1,3e^{-t}\}$ over $t\ge0$ gives $1+\log3$, hence an additive $4(1+\log3)/N<9/N$ in expectation.
\end{proof}

\begin{proof}[Proof of \Cref{thm:vc}]
Let $\eta=\eps/256$ and $t=\log(6(q+1)/\delta)$. The first block produces $g_0$ with $\E L(g_0)\le v/(N+1)$. On the second block choose an empirical-risk minimizer $f_1$ in $F$. Some $f_*\in F$ has risk at most $\eta$. The two binomial inequalities just proved, used for $f_*$ and all $q$ candidates, imply
\[
\Pr\{L(f_1)>4\eta+4t/N\}\le(q+1)e^{-t}=\delta/6.
\]
The third block selects $g$ between $g_0$ and $f_1$. Since $q\ge2$ implies $v\ge1$, \Cref{lem:validation} gives
\[
\E L(g)\le4v/(N+1)+9/N\le16v/N.
\]
Also $N\ge512t/\eps$ gives $L(f_1)\le3\eps/128$ outside the preceding exceptional event. A second application of the validation tail bound yields
\[
L(g)\le4(3\eps/128)+4t/N\le13\eps/128<\eps/8
\]
except on an event of probability at most $\delta/6+3e^{-t}<\delta$.

Write $r=L(g)$ and let $Q(h)\propto e^{-\lambda L(h)}$ on $F$, where $\lambda=16\log(2q)/\eps$. This reference distribution is independent of the sample. Since $\min_F L\le\eta$,
\[
Q\{L>\eps/8\}\le q\exp[-\lambda(\eps/8-\eta)]
=q(2q)^{-31/16}\le\tfrac12.
\]
The chosen sample budget gives $\lambda/N\le\rho/(64\sqrt v)\le1/4$. Apply \Cref{lem:control} to the fourth block, conditionally on $g$, and then Jensen's inequality. The unclipped empirical law $\widetilde Q$ obeys
\[
\E\TV(\widetilde Q,Q)\le\lambda\sqrt{\E r/(2N)}
\le2\sqrt2\lambda\sqrt v/N.
\]
The clipping estimate \Cref{lem:clip} applies also when $g$ is improper and the candidates are in $F$. If the clipped set is empty use the fixed fallback in $F$. Its two-sample disagreement bound is
\[
8\E\TV(\widetilde Q,Q)+32\E r/\eps
\le\frac{256\sqrt2\sqrt v\log(2q)+512v}{N\eps}\le\rho/2.
\]
On $r\le\eps/8$, a cover center of the target has distance at most $\eta+r<\eps/4$ from $g$, so clipping is nonempty for every chosen radius. Every possible clipped output has risk at most $\eps/2+r<\eps$. Thus the accuracy failure probability is below $\delta$, and the replication guarantee follows from the displayed bound.

For completeness we establish the required cover size. A maximal $\eta$-separated subset is a cover. If it has size $q\ge2$, take
$t_0=\lceil(2\log q+1)/\eta\rceil$ independent instances. The chance that some pair is not separated by the sample is at most $\binom q2e^{-t_0\eta}<1$. Some restriction therefore has $q$ distinct strings. Sauer's inequality, obtained by the same union--intersection induction as above, gives $q\le\sum_{j=0}^v\binom{t_0}j$. If $t_0<v$, then $\log q\le v\log2$. Otherwise put $s=\log q/v$ and use $\sum_{j=0}^v\binom{t_0}j\le(et_0/v)^v$ to obtain
\[
e^s\le(2e/\eta)(s+1).
\]
Here $t_0/v\le2(s+1)/\eta$. The function $e^s/(s+1)$ is increasing for $s\ge0$; at $s=2\log(8e/\eta)$ it exceeds $2e/\eta$. Hence $\log q\le2v\log(8e/\eta)$. Substitution proves the all-accuracy bound. The fixed-accuracy lower bound is \eqref{eq:vc-lower}, together with the two-target confidence lower bound in \Cref{lem:chain}. The product threshold class has VC dimension $d$: one can shatter one point in each chain, while two points in the same chain cannot be shattered. This verifies the class restriction required for the minimax conclusion.
\end{proof}

\subsection{A laminar representation for VC dimension one}\label{sec:vc1-proof}
We now prove the guarantee for the shared-phase tree module, \Cref{thm:vc1}. Fix $h_0\in H$ and put $A_x=\{h:h(x)\ne h_0(x)\}$. If $A_x,A_y$ intersect but neither contains the other, hypotheses in the intersection and in the two set differences, together with $h_0$, realize all four labelings of $x,y$. Thus the nonempty sets $A_x$ are laminar.

Build their inclusion tree, adjoining the root $H$ and every singleton. There are at most $2m-1$ vertices and $2m-2$ edges, since each internal vertex has at least two children. Assign to the edge entering a set $A$ the weight $p\{x:A_x=A\}$; an added set with no such instances has weight zero. The total edge weight is at most one. A vertex $u$ represents the Boolean function obtained from $h_0$ by flipping the labels whose support edge lies on the root-to-$u$ path. The singleton leaves represent exactly $H$. For vertices $u,v$, their $d_p$ distance equals their weighted tree distance.

Let $\sigma=\eps/(8m)$ and replace every edge of length below $\sigma$ by length zero. Write $\bar d$ for the resulting path pseudometric and $\Delta$ for the total removed length. Then
\begin{equation}\label{eq:tree-distances}
\Delta<\eps/4,\qquad \bar d(u,v)\le d_p(f_u,f_v)\le\bar d(u,v)+\Delta.
\end{equation}
Vertices and their fixed labels remain available even when their new distance is zero.

Given a realizable sample, select the deepest retained edge whose associated disagreement with $h_0$ was observed, and let $g$ be its lower endpoint. If no retained edge was witnessed, use the root. For target leaf $t$, all such edges lie on its root-to-leaf path, so $g$ is an ancestor of $t$. Put $r=\bar d(g,t)$. Enumerate the retained edges upward from $t$, with lengths $a_j\ge\sigma$ and partial sums $T_j$. The event that the anchor misses the first $j$ such edges is exactly the event that no sample hits their union, of mass $T_j$. Therefore
\begin{align}
\E r&=\sum_j a_j(1-T_j)^N
\le\sum_j a_je^{-NT_j}
\le\frac{\sigma}{e^{N\sigma}-1},\label{eq:tree-mean}\\
\Pr\{r>u\}&\le(1-u)^N\le e^{-Nu}.\label{eq:tree-tail}
\end{align}
For the last inequality in \eqref{eq:tree-mean}, the function $a/(e^{Na}-1)$ is decreasing. Thus
$a_je^{-NT_j}\le[\sigma/(e^{N\sigma}-1)](e^{-NT_{j-1}}-e^{-NT_j})$,
and the sum telescopes. To see \eqref{eq:tree-tail}, take the first partial sum exceeding $u$; the corresponding edges must all be unobserved if $r>u$.

\begin{proof}[Proof of \Cref{thm:vc1}]
Put $w=\eps/16$. A phase $R\in[0,w)$ removes every edge whose endpoint depths fall in different intervals $[R+jw,R+(j+1)w)$. All depths use $\bar d$ from the root. In each remaining component choose one fixed vertex $v_C$, and fix a nearest original leaf $\Pi(v_C)$ in $\bar d$, resolving ties in advance. Return $\Pi(v_{C(g)})$.

Every component has $\bar d$-diameter at most $2w$: the depth range has width at most $w$, and a path between two vertices goes through their common ancestor in that component. Consequently
\[
\bar d(\Pi(v_C),t)\le2\bar d(v_C,t)\le2r+4w.
\]
Together with \eqref{eq:tree-distances}, this bounds the true error by $2r+\eps/2$. By \eqref{eq:tree-tail}, accuracy fails with probability at most $e^{-N\eps/4}$, for every phase separately.

The two anchors lie on the same target path. Under a continuous uniform phase, a cut separates them with probability at most $\min\{1,\bar d(g,g')/w\}$. Their distance is at most $r+r'$. Hence their outputs disagree with probability at most
\[
\frac{2\E r}{w}\le\frac4{m(e^{N\eps/(8m)}-1)}.
\]
Use instead the $K$ equally spaced phases $wJ/K$, where $K=2^{\lceil\log_2(2/\rho)\rceil}$. The fraction of these phases in any half-open circular interval of length $a$ is at most $a/w+1/K$. This adds at most $1/K\le\rho/2$ to disagreement, including intervals whose endpoints are grid points. The stated $N$ makes the continuous term at most $\rho/2$ and the accuracy failure at most $\delta$.

For the upper minimax bound use that $m\log(1+c/m)$ is increasing in $m$ and $\log(1+8x)\le8\log(1+x)$. For the lower bound use the chain with $k=M-1$ in \Cref{lem:chain}. Its rate is comparable to $M\log(1+1/(M\rho))/\eps$, uniformly for $M\ge2$, and the same lemma supplies the confidence term. Its marginal is uniform and its VC dimension is one. Computing the support sets, the inclusion tree, their rational weights, nearest leaves, and the phase components takes polynomial time on explicit finite input. Exact arithmetic suffices for each comparison.
\end{proof}

\subsection{Compact threshold implementation and rational budgets}\label{sec:compact-threshold}
For thresholds $h_t(x)=\ind\{x\le t\}$ on $[bJ]$ with uniform mass $a=1/(bJ)$, no tree enumeration is needed. Let $g$ be the largest observed positive point, or zero. With $w=\eps/16$, a shared phase $R$, and $\ell=\lfloor(ag-R)/w\rfloor$, return
\[
\widehat t=\max\{0,\lceil(R+\ell w)/a\rceil\}.
\]
Then $0\le\widehat t\le g$ and $a(g-\widehat t)\le w$. Equal phase cells give identical outputs. The preceding anchor-tail and phase-interval arguments apply, with arithmetic cost polynomial in the sample size and the bit length of $bJ$, rather than in $bJ$ itself.

All logarithmic sample budgets can be implemented with rational upper bounds. For rational $x>0$ define $U(x)=x$ when $x\le1$, and $U(x)=\lceil\log_2(1+x)\rceil$ otherwise. Integer bit comparisons compute the latter value exactly. Since
$\log(1+x)\le U(x)\le3\log(1+x)$, replacing each occurrence of $\log(1+x)$ by $U(x)$ and then taking an integer ceiling preserves every guarantee up to an absolute constant. This avoids a rounding error in numerical evaluation of an exact sample budget.

\section{Fixed-block learning: construction and matching obstruction}\label{sec:locality-proof}
This appendix proves \Cref{thm:block}. The construction uses the deletion estimate from \Cref{lem:power}. The lower bound has two additional ingredients: a local collision inequality and a covariance estimate that keeps the fixed sample budget intact.

\subsection{Finite-bit block construction}
\begin{proof}[Upper bound in \Cref{thm:block}]
Set $g=\lceil d/s\rceil$, $k_0=4s$, and
\[
n=\left\lceil\max\left\{\frac{128k_0^2g^3}{\eps\rho^2},
\frac{8(d\log2+\log(1/\delta))}{\eps}\right\}\right\rceil,
\qquad B=\lfloor n\eps/(2g)\rfloor.
\]
In a block of size $s_j\le s$, enumerate $m_j=2^{s_j}$ label vectors. Give vector $v$ weight $(B-C_j(v))_+^{k_0}$. A locally consistent vector has positive weight. The hard cutoff gives total empirical errors below $gB\le n\eps/2$. For any fixed vector of true error above $\eps$, the binomial lower tail bounds the probability of at most $n\eps/2$ errors by $e^{-n\eps/8}$. A union bound over $2^d$ vectors proves accuracy for every sampling seed and every fallback described below.

The proof of \Cref{lem:power} applies to a block, treating all observations outside it as zero contributions to its cost. Since $k_0\ge2\log m_j$, common rejection sampling of the normalized weights has expected disagreement at most $\sqrt2 e k_0/\sqrt B$. Summing over the $g$ blocks gives $\sqrt2 e k_0g/\sqrt B$. The sample budget implies
$B\ge63k_0^2g^2/\rho^2$, so this is less than $0.485\rho$.

Here is a bounded finite-bit implementation. A trial chooses a uniform block vector and an independent uniform acceptance number, accepting with its normalized probability. Its acceptance probability is $1/m_j$. Cap the trials at
$T_j=\lceil m_j\log(16g/\rho)\rceil$, so the failure probability is at most $\rho/(16g)$. Allocate every block and trial a fixed segment of the public tape. Choose $L$ with $2^L\ge16g\max_jT_j/\rho$. Replace the acceptance number by a uniform integer in $\{0,\ldots,2^L-1\}$ and accept below the integer floor of $2^L$ times the probability. Under the natural continuous coupling, each decision changes with probability at most $2^{-L}$, so quantization changes some trial in some block with probability at most $\rho/16$. After an unsuccessful cap return the first locally consistent vector. One execution differs from its ideal version with probability at most $\rho/8$; two executions add at most $\rho/4$. Total disagreement is therefore below $\rho$.

All nonzero weights are integers of bit length $O(k_0\log B)$. Enumerating them, normalizing by integer sums, and comparing the rational probabilities take polynomial bit complexity per enumerated vector and trial. Since $g\le2d/s$, the stated sample and time bounds follow. This implementation also covers a final block smaller than $s$.
\end{proof}

\subsection{Local accuracy forces collisions}
For the next lemma $B_s$ is the unit Euclidean ball in $\R^s$, with normalized volume measure. A distribution $Q_t$ is a vector indexed by $v\in\{0,1\}^s$. Put
$\ell_v(t)=\sum_i t_i^2\ind\{v_i\ne\ind\{t_i>0\}\}$.

\begin{lemma}[Local collision inequality]\label{lem:local-collision}
Suppose the output law satisfies
\[
\TV(Q_t,Q_u)\le L\|t-u\|_2,\qquad
\int_{B_s}\sum_vQ_t(v)\ell_v(t)\,dt\le1/8.
\]
Then
\[
\int_{B_s}\left(1-\sum_vQ_t(v)^2\right)dt
\ge c_0\min\{1,s/L\},\qquad c_0=2^{-35}.
\]
The assumptions cannot hold with $L=0$.
\end{lemma}
\begin{proof}
For a fixed $v$, symmetry gives $\int\ell_v=s/[2(s+2)]\ge1/6$. We also need a small-volume estimate. Write $t=G/\sqrt{\|G\|^2+Z}$, where the coordinates of $G$ are independent standard normals and $Z$ is independent chi-square with two degrees of freedom. The condition $\ell_v(t)\le1/4$ is equivalent to
$\sum_i(4\ind\{\text{wrong sign}\}-1)G_i^2\le Z$.
Exponential Markov with parameter $1/8$ gives
\[
\vol\{\ell_v\le1/4\}\le\tfrac43(1/\sqrt3+1/\sqrt7)^s
\le(49/50)^s\quad(s\ge16).
\]
Indeed $1/\sqrt3+1/\sqrt7<24/25$ and $(49/48)^{16}\ge4/3$.

Define the closed cores $F_v=\{t:Q_t(v)\ge7/8\}$, their volumes $p_v$, and $b=1-\sum_vp_v$. The cores are disjoint. Outside them, $\sum_vQ_t(v)^2\le\max_vQ_t(v)\le7/8$, so collision is at least $b/8$. If $b>1/16$ this already exceeds the asserted bound. Assume henceforth $b\le1/16$ and put $\kappa=2^{-14}$. We claim
\begin{equation}\label{eq:core-power}
\sum_vp_v^{1-1/s}\ge1-b+\kappa,
\end{equation}
where for $s=1$ a nonempty core contributes one.

First suppose $s<256$. Since $0\le\ell_v\le1$,
$1/8\ge(7/8)\int_{F_v}\ell_v\ge(7/8)(1/6-(1-p_v))$,
whence $p_v\le41/42$. For $s\ge2$,
$p_v^{-1/s}-1\ge(1-p_v)/s\ge1/(42s)$;
for $s=1$, each positive $p_v$ similarly contributes $1-p_v\ge p_v/42$.
Summing proves an increment at least $15/(16\cdot42\cdot255)>\kappa$.
For $s\ge256$, let $p_0=(49/50)^{s/2}\le1/8$. On a core of mass $p_v>p_0$, its intersection with $\{\ell_v\le1/4\}$ has mass at most $p_0^2\le p_v/8$. Therefore its contribution to expected loss is at least $(7/8)(1/4)(7p_v/8)=49p_v/256$. The total mass of such cores is at most $32/49$. The remaining positive cores have total mass at least $15/16-32/49>1/4$. On each, $p_v^{-1/s}\ge\sqrt{50/49}>101/100$. Their increment is at least $1/400>\kappa$. This proves \eqref{eq:core-power}.

For distinct cores, total variation is at least $3/4$, so their distance is at least $3/(4L)$. Take
$a=\min\{1/(8L),\kappa/(4s)\}$ and $u=sa\le\kappa/4$.
The sets $F_v+aB_s$ are disjoint and contained in $(1+a)B_s$. The compact Brunn--Minkowski inequality proved in \Cref{sec:geometry} gives
\[
(1+a)^s\ge\sum_v(p_v^{1/s}+a)^s
\ge1-b+sa\sum_vp_v^{1-1/s}.
\]
Since $(1+a)^s\le e^u\le1+u+u^2$, \eqref{eq:core-power} yields
$b(1+u)\ge\kappa u-u^2\ge3\kappa u/4$.
Thus $b\ge\kappa u/2$ and collision is at least $\kappa u/16\ge\kappa^2\min\{1,s/L\}/128=2^{-35}\min\{1,s/L\}$. If $L=0$, the law is constant and its integrated loss is at least $1/6$, contradicting the assumption.
\end{proof}

\subsection{Fixed-size blocks are dependent, but their covariances are controlled}
\begin{lemma}[Disjoint-view covariance]\label{lem:block-covariance}
Let an observation fall into disjoint blocks of masses $p_j,p_k$. The local view of a block records its internal observation or the symbol $\bot$. The maximal correlation between the two local views is at most
\[
\sqrt{\frac{p_jp_k}{(1-p_j)(1-p_k)}}.
\]
The same bound holds for any number of independent observations. Suppose block masses are at most $\gamma\le1/2$, there are $g$ blocks with $g\gamma\le1$, and $Z_j$ are local two-sample disagreement indicators, after fixing the public seed. Independent block parameters may be averaged as well. Then, writing $\mu=\E\sum_jZ_j$,
\[
\Var\left(\sum_jZ_j\right)\le2\mu,
\qquad \Pr\{\text{some }Z_j=1\}\ge\frac\mu{\mu+2}.
\]
\end{lemma}
\begin{proof}
For functions of one local view, subtract their values at $\bot$, and let $a_j$ be the conditional mean of the resulting function inside block $j$. Their covariance is $-p_jp_ka_ja_k$, because the two nonzero events are disjoint. The variance of the first is at least $p_j(1-p_j)a_j^2$, and similarly for the second. Cauchy--Schwarz proves the bound. Tensorization follows by decomposing each product $L^2$ space into tensor products of constant and mean-zero subspaces. The conditional expectation operator on each nonconstant component has norm at most the one-observation bound, and multiplying more such norms cannot increase it.

Condition on the independent block parameters. The conditional mean of $Z_j$ depends only on parameter $j$, so the covariance between conditional means for different blocks vanishes when averaging those parameters. The conditional covariance bound, Cauchy--Schwarz, and total variance give
\[
\Var\sum_jZ_j\le\left(1+\frac{(g-1)\gamma}{1-\gamma}\right)\sum_j\Var Z_j\le2\mu.
\]
Finally Cauchy--Schwarz applied to $Z=\sum_jZ_j$ gives $\Pr\{Z>0\}\ge(\E Z)^2/\E Z^2\ge\mu/(\mu+2)$.
\end{proof}

\begin{proof}[Lower bound in \Cref{thm:block}]
First take $g\ge2$ active blocks of $s$ coordinates and an additional anchor. Put $\gamma=16\eps/g$. Independently choose $t_j\sim\Unif(B_s)$, give coordinate $i$ in block $j$ mass $\gamma t_{ji}^2$ and label $\ind\{t_{ji}>0\}$, and give the remaining mass to the zero-labeled anchor. For a fixed seed, let $e_j$ be the expected normalized loss in block $j$, averaged over its parameter and sample. Accuracy implies
\[
\E_{\rm seed}\frac1g\sum_j e_j\le\frac1{16}+\delta.
\]
To justify the $\delta$ term, on the failure event normalized average loss is at most one, while on success it is at most $1/16$. For $\delta\le1/128$, the expected proportion of blocks with $e_j\le1/8$ is at least $7/16$. A random variable in $[0,1]$ with this mean is at least $1/4$ with probability at least $1/4$. Thus at least one quarter of seeds have at least $g/4$ good blocks.

The local single-observation experiment has squared Hellinger distance at most $\gamma\|t-u\|^2/(1-\gamma)\le2\gamma\|t-u\|^2$. Tensorization and $\TV\le\HH$ give local output Lipschitz constant $L=\sqrt{2n\gamma}$. Apply \Cref{lem:local-collision} on every good block. Conditional on a seed in the preceding quarter, the sum of mean local disagreements is at least
\[
a_n=\frac{c_0g}{4}\min\left\{1,s\sqrt{\frac g{32n\eps}}\right\}.
\]
The total block mass is at most $g\gamma=16\eps\le1$. \Cref{lem:block-covariance} therefore gives
$\rho\ge\tfrac14a_n/(a_n+2)$.
For $\rho\le1/8$ this implies $a_n\le16\rho$. If also $\rho\le c_0g/128$, the minimum cannot equal one. Solving for $n$ gives
\begin{equation}\label{eq:block-lower}
n\ge2^{-87}\frac{s^2g^3}{\eps\rho^2}.
\end{equation}

For a partition of $d=Gs$ coordinates with $G\ge3$, use $G-1$ blocks as active blocks and one coordinate of the remaining block as the anchor. The unused coordinates have mass zero. If $G$ is large enough, the extra condition on $\rho$ holds throughout \eqref{eq:domain}; for example $G>2^{44}$ suffices. Then $g=G-1\ge2G/3$, and \eqref{eq:block-lower} has order $d^3/(s\eps\rho^2)$. For $G\le2^{44}$, including $G=1,2$, the unrestricted anchored lower bound on $d-1$ coordinates gives
\[
n\ge\cstar\frac{(d-1)^2}{\eps\rho^2}
\ge\frac{\cstar}{4\cdot2^{44}}\frac{d^3}{s\eps\rho^2},\qquad d\ge2.
\]
The ordinary two-target rare-point argument gives $\Omega(\log(1/\delta)/\eps)$ even without replication. Taking the larger lower bound is equivalent, up to a factor two, to their sum. This completes the matching sample lower bound in the stated locality model.
\end{proof}

\section{Shared randomness on finite empirical grids}\label{sec:randomness-proof}
The purpose of this appendix is to prove \Cref{thm:seed}, including the storage cost of the construction. It also records why the finite empirical grid matters: a continuous, additive-error-free requirement would exclude atomic public seeds.

\subsection{A finite table of ball-rounding maps}
Let
\[
\mathcal X_{n,d}=\{x:x_i=\operatorname{sgn}(z_i)\sqrt{|z_i|/n},\quad z_i\in\mathbb Z,
\ \sum_i|z_i|=n\}.
\]
Its cardinality is at most $(2n+1)^d$. The signed count vector of a realizable sample belongs to this set. Points $x,y$ are sign-compatible if $x_iy_i\ge0$ for every $i$.

\begin{lemma}[Finite-grid map compression]\label{lem:finite-seeds}
For $0<\eta<1$ there are $K$ deterministic maps from $\mathcal X_{n,d}$ to $\{0,1\}^d$, where $K$ is the least power of two at least
\[
4[\log2+2d\log(2n+1)]/\eta,
\]
such that every map has empirical sign-error weight at most $\eps/2$, and the uniform choice of a map satisfies, for every pair $x,y$,
\[
\Pr\{f(x)\ne f(y)\}\le4\sqrt{2d/\eps}\,\|x-y\|_2+\eta.
\]
\end{lemma}
\begin{proof}
Put $r=\sqrt{\eps/2}$. Use a shared sequence of uniform points in $[-2,2]^d$, and accept the first one in $x+rB_d$. Almost surely the sequence is dense and every input has an accepted point. Restrict the seed space to that probability-one event. Return its coordinate signs, with one fixed convention at zero. If a coordinate sign is wrong, $x_i^2\le(x_i-Z_i)^2$, so the total empirical error is at most $\|x-Z\|^2\le\eps/2$ for every input and every retained seed.

The common rejection sampler of two uniform ball laws disagrees with probability at most twice their TV distance. Slicing balls in the translation direction bounds that TV distance by
$(V_{d-1}/V_d)\|x-y\|/r\le\sqrt d\|x-y\|/r$.
Thus the original map has disagreement probability at most $L_0\|x-y\|$, with $L_0=2\sqrt{2d/\eps}$. Choose $K$ independent such maps. For a fixed pair with original disagreement probability $p$, exponential Markov at $\log2$ gives
\[
\Pr\left\{\frac1K\sum_{j=1}^K\ind\{f_j(x)\ne f_j(y)\}>2p+\eta/2\right\}
\le e^{-K[(2\log2-1)p+\eta\log2/2]}
\le e^{-K\eta/4}.
\]
A union bound over at most $(2n+1)^{2d}$ ordered pairs is at most $1/2$. Hence some collection works simultaneously for every pair. Its error bound is stronger than the displayed $2L_0\|x-y\|+\eta$.

The table can also be specified by finite exhaustive search. Enumerate the finitely many output tables satisfying the hard error constraints, and check their pairwise disagreement fractions against the last displayed bound. All grid coordinates and distances are algebraic when $\eps$ is rational. The unused slack $\eta/2$ ensures that interval refinement eventually certifies a successful table. The search is finite; its time is not asserted to be polynomial. Storing the selected table takes at most $Kd(2n+1)^d$ bits.
\end{proof}

\begin{proof}[Upper bound in \Cref{thm:seed}]
Apply \Cref{lem:finite-seeds} with $\eta=\rho/4$. On two realizable samples the empirical vectors are sign-compatible. For each coordinate count $C_i\sim\operatorname{Bin}(n,p_i)$,
$\E(\sqrt{C_i/n}-\sqrt{p_i})^2\le(1-p_i)/n$ when $p_i>0$, using $(\sqrt a-\sqrt b)^2\le(a-b)^2/b$; zero-mass coordinates contribute zero. Independence of the samples therefore gives
$\E\|x-x'\|^2=2\sum_i\Var(\sqrt{C_i/n})\le2d/n$.
The disagreement probability is at most
$8d/\sqrt{\eps n}+\rho/4<\rho$
under the stated sample budget. Every table row has empirical error at most $\eps/2$, so the same $2^de^{-n\eps/8}$ union bound used in \Cref{sec:locality-proof} proves accuracy. Uniformly selecting one of the $K$ rows uses exactly $\log_2K$ bits, which has the stated order.
\end{proof}

\subsection{An unavoidable logarithmic seed cost}
\begin{proof}[Lower bound in \Cref{thm:seed}]
Use an anchor with label zero and a second point with mass $8\eps t^2$ and label $\ind\{t>0\}$, for $t\in[-1,1]$. For a fixed seed $r$, let $u_r(t)$ be the probability over the finite sample that the output labels the second point one. It is continuous: on each side of zero it is a polynomial in the observation probabilities, and both sides converge at zero to the all-anchor sample. Accuracy at the endpoints implies $\E_r u_r(1)\ge1-\delta$ and $\E_r u_r(-1)\le\delta$. By Markov, seeds with $u_r(1)>1/2$ and $u_r(-1)<1/2$ have total mass at least $1-4\delta$.

For each such seed, continuity gives a parameter $t_r$ with $u_r(t_r)=1/2$. Conditional on that seed, two independent samples disagree at the second point with probability $1/2$. If the seed is an atom of mass $w_r$, validity at $t_r$ forces $w_r/2\le\rho$. A support of at most $K$ atoms therefore has $1-4\delta\le2K\rho$, as claimed. This argument permits improper predictors and arbitrary finite sample size.
\end{proof}

\begin{proposition}[Atomic obstruction on the continuous sphere]\label{prop:atomic}
Let $d\ge2$ and $\eps/2<1$. Suppose a public random map on the unit sphere always has sign-error weight at most $\eps/2$ and has disagreement probability at most $L\|x-y\|$ for all sign-compatible $x,y$. Its seed law has no atoms. On the empirical grid $\mathcal X_{n,d}$, each atom instead has mass at most $L\sqrt{2/n}$; a finite seed support must have size at least $\sqrt n/(\sqrt2L)$.
\end{proposition}
\begin{proof}
An atom of mass $w>0$ must give the same output on each compatible pair at distance below $w/L$. Follow the quarter-circle from $e_1$ to $e_2$ and then the quarter-circle to $-e_1$. Within either arc, sufficiently close points are sign-compatible; subdividing each arc forces the atom's output to be constant along the whole path. At $e_1$ the first output bit must be one, whereas at $-e_1$ it must be zero, by the hard error bound. This is a contradiction. The case $L=0$ is immediate by the same path argument.

For the finite grid use consecutive points
$(\sqrt{(n-j)/n},\sqrt{j/n},0,\ldots,0)$, $j=0,\ldots,n$, followed by the analogous second arc with negative first coordinate. Consecutive distances are at most $\sqrt{2/n}$. If $w>L\sqrt{2/n}$ the same constancy argument applies. The support-size statement follows by summing atom masses.
\end{proof}

\subsection{Distribution-averaged seed compression}\label{sec:seed-average}
The finite-grid construction gives a pairwise guarantee. A second compression argument controls the average over two samples from each realizable distribution. Start with the truncated-power learner using disagreement budget $\rho/8$, and retain its hard empirical budget. For target vector $h$ and marginal $p$, let $F_{p,h}(r)$ be its two-sample disagreement probability conditional on seed $r$. Coupling the $2n$ observations gives
\[
|F_{p,h}(r)-F_{q,h}(r)|\le n\|p-q\|_1.
\]
Take $Q=\lceil16nd/\rho\rceil$ and a rational simplex net with denominator $Q$ and $\ell_1$ covering radius at most $d/Q$. One obtains such a net by rounding coordinates down and assigning the remaining units to coordinates with the largest fractional parts. The net has at most $(Q+1)^d$ points. Including all $2^d$ targets gives at most $[2(Q+1)]^d$ tests.

Choose $K\ge(24/\rho)\log(2[2(Q+1)]^d)$ independent seeds. For a bounded random variable of mean at most $\rho/8$, exponential Markov at $\log2$ shows that the probability its empirical average exceeds $\rho/4$ is at most $e^{-K\rho/24}$. The union bound leaves positive probability that all tests pass. Lipschitz extension from the net then bounds every distribution's average disagreement by $\rho/4+\rho/16$. All selected seeds retain the empirical-error constraint, so their accuracy guarantee is unchanged. This alternative again establishes existence of a finite public seed set, not an efficient uniform construction of it.

\section{A shared rejection filter for untruncated Gibbs proposals}\label{sec:filter-proof}
We prove \Cref{thm:filter}, the reduction in \Cref{sec:module-filter}. The proof controls the proposal law, the empirical acceptance score, and the stopping rule separately. This order is important: an exact marginal sampler alone supplies only the first of these requirements.

\subsection{Entropy and deletion bounds for the proposal}
Let $m\ge2$, $L_0=\lceil\log_2m\rceil$, $\lambda_0=32L_0/\eps$, and $n\ge2\lambda_0$. Write $a=1-\lambda_0/n$ and $\beta=-\log a$, so
\[
\lambda_0/n\le\beta\le2\lambda_0/n\le1.
\]
For a realizable sample put $P_S(h)=e^{-\beta C_h}/Z_S$. Since a consistent hypothesis exists, $Z_S\ge1$. The entropy identity gives
\begin{equation}\label{eq:filter-entropy}
\beta\E_{P_S}C_h+\log Z_S=H(P_S)\le\log m,
\qquad \E_{P_S}\widehat L_S\le\eps/32.
\end{equation}
In particular $P_S\{\widehat L_S\le\eps/4\}\ge7/8$. For rational $a=u/v$, the weights can be represented by integers $u^{C_h}v^{n-C_h}$. Their bit lengths are polynomial in $n$ and the bit lengths of the parameters.

\begin{lemma}[Expected proposal displacement]\label{lem:gibbs-displacement}
For two independent realizable samples of size $n$,
\[
\E\HH(P_S,P_{S'})^2\le2\beta\log m
\le\frac{128L_0^2}{n\eps},\qquad
\E\HH(P_S,P_{S'})\le\frac{8\sqrt2L_0}{\sqrt{n\eps}}.
\]
\end{lemma}
\begin{proof}
Take the unnormalized square-root weights $\psi_S(h)=e^{-\beta C_h/2}$. Deleting observation $i$ increases a weight by the factor $e^{\beta/2}$ exactly when that hypothesis errs on observation $i$. The normalization inequality in \Cref{lem:deletion} consequently bounds the sum of squared deletion displacements by
\[
(e^{\beta/2}-1)^2\E_{P_S}C_h\le\beta\log m.
\]
Here $e^{\beta/2}-1\le\beta$ for $0\le\beta\le1$, and \eqref{eq:filter-entropy} bounds the expected count by $\log m/\beta$. The Hilbert-space deletion inequality then gives twice this quantity for independent full samples. Substitute $\beta\le64L_0/(n\eps)$ and $\log m\le L_0$. Jensen gives the first-moment bound.
\end{proof}

\subsection{The empirical score under a data-dependent proposal}
\begin{lemma}[Gibbs-weighted score envelope]\label{lem:score-envelope}
Let $A_0=1+\log(2m)$ and
$D(S,S')=\E_{h\sim P_S}|\widehat L_S(h)-\widehat L_{S'}(h)|$.
Then
\[
\E D\le\tfrac12\sqrt{A_0\eps/n}+5A_0/n.
\]
If $n\ge A_0/\eps$, it follows that $4\E D/\eps\le40L_0/\sqrt{n\eps}$.
\end{lemma}
\begin{proof}
For a fixed hypothesis of true error $L$, the elementary binomial exponential bound gives
$\Pr\{L>2\widehat L_S+2t/n\}\le e^{-t}$.
For two independent empirical means, Bernstein's bound gives
\[
\Pr\{|\widehat L_S-\widehat L_{S'}|>2\sqrt{Lt/n}+2t/n\}\le2e^{-t}.
\]
One can obtain the latter directly by writing the difference as a sum of $2n$ centered bounded variables. Its total variance is at most $2nL$; expanding their exponential moments and using $|X|^j\le X^2$ for $j\ge2$ gives the Bernstein moment bound, whose quadratic optimization yields the displayed (slightly enlarged) constants.

Let $T_1,T_2$ be the smallest nonnegative parameters making the first and second inequalities hold simultaneously for all $m$ hypotheses. A union bound shows
$\Pr\{T_1>t\}\le\min(1,me^{-t})$ and
$\Pr\{T_2>t\}\le\min(1,2me^{-t})$.
Thus $\E T_i\le A_0$. Put $X=\E_{P_S}L(h)$. Pointwise in the data, the first envelope and \eqref{eq:filter-entropy} give
$X\le\eps/16+2T_1/n$.
The second envelope and Jensen give
$D\le2\sqrt{T_2X/n}+2T_2/n$.
Taking expectations and applying Cauchy--Schwarz, rather than independence between $T_2$ and $X$, yields
\[
\E D\le2\sqrt{A_0(\eps/16+2A_0/n)/n}+2A_0/n
\le\tfrac12\sqrt{A_0\eps/n}+5A_0/n.
\]
Finally $A_0\le3L_0$, $n\eps\ge A_0$, and $L_0\ge1$ imply the stated coarser bound.
\end{proof}

\subsection{Stopping-time synchronization}
\begin{proof}[Proof of \Cref{thm:filter}]
At each trial take a fresh, independently addressed public proposal seed and threshold seed. A marginal proposal has empirical loss at most $\eps/4$ with probability at least $7/8-\zeta\ge3/4$. The midpoint threshold is always above $\eps/4$, so the acceptance probability is at least $3/4$. This proves the expected $4/3$ trial bound.

Fix $S,S'$. Call a trial bad if the two proposed hypotheses differ or, when they are equal, their accept/reject decisions differ. A $K$-point midpoint grid in an interval of length $\eps/4$ distinguishes two scores with probability at most
$\min\{1,4|\widehat L_S(h)-\widehat L_{S'}(h)|/\eps+1/K\}$.
The joint mass of identical proposals at $h$ is at most the first proposal's marginal mass at $h$. Replacing that marginal by $P_S$ costs at most $\zeta$, because the function being integrated is bounded by one. It follows that the bad-trial probability is at most
\[
b(S,S')=c\HH(P_S,P_{S'})+2\zeta+4D(S,S')/\eps+1/K.
\]
The probability that both reject is at most $1/4$. If the two infinite executions ultimately disagree, either the first trial was bad or both rejected and the independent continuation disagreed. Thus their disagreement probability $q$ satisfies $q\le b+q/4$, and hence $q\le4b/3$. This argument uses the actual joint proposal law; it neither assumes independent proposals across the two executions nor conditions on a successful trial.

Cap each execution at $T$ trials and use its fixed consistent fallback. The chance that this changes either infinite output is at most $2\cdot4^{-T}$. Averaging the last bound and applying \Cref{lem:gibbs-displacement,lem:score-envelope} gives
\begin{equation}\label{eq:filter-final}
\Pr\{\text{different outputs}\}\le\frac43\left[
\frac{(8\sqrt2c+40)L_0}{\sqrt{n\eps}}+2\zeta+1/K\right]+2\cdot4^{-T}.
\end{equation}
The stated sample budget ensures all preliminary conditions on $n$. Since $8\sqrt2c+40\le26(c+1)$ for $c\ge1$, its first contribution to \eqref{eq:filter-final} is at most $13\rho/48$. The grid and proposal terms contribute at most $\rho/16$, and the cap contributes at most $\rho/16$. Their sum is $19\rho/48<\rho$.

Every accepted hypothesis has empirical loss below $\eps/2$, and the fallback has loss zero. A union bound over $H$ gives accuracy failure at most $me^{-n\eps/8}\le\delta$, independently of all sampling approximations. Thus the sampler interface suffices for both conclusions.
\end{proof}

\subsection{An explicit bounded-bit proposal sampler}\label{sec:finite-proposal}
To complete the implementation assertion, first consider the ideal public sampler which repeatedly proposes a uniform index $I\in[m]$ and accepts when an independent $U\in[0,1]$ is below $P(I)$. It returns law $P$ and accepts each trial with probability $1/m$. For two laws $P,Q$, the common-proposal coupling has disagreement at most $2\TV(P,Q)\le2\HH(P,Q)$. More explicitly, its joint output law is
\begin{equation}\label{eq:rejection-joint}
J_{ij}=\frac{\ind\{i=j\}\min(P_i,Q_i)+(P_i-Q_i)_+Q_j+P_i(Q_j-P_j)_+}
{1+\TV(P,Q)}.
\end{equation}
To derive this, ignore trials where both reject. The first trial accepted by at least one side has total probability $(1+\TV(P,Q))/m$. If both accept its index, the diagonal term results. If only the first accepts, the second continues with marginal law $Q$, giving the second term; the symmetric event gives the third. Summing the diagonal common-acceptance mass proves the bound.

For $0<\eta<1$, let
\[
Q_0=2m(\lceil\log_2(8/\eta)\rceil+1),\quad
b_I=\lceil\log_2(8Q_0m/\eta)\rceil,\quad
b_U=\lceil\log_2(8Q_0/\eta)\rceil.
\]
At each of $Q_0$ preallocated slots choose $A$ uniformly from $\{0,\ldots,2^{b_I}-1\}$ and propose $1+\lfloor mA/2^{b_I}\rfloor$. Independently choose $B$ uniformly from $\{0,\ldots,2^{b_U}-1\}$ and accept below $\lfloor2^{b_U}P(I)\rfloor$. If the cap is exhausted return one fixed index. Couple $A,B$ to continuous uniforms by adding independent uniform remainders. The index mismatch per trial is at most $m2^{-b_I}$ and the acceptance mismatch is at most $2^{-b_U}$. Over all trials the total is at most $\eta/4$. The ideal cap failure is at most $e^{-Q_0/m}\le\eta/8$. Consequently the single-execution TV error is at most $3\eta/8$, and the two-execution disagreement increases by at most $3\eta/4$.

Taking $\eta=\zeta$ proves the interface with $c=2$ and error at most $\zeta$. Nest these fixed trial slots inside the filter's fixed outer slots. Integer weights from \eqref{eq:filter-entropy} allow exact probability comparisons, so total bit complexity is polynomial in $m,n,1/\zeta$ and parameter bit lengths. In particular it is polynomial in $m,n$ under the theorem's sample budget, which bounds $1/\rho$ by a polynomial in $n$. For $P_d$ an explicit list has $m=2^d$ entries; obtaining a constant-distortion public sampler directly from its $d$ product parameters is the separate computational interface described in the main text.

\section{Computational interfaces and mechanism-specific obstructions}\label{sec:obstructions}
This appendix supplies the supporting propositions for \Cref{sec:module-obstructions}. Each negative result specifies the information or coupling restriction responsible for it. The positive sufficient-statistic result complements the filter by identifying a product subfamily that can be coupled efficiently.

\subsection{A hypothesis-evaluation oracle cannot hide enumeration cost}
\begin{proposition}[Black-box query obstruction]\label{prop:oracle}
In an indexed hypothesis-evaluation model with $m$ hypotheses and $L$ ordinary instances, any learner using $n$ samples and at most $q$ oracle queries, including preprocessing and prediction queries, has an instance with expected prediction error at least
\[
\tfrac12(1-q/m-n/L).
\]
In particular, if $\eps,\delta\le1/16$ and $L\ge8n$, PAC accuracy requires $q\ge5m/8$.
\end{proposition}
\begin{proof}
Draw $m$ independent uniformly random Boolean rows on the $L$ ordinary instances, and choose the target row uniformly. Add $\lceil\log_2m\rceil$ zero-mass code instances that encode the index; this ensures that the hypotheses are distinct. Use the uniform marginal on ordinary instances. An oracle query specifies a row and an instance. Include in $q$ every query made before or after training, including those used to predict an independent test instance.

Until the target row is queried, its index is exchangeable among unqueried rows. The training labels reveal random bits of that row but do not identify it among the unqueried indices. Adaptive queries therefore touch the target with probability at most $q/m$. The independent test instance was seen in the training sample with probability at most $n/L$. On the complement of these two events, its target bit is still fair conditional on the entire transcript, so error is $1/2$. Averaging gives the bound and hence an individual instance attaining it. A PAC learner has expected loss at most $\eps+\delta$, so
$q/m\ge1-2(\eps+\delta)-n/L\ge5/8$.
The argument counts oracle access rather than treating a compiled hypothesis table as free advice.
\end{proof}

\subsection{Normalization is easier than shared coupling}\label{sec:normalization}
For the full Boolean class, write the empirical cost as
$C(v)=\sum_i c_i\ind\{v_i\ne b_i\}$, where $c_i$ are coordinate counts and unseen labels use a fixed convention. The numbers of assignments with cost $t$ satisfy
\[
F_0(0)=1,\qquad F_i(t)=F_{i-1}(t)+F_{i-1}(t-c_i),
\]
with negative indices interpreted as zero. Hence a truncated-power normalizer is
$Z=\sum_{t<B}F_d(t)(B-t)^k$ and can be computed in $O(dn)$ arithmetic operations. Sampling the cost with these weights and then backtracking through the recurrence gives the exact desired marginal. This calculation establishes a normalization and sampling procedure; its backtracking coupling is a further property to analyze.

The distinction is visible already for product Bernoulli laws. Let
$P=\operatorname{Ber}(1/2)^{\otimes d}$ and
$Q=\operatorname{Ber}(1/2+\eta)^{\otimes d}$.
Using the same uniform number separately in each coordinate gives mismatch
$1-(1-|\eta|)^d$, asymptotic to $d|\eta|$ as $\eta\to0$ at fixed $d$. On the other hand
$\KL(P\|Q)=-(d/2)\log(1-4\eta^2)$, so Pinsker gives $\TV(P,Q)=O(\sqrt d|\eta|)$. Thus exact coordinate marginals alone do not give a constant-TV-distortion common sampler.

\subsection{A fixed statistic preserves the relevant distances}\label{sec:fixed-statistic}
\begin{proposition}[Sufficient-statistic lift]\label{prop:lift}
Suppose $T$ is fixed and
$P_\theta(v)=\pi_\theta(T(v))Q_{T(v)}(v)$, where $Q_z$ is independent of $\theta$ and supported on the fiber $T^{-1}(z)$. Then
\[
\TV(P_\theta,P_{\theta'})=\TV(\pi_\theta,\pi_{\theta'}),\qquad
\HH(P_\theta,P_{\theta'})=\HH(\pi_\theta,\pi_{\theta'}).
\]
A public sampler for the statistic followed by a shared conditional sampler has exactly the statistic sampler's disagreement probability.
\end{proposition}
\begin{proof}
Sum the absolute difference, or the squared square-root difference, separately over each fiber. In either case the common factor $Q_z$ sums to one, proving the identities. If the two statistic values agree, the same conditional input and seed give the same full output. If the statistic values differ, the outputs lie in disjoint fibers and necessarily differ.
\end{proof}

For fixed positive rational $a_i$, consider
$P_t(v)\propto t^{|v|}\prod_i a_i^{v_i}$.
Define the elementary-symmetric sums recursively by
\[
E_{d+1}(0)=1,\qquad E_i(k)=E_{i+1}(k)+a_iE_{i+1}(k-1).
\]
Then $\pi_t(k)=E_1(k)t^k/\prod_i(1+ta_i)$, while the conditional probability of $v_i=1$ given $k$ remaining ones is
$a_iE_{i+1}(k-1)/E_i(k)$.
It is independent of $t$. The recurrence costs $O(d^2)$ rational operations. Sampling the $d+1$-state statistic by common rejection and then using these fixed conditionals realizes \Cref{prop:lift}; downward quantization to $b$ bits changes the conditional output with probability at most $d2^{-b}$. This gives a finite-bit approximation with explicitly controlled error.

A sample-dependent statistic does not have that property. On two bits take
$P(v)\propto r^{v_1+2v_2}$ and $Q(v)\propto r^{2v_1+v_2}$, $0<r<1$. Their respective cost statistics have the identical distribution proportional to $(1,r,r^2,r^3)$. Coupling these costs identically nevertheless gives output mismatch
\[
\frac{r+r^2}{(1+r)(1+r^2)}=\frac r{1+r^2},
\quad\text{whereas}\quad
\TV(P,Q)=\frac{r-r^2}{(1+r)(1+r^2)}.
\]
The ratio is $(1+r)/(1-r)$ and diverges. This example occurs in the actual filter proposal family: use three coordinate counts $(1,2,n-3)$ and $(2,1,n-3)$, all labels zero, and $r=1-(96/\eps)/n$ with $n>6$ and $n\ge192/\eps$. The third cost distinguishes two disjoint four-point cost ranges. The same mismatch persists and tends to $1/2$, while the TV distance tends to zero.

Nor can a nontrivial fixed statistic suffice for the entire family of positive Bernoulli products. Write their probabilities as proportional to $e^{\theta\cdot v}$. If distinct $v,w$ belonged to one fiber with a parameter-independent conditional law, their likelihood ratio $e^{\theta\cdot(v-w)}$ would be independent of $\theta\in\R^d$, which forces $v=w$. Thus such a statistic has at least $2^d$ states on this family.

\subsection{Greedy shared-capacity rounding can be highly discontinuous}\label{sec:greedy}
Consider costs $2^{-1},\ldots,2^{-d}$ in this order. A greedy rule takes an item whenever it fits within capacity $R\sim\Unif[1/4,1/2]$. Except at finitely many boundaries, the output subset is the binary expansion of $\lfloor2^dR\rfloor$. Scale every cost by $1+\eta$, where $\eta=2^{-2d}$. The boundary at $j/2^d$ moves to $(1+\eta)j/2^d$. For $L=2^{d-2}$, the intervals between old and new boundaries are disjoint for $j=L,\ldots,2L-1$, and remain in the capacity interval up to null endpoints. Their total relative length is
\[
4\sum_{j=L}^{2L-1}\frac{\eta j}{2^d}
=\eta\frac{3\cdot2^{d-2}-1}{2}.
\]
This is the exact disagreement probability under the same capacity. It rules out a dimension-independent Lipschitz claim for this particular greedy map. It makes no assertion about unrestricted learning algorithms.

\subsection{A sufficient square-root rounding interface}\label{sec:rounding-interface}
The following conditional implication explains the normalization relevant to efficient implementations. Suppose a polynomial-time public map on signed empirical vectors always has error weight at most $\eps/2$ and, for sign-compatible vectors, has mismatch at most
$C\sqrt{d/\eps}\|x-x'\|_2$.
The empirical estimate in \Cref{sec:randomness-proof} gives mismatch at most $C\sqrt2d/\sqrt{\eps n}$. Thus
\[
n\ge\max\{2C^2d^2/(\eps\rho^2),\ 8(d\log2+\log(1/\delta))/\eps\}
\]
would suffice for optimal-rate polynomial-time learning of $P_d$. An additive term $\rho/4$ can be incorporated with the corresponding constant slack. \Cref{lem:finite-seeds} supplies existence on the empirical grid, with large table cost; \Cref{thm:filter} provides an alternative interface using ordinary product proposals.

\subsection{A cardinality-first knapsack rule: a same-distribution obstruction}
We prove a stronger obstruction for a natural globally optimized rounding rule. On $D$ coordinates, draw a shared random permutation with ranks $\pi(i)\in[D]$ and a shared $R\sim\Unif[\eps/4,\eps/2]$. Give ignored coordinate $i\in\{0,\ldots,D-1\}$ profit
\[
z_i=2^D(D^3+\pi(i))-2^i.
\]
Choose the subset of maximum total profit subject to total empirical cost at most $R$, set ignored labels to zero, and retain observed labels on the others. Profits strictly order subsets first by cardinality, then by rank sum, then by smaller binary mask. The $D^3$ term dominates every rank-sum difference, and the mask removes all ties. An integer knapsack dynamic program runs in $O(Dn)$ arithmetic operations. Every output has empirical error at most $R$.

\begin{proposition}[Knapsack lower bound at fixed dimension]\label{prop:knapsack}
For each $d\ge64$ and $0<\eps\le1/16$, there is a fixed realizable distribution on $D=d+1$ coordinates for which the preceding rule has
\[
\liminf_{n\to\infty}\sqrt{\eps n}\,
\Pr\{\text{two outputs differ}\}\ge d^2/256.
\]
Consequently, for fixed $d,\eps$ and sufficiently small $\rho$, any valid choice of sample size for this rule obeys
$n\ge2^{-18}d^4/(\eps\rho^2)$.
\end{proposition}
\begin{proof}
Use $d$ active coordinates of label one and one anchor of label zero. Put
\[
L=4d2^d,\qquad a=2\eps/d,\qquad
p_i=a(1+2^{i-1}/L),\quad i=1,\ldots,d.
\]
The remaining mass is at the anchor. The total active mass is less than $2\eps(1+1/(4d^2))$, so the anchor has mass greater than $1/2$. At the population costs the anchor cannot be ignored, and each active subset $S$ has distinct cost
$W(S)=a(|S|+\operatorname{mask}(S)/L)$.
Cost order is cardinality followed by binary mask. All cost gaps are at least $a/L$.

Fix a permutation. For $r\in[\lceil d/8\rceil+1,\lfloor d/4\rfloor-1]$ and $j\in[\lceil d/2\rceil,d]$, let $S_{r,j}$ be the $r$ highest-ranked active coordinates among the first $j$. If $j\in S_{r,j}$, this subset becomes a new optimal solution exactly at its cost $W(S_{r,j})$. To check this, any equal-cardinality subset using an index above $j$ has larger cost. Among those using only indices at most $j$, $S_{r,j}$ uniquely maximizes rank sum. Lower cardinalities have lower profit, and higher cardinalities have larger cost. These record costs lie strictly in $I=[\eps/4,\eps/2]$. Distinct $(r,j)$ witnesses give distinct subsets. Since $\Pr\{j\in S_{r,j}\}=r/j\ge1/8$, there are in expectation at least
\begin{equation}\label{eq:knapsack-records}
(d/16)(d/2)(1/8)=d^2/256
\end{equation}
such records. The permutation may include the anchor; its restriction to active coordinates has uniform relative order, which is all that this argument uses.

One may first see the instability by scaling all active costs by $1+t$ and adjusting the anchor mass, with $0<t\le1/(16dL)$. The old and new intervals of every witnessed record boundary are disjoint. Their total relative capacity length is at least
$(4t/\eps)\E_\pi\sum W(S_{r,j})\ge td^2/256$.
The signed square-root vectors move by at most $2t\sqrt\eps$. Thus a bound $C\sqrt{D/\eps}\|x-x'\|$ for this rule would require $C\ge d^2/(512\sqrt{d+1})$.

For the learning claim, keep the distribution fixed and take two independent samples. There are finitely many subset costs and permutation orders. No active subset cost equals an endpoint of $I$. Indeed in units $a/L$ the endpoints are $dL/8$ and $dL/4$; a nonzero binary mask is less than $2^d$, whereas any nonzero fractional-cardinality offset has magnitude at least $L/8>2^d$. Thus a positive separation exists among all relevant costs and the endpoints. By finitely many binomial concentration bounds, with probability $1-O(e^{-c n})$ both empirical subset-cost lists preserve their population order, their record list, and the position of each record relative to the endpoints. The anchor remains infeasible as well. Constants here may depend on the fixed $d,\eps$.

On this event each record boundary merely moves inside a disjoint small neighborhood. Since every active subset has a different output vector, the conditional shared-capacity disagreement is exactly
\[
\frac4\eps\sum_{S\text{ record},\ W(S)\in I}
|\widehat W_S(S)-\widehat W_{S'}(S)|.
\]
For a fixed subset of mass $W$, the two counts are independent binomials. The central limit theorem and uniform integrability (from their bounded scaled second moments) give
\[
\sqrt n\,\E|\widehat W_S-\widehat W_{S'}|
\longrightarrow\frac2{\sqrt\pi}\sqrt{W(1-W)}.
\]
The exceptional exponential event has negligible contribution after multiplication by $\sqrt n$. Finite sums over subsets and permutations therefore yield
\[
\lim_{n\to\infty}\sqrt n\,\Pr\{\text{different outputs}\}
=\frac8{\eps\sqrt\pi}\E_\pi\sum_{S\text{ record},\ W(S)\in I}\sqrt{W(S)(1-W(S))}.
\]
Each witnessed term is at least $\sqrt{\eps/8}$. Combining with \eqref{eq:knapsack-records} proves the stated, weaker, constant $d^2/256$ after multiplying by $\sqrt\eps$.

For every fixed finite $n$, disagreement is positive. For example, one sample can consist entirely of anchor observations and the other entirely of one active coordinate. On the first sample every active coordinate has zero cost and is ignored. On the second, the observed active coordinate has empirical cost one and cannot be ignored. Both events have positive probability and force different outputs. Consequently sample sizes achieving $\rho\to0$ must tend to infinity. The limit inferior bound then gives mismatch at least $d^2/(512\sqrt{\eps n})$ for all sufficiently large $n$, proving the last assertion.
\end{proof}

The same asymptotic obstruction has a bounded-bit realization. Quantize the capacity to $b\ge D+4\lceil\log_2(n+1)\rceil+4$ bits. There are at most $2^D$ subset boundaries, so the probability of changing a continuous-capacity decision is $O(n^{-4})$. Generate the permutation by sorting independent public keys of $4\lceil\log_2(n+1)\rceil+2\lceil\log_2D\rceil+4$ bits. Key ties have probability $O(n^{-4})$; conditional on no tie the order is uniform. These errors vanish after scaling by $\sqrt n$. The proposition therefore concerns an implementable rule rather than an infinite-precision artifact.

\end{document}